%% file: acl2023.tex
\pdfoutput=1

\documentclass[11pt]{article}

\usepackage{ACL2023}

\usepackage{times}
\usepackage{latexsym}

\usepackage[T1]{fontenc}

\usepackage[utf8]{inputenc}

\usepackage{microtype}

\usepackage{inconsolata}

\usepackage{hyperref}
\usepackage{microtype}
\usepackage{graphicx}
\usepackage{booktabs}
\usepackage{subfig} 
\usepackage{placeins} 
\usepackage{pdfpages}

\title{Crosslingual Generalization through Multitask Finetuning}

\author{
Niklas Muennighoff$^{1}$
\quad Thomas Wang$^1$ 
\quad Lintang Sutawika$^{2,3}$ 
\quad Adam Roberts$^4$
\quad Stella Biderman$^{3,5}$\\
\bf \quad Teven Le Scao$^1$
\bf \quad M Saiful Bari$^6$ 
\bf \quad Sheng Shen$^7$ 
\bf \quad Zheng-Xin Yong$^8$
\bf \quad Hailey Schoelkopf $^{3,9}$\\
\bf \quad Xiangru Tang $^9$ 
\bf \quad Dragomir Radev$^9$ 
\bf \quad Alham Fikri Aji$^{10}$
\bf \quad Khalid Almubarak$^{11}$\\
\bf \quad Samuel Albanie$^{12}$
\bf \quad Zaid Alyafeai$^{13}$
\bf \quad Albert Webson$^8$
\bf \quad Edward Raff$^{5}$
\bf \quad Colin Raffel$^1$ \\
  $^1$Hugging Face \quad
  $^2$Datasaur.ai \quad
  $^3$EleutherAI \quad
  $^4$Google Research, Brain Team \quad
  $^5$Booz Allen Hamilton \quad\\
  $^6$Nanyang Technological University \quad
  $^7$ UC Berkeley \quad
  $^8$ Brown University \quad \\
  $^9$ Yale University \quad
  $^{10}$ MBZUAI \quad  
  $^{11}$ PSAU \quad
  $^{12}$ University of Cambridge 
  $^{13}$ KFUPM\\
  {\tt \href{mailto:niklas@hf.co}{niklas@hf.co}}
}

\begin{document}
\maketitle
\begin{abstract}

Multitask prompted finetuning (MTF) has been shown to help large language models generalize to new tasks in a zero-shot setting, but so far explorations of MTF have focused on English data and models. We apply MTF to the pretrained multilingual BLOOM and mT5 model families to produce finetuned variants called BLOOMZ and mT0. We find finetuning large multilingual language models on English tasks with English prompts allows for task generalization to non-English languages that appear only in the pretraining corpus. Finetuning on multilingual tasks with English prompts further improves performance on English and non-English tasks leading to various state-of-the-art zero-shot results. We also investigate finetuning on multilingual tasks with prompts that have been machine-translated from English to match the language of each dataset. We find training on these machine-translated prompts leads to better performance on human-written prompts in the respective languages. Surprisingly, we find models are capable of zero-shot generalization to tasks in languages they have never intentionally seen. We conjecture that the models are learning higher-level capabilities that are both task- and language-agnostic. In addition, we introduce xP3, a composite of supervised datasets in 46 languages with English and machine-translated prompts. Our code, datasets and models are freely available at \url{https://github.com/bigscience-workshop/xmtf}.

\end{abstract}

\section{Introduction}

\begin{figure*}[ht]
    \centering
    \begin{center}
        \includegraphics[width=\textwidth]{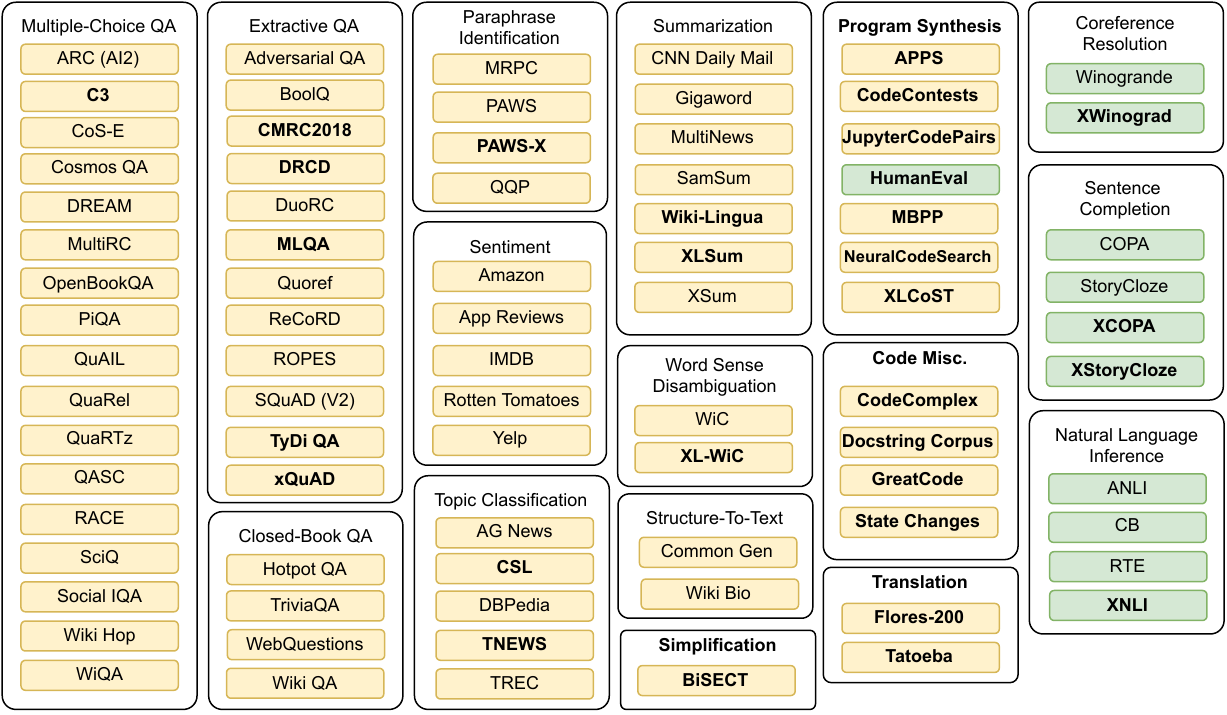}
        \caption{An overview of datasets in xP3. Datasets added to P3 in this work are marked \textbf{bold}. Yellow datasets are trained on. Green datasets are held out for evaluation.}
        \label{fig:xp3tasks}
    \end{center}
\end{figure*}

\begin{figure*}[ht]
    \centering
    \begin{center}
        \includegraphics[width=\textwidth]{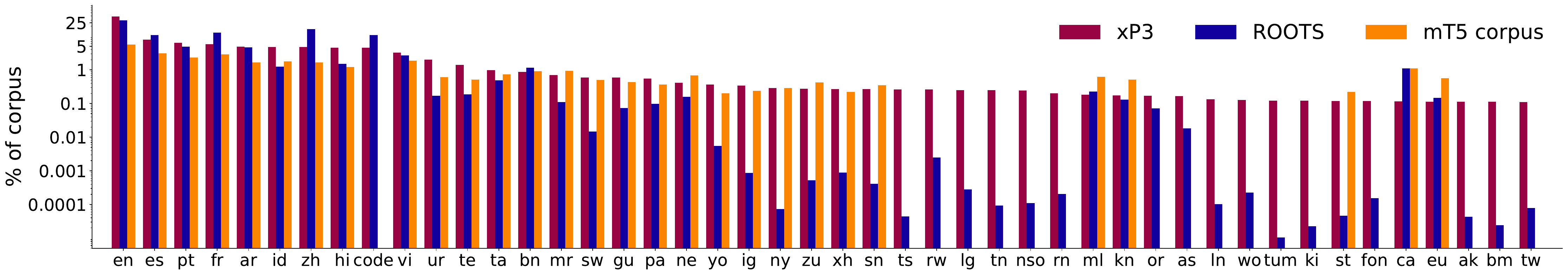}
        \caption{Language composition of xP3, ROOTS, and the corpus of mT5. All ROOTS and xP3 languages are depicted. The mT5 corpus covers additional languages that are not included in the graph.}
        \label{fig:xp3langs}
    \end{center}
\end{figure*}

\begin{figure*}[ht]
\includegraphics[width=\textwidth]{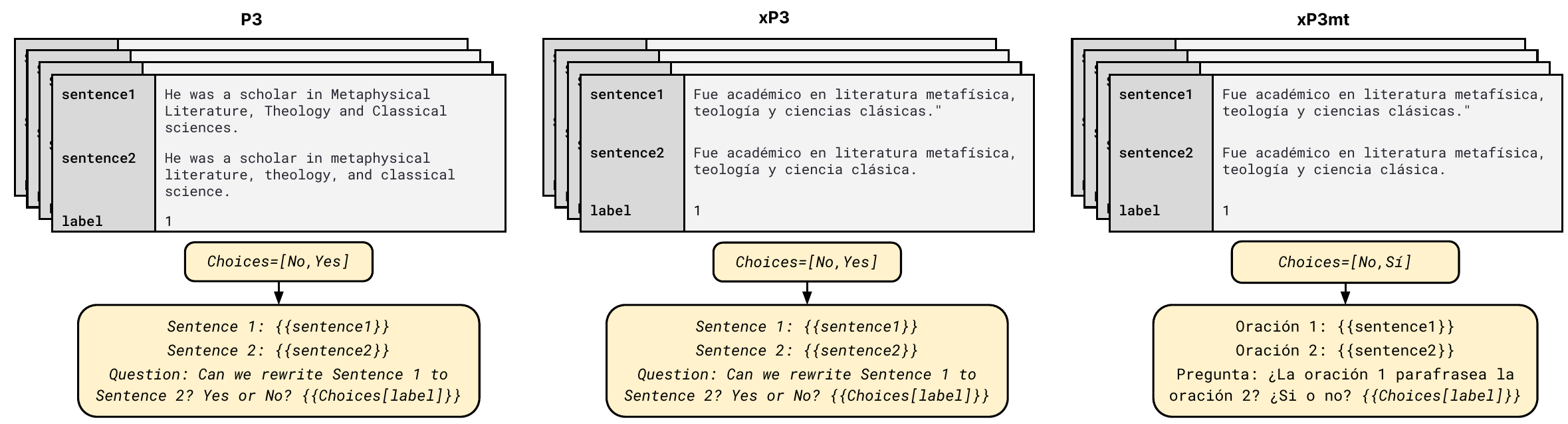}
\centering
\caption{Comparison of dataset variants P3, xP3, and xP3mt on a sample from PAWS for P3 \cite{zhang2019paws} and PAWS-X \cite{pawsx2019emnlp} for xP3 and xP3mt. P3 pairs English datasets with English prompts, xP3 pairs multilingual datasets with English prompts and xP3mt pairs multilingual datasets with prompts machine-translated from English to match the dataset language. Expressions in curly brackets are replaced, e.g. for xP3mt the target shown as \texttt{\{\{Choices[label]\}\}}) becomes  \texttt{Sí}.}\label{fig:dataset-diagram}
\end{figure*}

Large language models pretrained on vast amounts of text show some capability of solving tasks expressed in natural language, even without explicit training on these tasks \cite{gpt3}. Finetuning on groups of language tasks has been shown to significantly boost this zero-shot task generalization of language models~\cite{wei2021finetuned, sanh2022multitask, min2021metaicl}. For example, \citet{sanh2022multitask} finetune on tasks like summarization and question answering leading to better performance on unseen tasks like natural language inference. Previous work has focused on multitask finetuning in the context of large English language models and tasks.

Multilingual large language models show the same zero-shot learning capabilities for both monolingual and crosslingual tasks \cite{goyal2021larger,lin2021few,patel2022bidirectional,saleh2022}. However, zero-shot performance tends to be significantly lower than finetuned performance. Thus, task-specific or language-specific transfer learning via finetuning remains the predominant practice \cite{devlin2018bert,conneau2019unsupervised,ExT5}. This is particularly challenging for low-resource languages or tasks with limited data available, such as writing a fable that teaches a specified moral. In the spirit of multitask finetuning, it would be desirable to improve the zero-shot task generalization of multilingual models to make them usable on tasks from low-resource languages without requiring further finetuning.

To address this goal, we focus on crosslingual multitask finetuning. Due to the difficulty of collecting supervised task data in low-resource languages, previous work typically aims to transfer capabilities learned from finetuning on English data, which can improve performance on non-English language tasks \cite{wu-dredze-2019-beto,phang2020english,chalkidis2021multieurlex,vu2022overcoming}. We investigate whether English-only multitask finetuning also improves performance on non-English held-out tasks using the multilingual BLOOM~\cite{scao2022bloom} and mT5~\cite{xue2020mt5} models. We find that after finetuning on the English-only multitask mixture used for T0~\cite{sanh2022multitask} (P3), performance on a diverse set of non-English held-out tasks increases.

To investigate whether multilingual task data can further improve performance, we extend P3 to xP3 by adding datasets from 46 different languages that cover tasks previously not present in P3 (such as translation and program synthesis). Finetuning on xP3 leads to even better zero-shot task generalization in both English and non-English compared to the P3-trained baseline. Models finetuned on xP3 perform best on English prompts, even for non-English samples. Hypothesizing that better performance could be attained by training on non-English prompts, we construct a variant of xP3 with machine-translated prompts called xP3mt. We find that finetuning on machine-translated prompts is enough to significantly increase performance on held-out tasks with non-English human-written prompts. However, reducing the number of English prompts in the finetuning also worsens English prompt performance on multilingual tasks. 

Notably, we also find that models finetuned on xP3 generalize to held-out tasks in languages never intentionally seen during pretraining nor finetuning. We conduct a contamination analysis and find that only small amounts of these languages were included in the pretraining corpus. Thus, we hypothesize the models learn some language- and task-agnostic capabilities.

We publicly release all our datasets and models (URLs in Appendix \S\ref{sec:artifacts}).

\section{Related work}

\subsection{Multitask learning}

Multitask finetuning \cite{sanh2022multitask} (or instruction tuning \cite{wei2021finetuned}) has emerged as a recipe for improving the zero-shot task generalization of large language models. Typically, these works define a task as a collection of datasets that require a certain set of skills. To inform large language models which task to perform given an input, a prompt is used to add natural language instructions to dataset instances \cite{schick2020exploiting,scao2021many}. In this line of work, zero-shot task generalization refers to the ability to perform a held-out task based on prompted instructions alone. Our work builds on T0~\citep{sanh2022multitask}, a variant of T5~\citep{raffel2020exploring} that underwent MTF and was subsequently shown to have strong zero-shot task generalization capabilities.

Increasing the number and diversity of finetuning tasks and datasets has been shown to increase model performance \cite{min2021metaicl,fries2022bigbio,wang2022zemi,scialom2022continual,chung2022scaling,mishra2021}. PromptSource \cite{promptsource} is a software application that provides a framework for developing and applying prompts. PromptSource was used to construct P3, the training dataset of T0. While most prior work has focused on using English prompts on English datasets, \citet{wang2022super} trained both English and multilingual models on prompted datasets. Their multilingual model, called mTk-Instruct, attains strong crosslingual performance. In contrast with \citet{wang2022super}, our sole focus is crosslingual zero-shot generalization. Therefore, we consider a wider variety of prompting settings and perform a more detailed evaluation of multilingual capabilities. Separately, \citet{radford2019language} find that accidental inclusion of non-English text gave the GPT-2 model a limited ability to process and generate non-English text. We similarly discover that our finetuned models can process text in languages not intentionally trained on.

\subsection{Multilingual models}

Many language models are pretrained on English data only. Multilingual pretrained language models \cite{lample2019,conneau2019unsupervised,fan2021beyond} aim to enable processing a wide variety of non-English languages. Unlike monolingual models, multilingual models can also be used for crosslingual tasks, such as translation. For language generation, recent efforts have focused on two different model architectures based on the Transformer~\cite{vaswani2017attention}. On the one hand, encoder-decoder transformers trained with a denoising objective such as mBART \cite{liu2020multilingual} and mT5 \cite{xue2020mt5} learn to predict tokens masked out in the input sequence. Predicting masked tokens is only a pretraining task and these models are generally finetuned on downstream datasets before being used. On the other hand, decoder-only models pretrained on next token prediction such as mGPT \cite{shliazhko2022mgpt}, XGLM \cite{lin2021few} and BLOOM \cite{scao2022bloom} can be used to solve tasks expressed in natural language directly in a zero-shot or few-shot setting \cite{gpt3}. XGLM demonstrated competitive few-shot performance even when the model was prompted in a language different than the sample being processed. In particular, using English prompts for multilingual datasets provides better performance with XGLM than human-translating the English prompt to the dataset language. 

In this work, we use the BLOOM models~\cite{scao2022bloom,scao2022language}, which were pretrained on the ROOTS corpus \cite{laurencconbigscience} in 46 natural languages and 13 programming languages. We also finetune mT5~\cite{xue2020mt5} to compare encoder-decoder and decoder-only performance. mT5 is pretrained on a corpus sampled from mC4 covering 101 languages.

\section{Finetuning data and models}

To study crosslingual multitask prompted finetuning, we create xP3 by extending the P3 dataset collection with additional non-English tasks. We finetune both BLOOM and mT5 models on xP3. We refer to Appendix~\S\ref{sec:artifacts} for public links to released models and datasets. 

\subsection{Finetuning data}

We build on the P3~\cite{sanh2022multitask} task taxonomy and add 30 new multilingual datasets illustrated in Figure~\ref{fig:xp3tasks}. We define four task clusters previously not present in P3: translation, simplification, program synthesis, and miscellaneous code datasets. As 11\% of BLOOM's pretraining data is code, we add code datasets classified as program synthesis (text-to-code) or miscellaneous. The latter includes tasks such as estimating the computational complexity of a provided code snippet and generating a name for a given function. We extend the XWinograd dataset~\cite{tikhonov2021heads} with winograd schemas from CLUE~\cite{xu2020clue} to increase its Chinese samples from 16 to 504. Similar to P3, a fraction of our prompts invert the task at hand. For example, a prompt may invert a closed-book QA sample by asking the model to generate a question given an answer.

With xP3 we aim to replicate the language distribution of the ROOTS corpus \cite{laurencconbigscience} used to pretrain BLOOM. Thus, xP3 consists of the same 46 natural languages and code as ROOTS. ROOTS, xP3 and the mT5 corpus~\cite{xue2020mt5} language distributions are visualized in Figure~\ref{fig:xp3langs}. 39\% of xP3 data is English, slightly more than the 30\% of English data in ROOTS. Various African languages such as Twi (tw) and Bambara (bm) form the tail of xP3's language distribution. Many of them are not included in the mT5 pretraining corpus. In xP3, Twi and others are represented solely as a translation task using data from Flores-200~\cite{nllb2022}. 

To study the importance of non-English prompts, we construct a machine-translated variant of xP3, xP3mt. We translate prompts of monolingual datasets into the respective dataset language. For example, for the Chinese dataset C3~\cite{sun2020investigating} prompts in xP3mt are in Chinese instead of English in xP3. For crosslingual datasets prompts remain in English in xP3mt (such as Wiki-Lingua, which involves producing a summary in one language based on text in another language). We use the Google Cloud API for machine translation\footnote{\url{https://cloud.google.com/translate}}. Figure \ref{fig:dataset-diagram} compares the dataset variants we train on.

\subsection{Models}
\label{sec:models}

\begin{figure*}[ht]
\includegraphics[width=\textwidth]{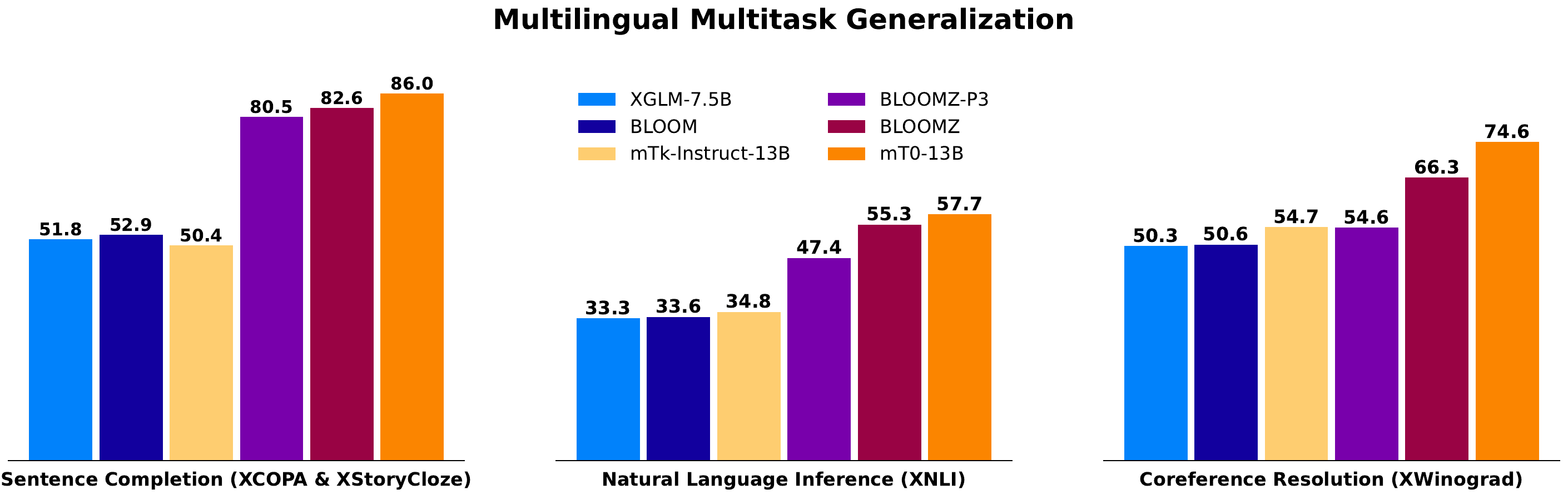}
\centering
\caption{Zero-shot multilingual task generalization with English prompts. BLOOM models have 176 billion parameters. Scores are the language average for each task. Appendix~\S\ref{sec:taskgenlang} breaks down performance by language.}
\label{fig:taskgen}
\end{figure*}

We use publicly available pretrained BLOOM models ranging from 560 million to 176 billion parameters. BLOOM models are large decoder-only language models pretrained for around 350 billion tokens with an architecture similar to GPT-3 \cite{gpt3}. We finetune the models for an additional 13 billion tokens with loss only being computed on target tokens. For example, given the input ``Translate to English: Je t'aime." and a space-separated target ``I love you.", the model is trained to predict only the targets. As targets vary in length from just one to hundreds of tokens, we downscale the loss of each token by the length of the target it belongs to. This ensures short targets (e.g.\ for multiple-choice QA) get the same weight as long targets (e.g. for translation). We skip samples longer than 2048 tokens and use packing to train efficiently on multiple samples at a time \cite{kosec2021packing}. We select the final checkpoint based on validation performance.

For mT5 models, we finetune using the T5X~\cite{roberts2022t5x} framework on TPUs. mT5 uses the same encoder-decoder architecture, pretraining objective (masked language modeling), and pretraining length (1 trillion tokens) as T5~\cite{raffel2020exploring}. For finetuning mT5, we follow the same procedure as described above for BLOOM, except that inputs are fed into the encoder and thus are not space-separated from targets. 

We produce three core model variants available in different sizes:

\begin{itemize}
    \item \textbf{BLOOMZ-P3 / mT0-P3:} Models finetuned on the English-only P3.
    \item \textbf{BLOOMZ / mT0:} Models finetuned on xP3, which consists of multilingual datasets with English prompts.
    \item \textbf{BLOOMZ-MT / mT0-MT:} Models finetuned on xP3mt, which consists of multilingual datasets with English and machine-translated prompts.
\end{itemize}

We evaluate on three held-out tasks: coreference resolution, sentence completion and natural language inference (NLI) as depicted in Figure~\ref{fig:xp3tasks}. We also evaluate on HumanEval due to its popularity for code evaluations using the pass@k metric~\cite{chen2021evaluating}. For datasets that involve choosing the correct completion from several options, we follow prior work~\cite{sanh2022multitask,gpt3} and use rank classification: We compute the log-likelihood of each possible completion and select the highest scoring option. For each evaluation dataset, we select 5 prompts at random from PromptSource and use them for all language splits of the dataset. We report the median of the 5 prompts for results per language split. Thus, in constrast to XGLM~\cite{lin2021few}, we do not tune prompts based on performance on validation data. A selection of prompts can be found in Appendix~\S\ref{sec:prompts}. For evaluation on generative tasks, such as translation, we use lm-evaluation-harness \cite{eval-harness} and report BLEU scores~\cite{papineni2002bleu}.

\section{Results}

\begin{figure*}[ht]
\includegraphics[width=\textwidth]{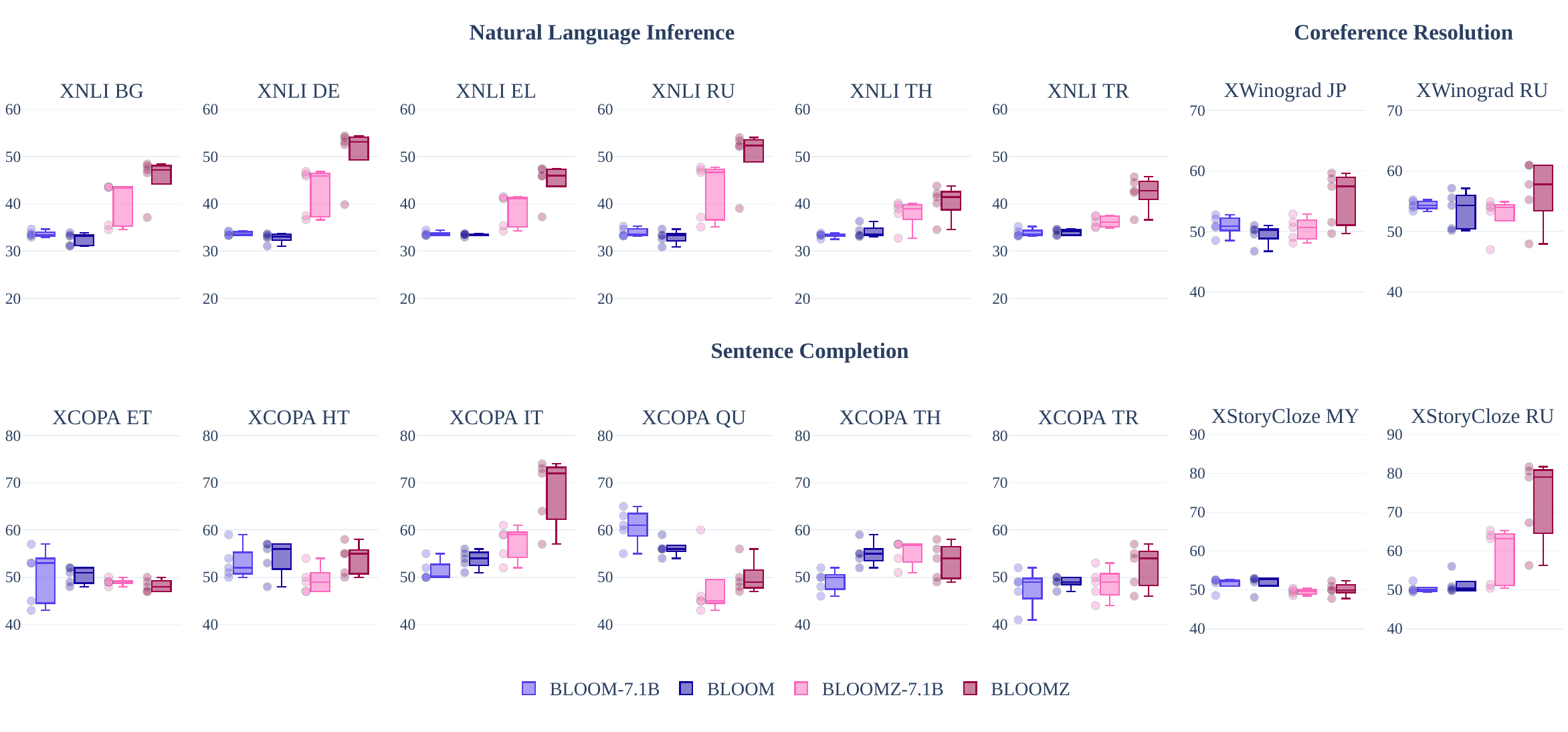}
\centering
\caption{Zero-shot task and language generalization using English prompts on tasks and languages not intentionally seen during pretraining nor finetuning. Language codes are ISO 639-1, except for JP (Japanese).}
\label{fig:langgen}
\end{figure*}

We first examine generalization to new tasks in languages included in finetuning in~\S\ref{sec:taskgen}. Then, in~\S\ref{sec:langgen}, we look at language generalization: Can models generalize to tasks in languages that \textbf{(a)} they have only seen during pretraining and \textbf{(b)} they have never seen intentionally? In~\S\ref{sec:multilingualprompting}, we investigate performance on multilingual prompts and finetuning on xP3mt. Scaling laws are analyzed in~\S\ref{sec:scaling}. Finally, \S\ref{sec:generation} looks at performance on generative tasks and \S\ref{sec:corpus} at the effect of language proportions on performance.

\subsection{Task generalization}
\label{sec:taskgen}

Previous work has shown that large language models finetuned on prompted multitask mixtures generalize to unseen tasks \cite{zhong2021adapting,wei2021finetuned,mishra2021,mishra2021cross,wang2022super}. In Figure \ref{fig:taskgen}, we show that the same applies to multilingual models: Finetuned BLOOMZ and BLOOMZ-P3 models significantly improve over BLOOM and XGLM on held-out tasks. Despite an order of magnitude fewer parameters, mT0 (13 billion parameters) is ahead of BLOOMZ (176 billion parameters). We attribute this to the encoder-decoder architecture paired with a masked language modeling pretraining objective \cite{wang2022language,tay2022unifying} as well as the longer pretraining of mT5 \cite{hoffmann2022training,su2022welm} (1 trillion tokens for mT5 vs.\ 366 billion for BLOOM). Despite also having gone through crosslingual multitask finetuning, mTk-Instruct performs significantly worse than the same-sized mT0. We attribute this to our prompting style, which aims to replicate natural human communication. mTk-Instruct is finetuned on more structured prompts with specific ``Definition", ``Input" and ``Output" fields. Similarly, \citet{wang2022super} find that T0 performs worse than Tk-Instruct on their prompts. We also find models finetuned on the 39\% English xP3 (BLOOMZ, mT0-13B) outperform models finetuned on the 100\% English P3 (BLOOMZ-P3, mT0-13B-P3) on \emph{English tasks} (Appendix~\S\ref{sec:taskgenlang}). Even the fully English T0-11B model \cite{sanh2022multitask} is outperformed by our mT0-13B model on entirely \emph{English tasks}. Ignoring embedding parameters T0-11B and mT0-13B have about the same size. This is likely due to xP3 adding additional tasks and prompts, which has been shown to help generalization~\cite{chung2022scaling,iyer2022opt}. mT0-13B beating T0-11B indicates that the benefit of scaling tasks is larger than the benefit of pre-training and finetuning on relatively more English tokens. 

\subsection{Language generalization}
\label{sec:langgen}

Here we add another layer of generalization: languages. Figure \ref{fig:taskgen} already shows that finetuning on English data only (P3) leads to better performance on non-English data: For example, BLOOMZ-P3 improves by over 50\% on multilingual sentence completion compared to BLOOM. Thus, zero-shot task performance in languages only seen during pretraining improves after finetuning on English. This has major practical benefits as it can be more difficult to collect data for low-resource languages.

Next, we investigate performance on languages the model has \textit{never intentionally seen}. Due to the scale of large language model pretraining, it is difficult to label tasks or languages as strictly unseen. It is likely that the training data unintentionally includes small fractions of these languages (just as many tasks might appear ``implicitly'' in the pretraining corpus~\cite{sanh2022multitask}). In Figure~\ref{fig:langgen} we show that after multitask finetuning on xP3, the models can perform unseen tasks in languages that were not intentionally trained on. After probing the pretraining corpus of BLOOM, we do find small amounts of these languages that were unintentionally included (Appendix~\S\ref{sec:contamination}). However, for XNLI, performance increases across all languages, many of which only show up in tiny fractions in our language contamination analysis, such as Thai with 0.006\%. If we extrapolate this proportion to the entire ROOTS corpus, the BLOOM models would have seen a mere 20 million tokens of Thai during pretraining. One possibility is that better-than-random XNLI performance can be attained with little or no language understanding. In Appendix~\S\ref{sec:levenshtein}, we investigate edit distances of XNLI samples and find that there are differences across labels, however, likely not significant enough to enable this kind of generalization.

\subsection{Multilingual prompting}
\label{sec:multilingualprompting}

\input{promptlangl1.tex}

Since all prompts in xP3 are in English (even for multilingual datasets), we created xP3mt, an extension with machine-translated prompts. To investigate performance on non-English prompts, we additionally human- and machine-translated the English evaluation prompts from Figure~\ref{fig:taskgen}. In Table~\ref{tab:promptlangl1}, we report performance on these. Results on machine-translated prompts in languages that are not part of the finetuning corpus, such as those in Figure \ref{fig:langgen}, are in Appendix \S\ref{sec:multiunseen}. Table~\ref{tab:promptlangl1} shows that BLOOMZ performs much better on English than on non-English prompts. BLOOMZ-MT, which is finetuned on xP3mt, significantly improves on multilingual prompts. On XNLI, BLOOMZ-MT raises the average performance on human-translated prompts from 41.13 to 45.55. This comes at the cost of a reduction in its performance on English prompts, from 53.58 to 49.74. For mT0, the MT version provides similar performance gains on XNLI and XWinograd non-English prompts, while results on XCOPA and XStoryCloze are mixed. Similar to~\citet{lin2021few}, we also find that models perform better on human-translated prompts than machine-translated ones for XNLI. 

\subsection{Scaling}
\label{sec:scaling}

\begin{figure}[htbp]
    \centering
    \includegraphics[width=\linewidth]{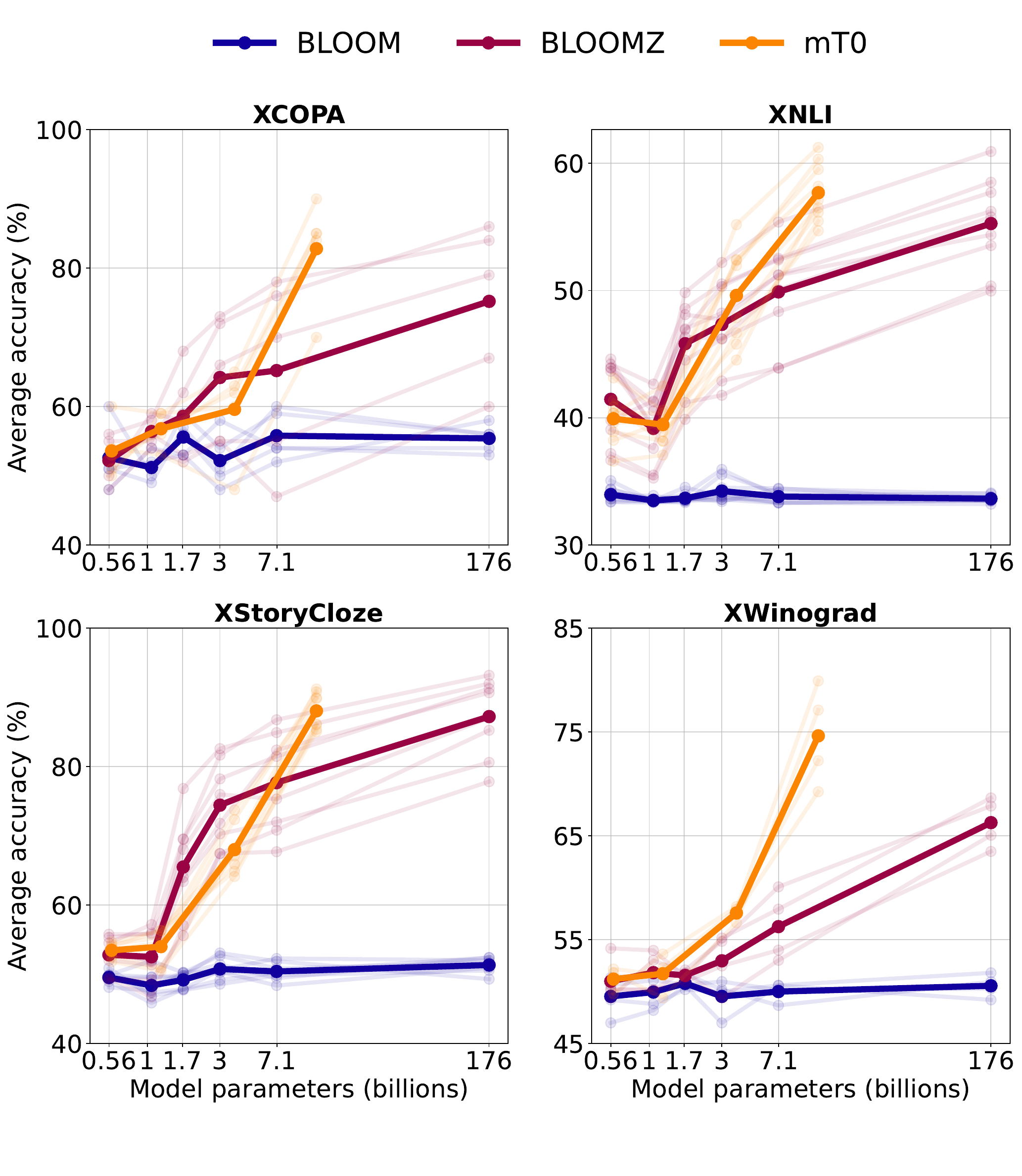}
    \caption{Aggregate performance vs. size. Transparent lines correspond to individual languages, while thick lines are average accuracy scores.}
    \label{fig:scale}
\end{figure}

In Figure \ref{fig:taskgen}, the average performance of BLOOM is near the random baselines of 0.50 for Sentence Completion and Coreference Resolution and 0.33 for NLI. We think this is due to all of our experiments being zero-shot and using untuned prompts~\cite{perez2021true}. We find in Figure~\ref{fig:scale} that even at 560M parameters, multitask finetuning improves zero-shot generalization. The gap between pretrained and multitask finetuned models grows significantly as parameters increase. Scaling up parameters benefits all languages evaluated.

\subsection{Generation tasks}
\label{sec:generation}

\begin{figure}[htbp]
    \centering
        \includegraphics[width=\linewidth]{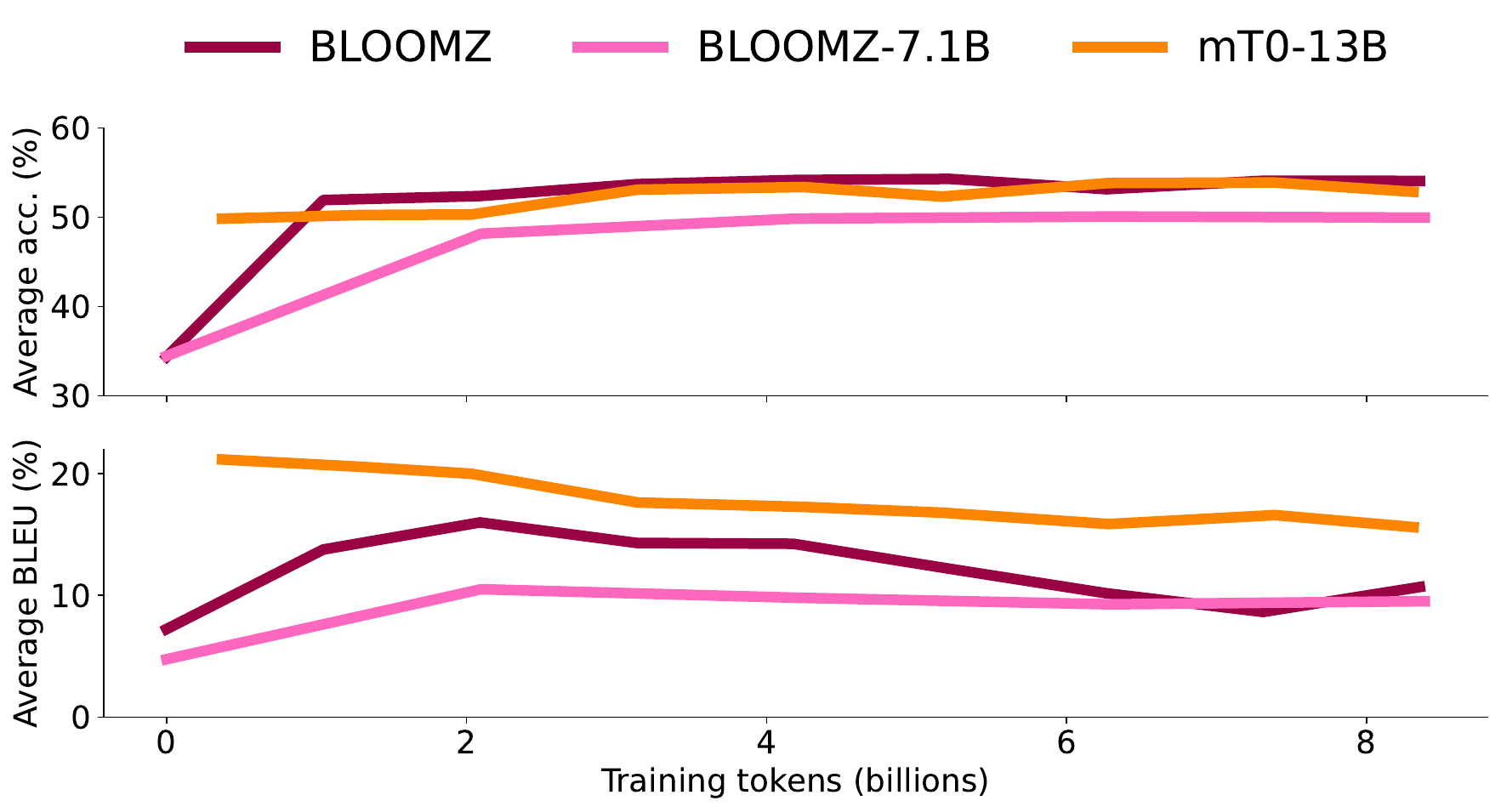}
        \caption{Validation performance during training on natural language understanding (NLU) and natural language generation (NLG) tasks. The former are scored using accuracy and the latter using BLEU \cite{papineni2002bleu}. The NLG tasks measured are translation and summarization. For BLOOMZ(-7.1B) the performance at 0 training tokens corresponds to the performance of BLOOM(-7.1B). For mT0 there is no data point at 0 tokens, as its base model, mT5, is not suitable for evaluation without finetuning. Performance on individual tasks is in Appendix \S\ref{sec:fullresults}.}
        \label{fig:validation}
\end{figure}

In this section, we investigate the impact of multitask finetuning on generative tasks. In Figure \ref{fig:validation}, we plot validation performance throughout the training process. We find that while performance on natural language understanding tasks continues to increase, generative performance jumps initially and then decreases. Relatedly, in Table~\ref{tab:humaneval}, we find that multitask finetuning does not improve performance on HumanEval~\cite{chen2021evaluating}. Only for small models, such as BLOOM-560M vs. BLOOMZ-560M, there are meaningful performance gains. When no code data is included in finetuning (BLOOMZ-P3) performance decreases significantly. mT0 models, which have not been pretrained on code, fail to solve any HumanEval problems (see full results in Appendix~\S\ref{sec:fullresults}). Given a Python docstring, HumanEval requires models to complete a function. Inspecting generations reveals that the multitask finetuned models are biased towards short generations. In Appendix~\S\ref{sec:codegen}, we show example solutions from HumanEval and compute average length statistics. BLOOMZ tries to solve problems with 70\% fewer characters than BLOOM.

\begin{table}[h!]
    \centering
    \footnotesize
    \begin{tabular}{lccc}
    \toprule
    & \multicolumn{3}{c}{Pass@$k$} \\
    & $k=1$ & $k=10$ & $k=100$ \\
    \midrule
    GPT-Neo 1.3B & 4.79\% & 7.47\% & 16.30\% \\
    GPT-Neo 2.7B & 6.41\% & 11.27\% & 21.37\% \\
    GPT-J 6B & 11.62\% & 15.74\% & 27.74\% \\
    GPT-NeoX 20B & 15.4\% & 25.6\% & 41.2\% \\
    \midrule
    Codex-300M & 13.17\% & 20.37\% & 36.27\% \\
    Codex-679M & 16.22\% & 25.7\% & 40.95\% \\
    Codex-2.5B & 21.36\% & 35.42\% & 59.5\% \\
    Codex-12B & 28.81\% & 46.81\% & 72.31\% \\
    \midrule
    BLOOM-560M & 0.82\% & 3.02\% & 5.91\% \\
    BLOOM-1.1B & 2.48\% & 5.93\% & 9.62\% \\
    BLOOM-1.7B & 4.03\% & 7.45\% & 12.75\% \\
    BLOOM-3B & 6.48\% & 11.35\% & 20.43\% \\
    BLOOM-7.1B & 7.73\% & 17.38\% & 29.47\% \\
    BLOOM & 15.52\% & 32.20\% & 55.45\% \\
    \midrule
    BLOOMZ-560M & 2.18 \% & 4.11\% & 9.00\% \\
    BLOOMZ-1.1B & 2.63\% & 6.22\% & 11.68\% \\
    BLOOMZ-1.7B & 4.38\% & 8.73\% & 16.09\% \\
    BLOOMZ-3B & 6.29\% & 11.94\% & 19.06\% \\
    BLOOMZ-7.1B & 8.06\% & 15.03\% & 27.49\% \\
    BLOOMZ & 12.06\% & 26.53\% & 48.44\% \\
    BLOOMZ-P3 & 6.13\% & 11.79\% & 18.73\% \\
    \bottomrule
    \end{tabular}
    \caption{
        Code continuation on HumanEval. Non-BLOOM results come from prior work~\cite{chen2021evaluating,fried2022incoder}. Codex is a language model finetuned on code, while the GPT models~\cite{black2021gpt,wang2021gpt,black2022gpt} are trained on a mix of code and text like BLOOM. Following \citet{chen2021evaluating} we generate 200 samples for each problem with top \emph{p} = 0.95 and compute pass rates. We perform this evaluation three times for temperatures 0.2, 0.6 and 0.8 and pick the best pass rate.
    }
    \label{tab:humaneval}
\end{table}

This bias towards short answers and the performance drop on generative tasks come from finetuning on short texts. Most tasks in our finetuning dataset, xP3, are single sentences. We show in Appendix~\S\ref{sec:generationlength} that finetuning on fewer short tasks via early stopping, adding long tasks or upweighting long tasks leads to longer generations and slightly better performance. We find it most effective, however, to force a minimum generation length at inference. This is done by ignoring any probability mass the model assigns to its end-of-sequence token for a desired number of tokens. Only after the generation has reached the desired length, can the model generate the end-of-sequence token, thus finishing the generation. Forcing a minimum generation length improves the BLEU score on a translation task by 9 points, see Appendix~\S\ref{sec:generationlength} for quantitative and Figure \ref{fig:qgen} for qualitative results.

\subsection{Effect of language proportions}
\label{sec:corpus}

\begin{figure}[t]
    \centering
    \includegraphics[width=\linewidth]{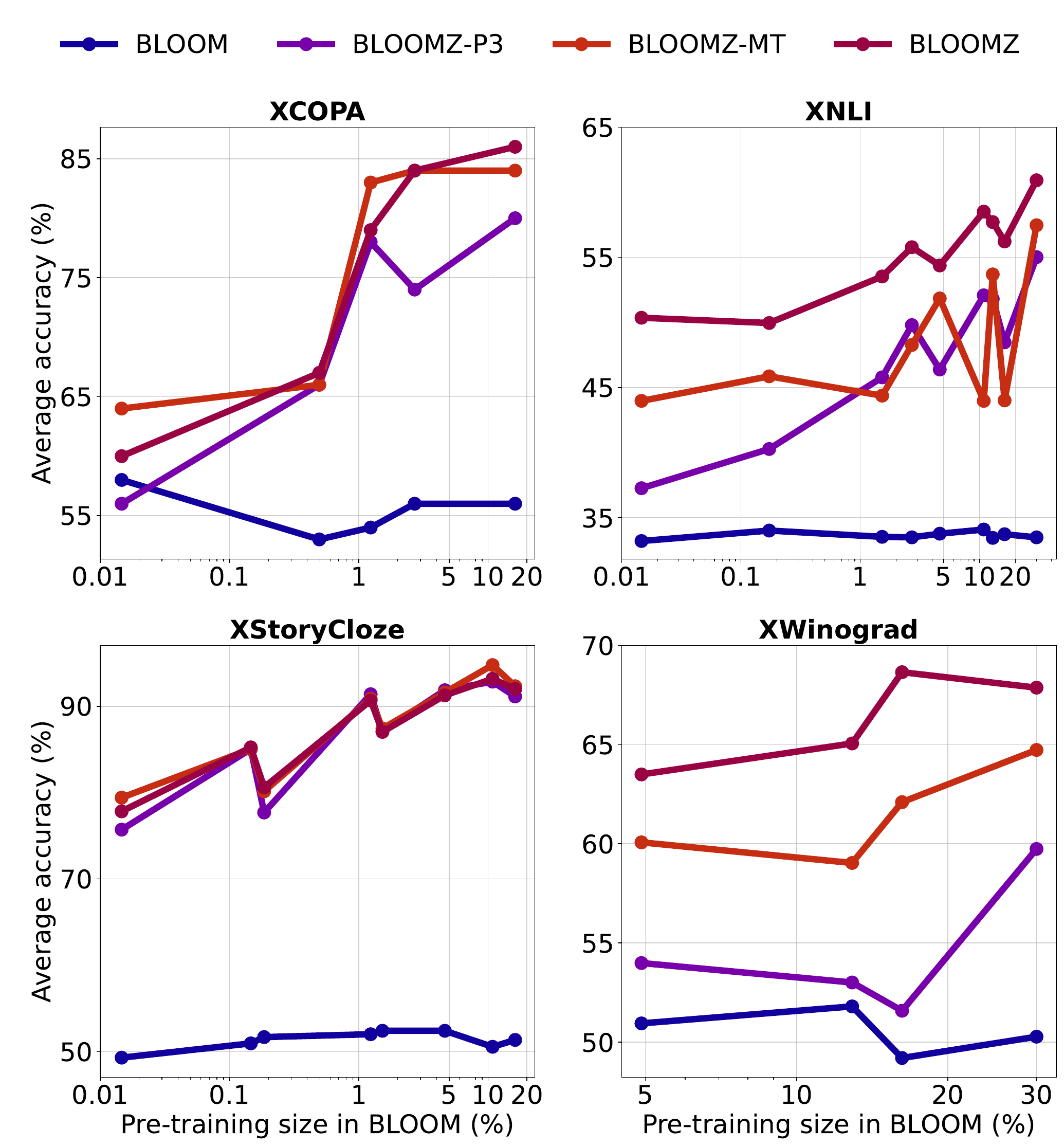}
    \caption{Performance across languages by size in the BLOOM pretraining corpus, ROOTS.}
    \label{fig:langscale}
\end{figure}

In Figure~\ref{fig:langscale}, we find that finetuned BLOOM models perform better on languages seen extensively during pretraining. As the language distribution in the finetuning dataset, xP3, closely follows that of pretraining, these languages are also seen most frequently during finetuning. Specifically, XCOPA and XNLI show significantly better performance on these high-resource languages, such as English, Spanish or French, which all make up more than 10\% of pretraining individually. The trend is less consistent for XWinograd. This may be caused by the fact that XWinograd language subsets are not translations of each other and have a significantly different number of samples. Thus, some language subsets of XWinograd may be inherently more difficult than others.

\section{Conclusion}

In this work we investigated crosslingual multitask finetuning. We developed xP3, a corpus consisting of tasks in 46 languages. Further, we have extended xP3 to xP3mt with machine-translated prompts. We have finetuned pretrained BLOOM and mT5 models on the newly created corpora as well as the English-only P3 corpus to produce BLOOMZ and mT0 models.

We found that English-only finetuning suffices for a multilingual pretrained large language model to generalize to tasks in other pretrained languages. However, finetuning on multiple languages using xP3 provided even better performance. We have further observed finetuned models to be capable of generalization to new tasks in languages they have never intentionally seen. We investigated multilingual prompting and found performance after finetuning on English prompts only to be poor. However, finetuning on a corpus with machine-translated prompts (xP3mt) lead to significantly better performance on human-written non-English prompts. Comparing models from 560 million up to 176 billion parameters revealed that the performance gap between only pretraining and finetuning widens as parameters increase. Lastly, we found multitask finetuning on billions of short targets biases models to produce short answers, which can hurt performance on generative tasks. We proposed a simple workaround by forcing a minimum generation length at inference.

To contribute to future progress on improving zero-shot generalization, we release all datasets and models introduced in this work.

\section{Limitations}

We highlight several limitations of our work:

\paragraph{Unnatural prompting format} The choice to separate inputs and targets using a space character has proven effective to multitask finetune our decoder-only models. Nonetheless, poorly formatted prompts may result in undesirable behavior. For example, given the following prompt: ``Translate to English: Je t'aime", the model may continue the input with additional French content before starting to solve the task, i.e. translating the input from French to English. This can be mitigated by improving the prompts with a trailing full stop or a newline symbol. Encoder-decoder models, such as our mT0, do not suffer from this problem, as inputs and targets are fed into different parts of the model.

\paragraph{Limited languages in xP3} The pretraining corpus of mT0 contains more than 101 languages~\cite{xue2020mt5}, however, we finetune on only 46 languages. Likely, finetuning on the full 101 languages mT0 has seen during pretraining would lead to better performance. However, we decided to use only the languages of BLOOM in order to study language generalization (\S\ref{sec:langgen}). Similarly, one could likely attain better performance by enhancing xP3 with more datasets, such as via BIG-Bench \cite{srivastava2022beyond,suzgun2022challenging}, or more prompts, such as via NL-Augmenter \cite{dhole2021nl}. We have released an extended version of xP3 dubbed xP3x that covers 277 languages and is around ten times larger than xP3, but are yet to finetune models on it.

\paragraph{Performance} While our models show strong capabilities of performing tasks zero-shot, there remain numerous failure modes that are common in large language models~\cite{rae2021scaling,bommasani2021opportunities,zhang2022opt,smith2022using,ouyang2022training,taylor2022galactica,chowdhery2022palm,biderman2023pythia,allal2023santacoder,li2023starcoder}. In Figure \ref{fig:fable} of Appendix \S\ref{sec:gen}, BLOOMZ fails to understand the moral of a fable resulting in an undesirable generation. Similarly, in Figure \ref{fig:qgen}, mT0-13B is asked to provide an explanation, but answers with a question. We have made several modifications to the multitask finetuning recipe, such as loss weighting, mixing in long tasks, and various multilingual aspects, leading to the strong zero-shot performance of our models. However, there are many other changes to the multitask finetuning procedure that are worth exploring to get better models~\cite{honovich2022unnatural,wang2022self,longpre2023flan,liu2023zero,dettmers2023qlora}. Further, the pre-trained models we use, BLOOM and mT5, are suboptimal in many aspects such as compute allocation~\cite{hoffmann2022training,muennighoff2023scaling}, pre-training datasets~\cite{longpre2023pretrainer,touvron2023llama,chung2023unimax}, pre-training objective~\cite{tay2022ul2} and possibly model architecture~\cite{komatsuzaki2022sparse,shen2023flan}. Future work should investigate multitask finetuning better base models.

\paragraph{Learning new languages during finetuning} While we have investigated generalization to languages only seen during pretraining, we did not investigate generalization to languages only seen during finetuning. Our mT0 models are finetuned on several new languages not seen in pretraining (see Figure \ref{fig:xp3langs}). Out of those, we only evaluated on code (HumanEval), where mT0 performed at the random baseline (0.00 in Table \ref{tab:allres}). We point to follow-up work that has investigated the question of teaching BLOOMZ new languages ~\cite{yong2022bloom+,cahyawijaya2023instruct} and work investigating adaptation of BLOOM~\cite{ennen2023extending,yong2022adapting}.

\newpage

\section*{Acknowledgments}

This work was granted access to the HPC resources of Institut du d\'eveloppement et des ressources en informatique scientifique (IDRIS) du Centre national de la recherche scientifique (CNRS) under the allocation 2021-A0101012475 made by Grand \'equipement national de calcul intensif (GENCI). In particular, all the evaluations and data processing ran on the Jean Zay cluster of IDRIS, and we want to thank the IDRIS team for responsive support throughout the project, in particular R\'emi Lacroix.

We thank the XGLM team for providing access to XStoryCloze. We thank volunteers who human-translated XNLI prompts. We thank Noah Constant and Douwe Kiela for feedback on drafts of this paper. We thank Victor Sanh, Stephen Bach, Sasha Rush and Jordan Clive for support throughout the project.


\bibliography{anthology,acl2020,prompt_appendix_generation/multilingual_promptgen.bib}
\bibliographystyle{acl_natbib}

\appendix

\onecolumn

\tableofcontents

\clearpage

\section{Contributions}
\label{sec:contributions}

This research was conducted under the BigScience project for open research, a year-long initiative targeting the study of large models and datasets. The goal of the project is to research language models in a public environment. The project has hundreds of researchers from more than 50 countries and over 250 institutions. The BigScience project was initiated by Thomas Wolf at Hugging Face, and this collaboration would not have been possible without his effort. In the following, we list contributions made to this work. \\

\input{contributions.tex}

\FloatBarrier

\section{Task generalization breakdown}
\label{sec:taskgenlang}

\begin{figure*}[htbp]
\includegraphics[width=\textwidth]{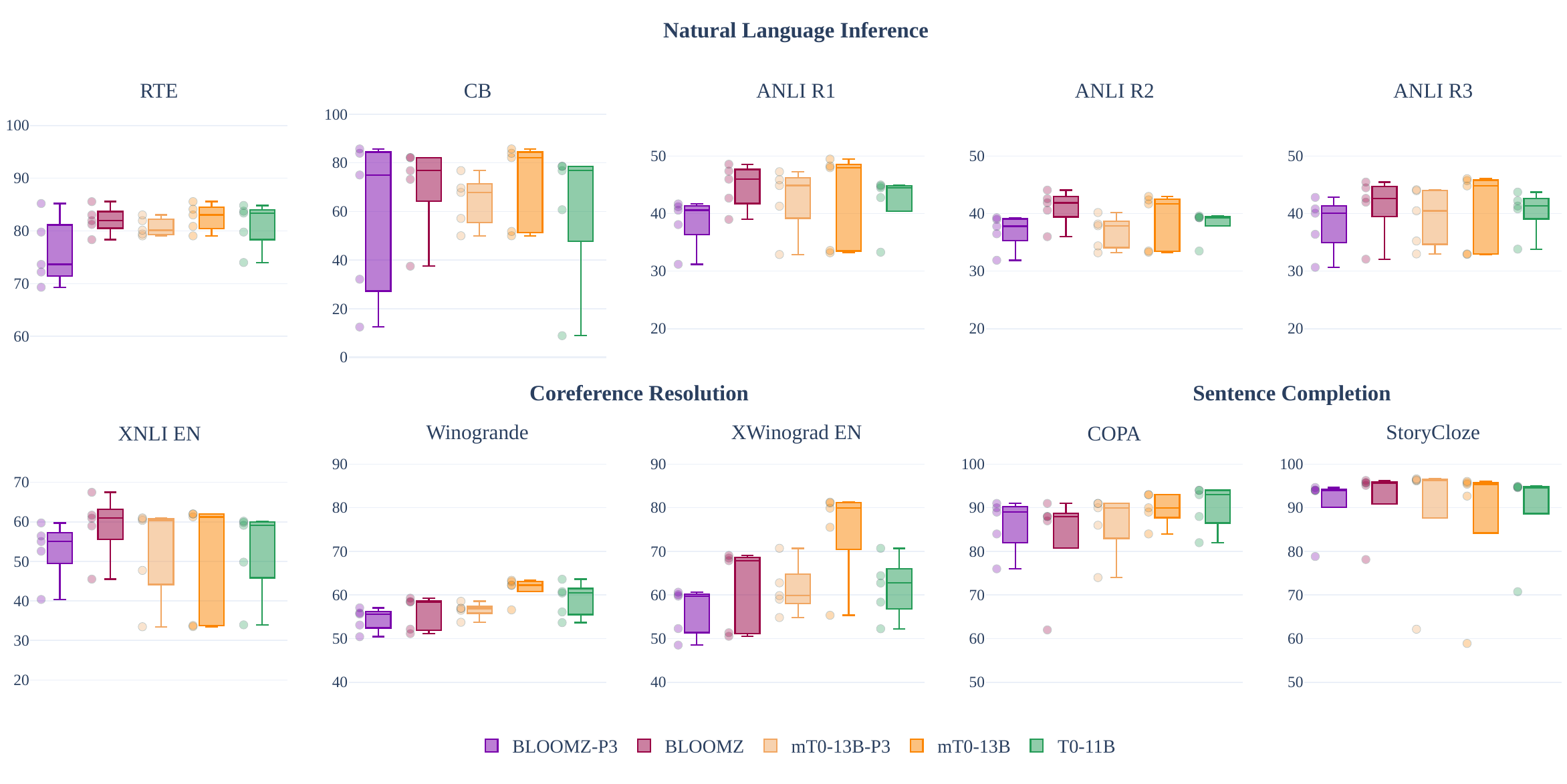}
\centering
\caption{Zero-shot English task generalization. Each dot represents performance on one English evaluation prompt.}
\label{fig:taskgeneng}
\end{figure*}

In Figure~\ref{fig:taskgeneng}, we compare performance on English held-out tasks. We find that \textbf{(a)} finetuning on xP3 outperforms P3 \textbf{(b)} multilingual mT0 is better than monolingual T0 on \emph{English tasks}. We think both improvements come from xP3 having more prompts and datasets than P3 \cite{chung2022scaling}.

\newpage

\begin{figure*}[htbp]
\includegraphics[width=\textwidth]{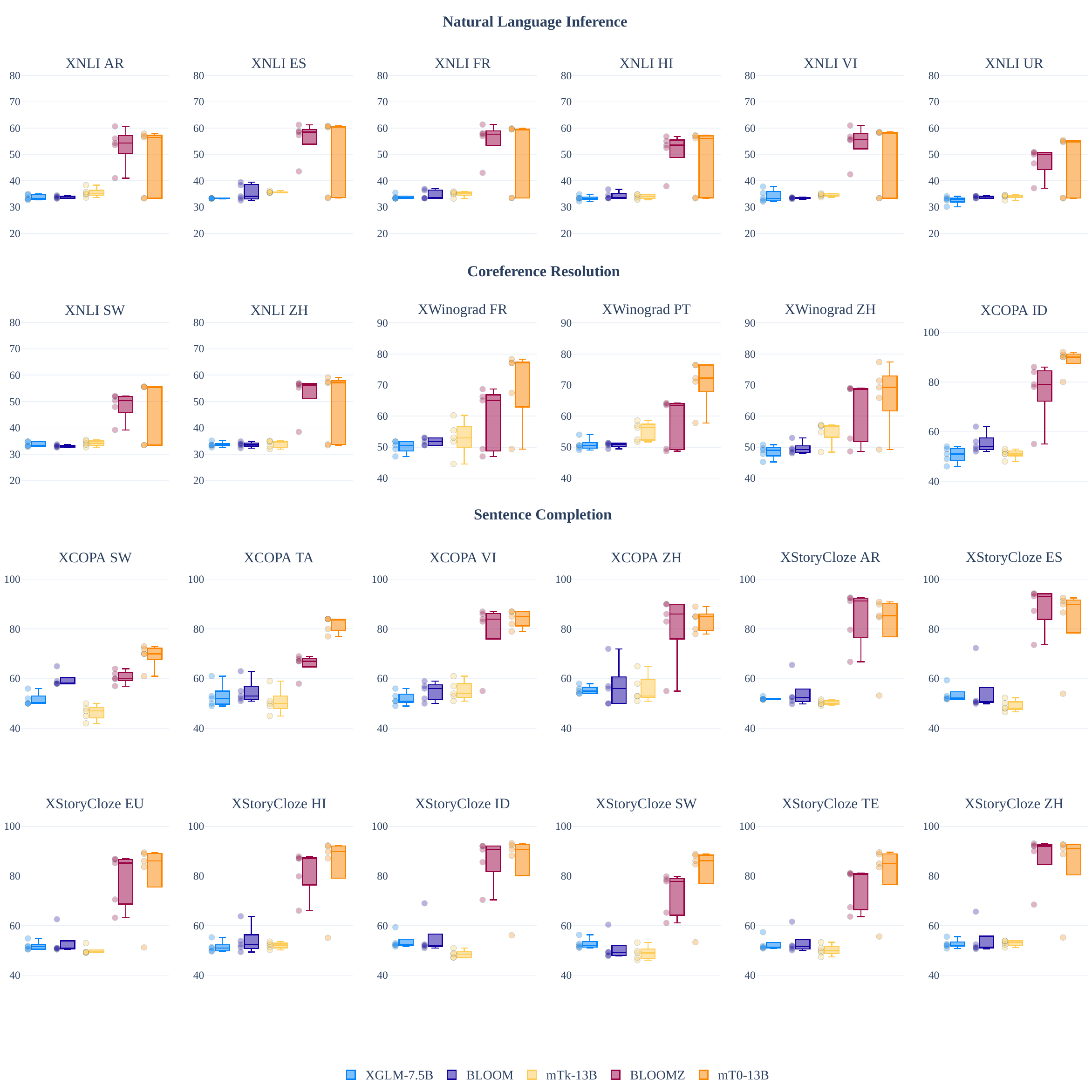}
\centering
\caption{Zero-shot multilingual task generalization on languages seen during pretraining and finetuning. Each dot represents performance on one English evaluation prompt.}
\label{fig:taskgenlang}
\end{figure*}

In Figure \ref{fig:taskgenlang}, we visualize task generalization to multilingual datasets. The same data is aggregated in Figure \ref{fig:taskgen}. Performance by prompt varies substantially highlighting that prompt engineering may still be necessary after MTF. We also find that mT0 consistently outperforms BLOOMZ on Swahili (SW), possibly due to it being a larger part of its pretraining corpus (see Figure \ref{fig:xp3langs} and \S\ref{sec:corpus}).

\FloatBarrier
\newpage

\section{Artifacts}
\label{sec:artifacts}

Table \ref{tab:ckpts} lists all artifacts used or released in this work. We make all our work accessible under the most permissive licenses available to us.

\begin{table*}[htbp]
    \centering
    \tiny
    \resizebox{\textwidth}{!}{\begin{tabular}{@{}l|c|l@{}}
    \toprule
\multicolumn{1}{l|}{\textbf{Artifact}} &
\multicolumn{1}{c|}{\textbf{Explanation}} &\multicolumn{1}{c}{\textbf{Public link}} \\
\midrule
\midrule
ROOTS & Multilingual pretraining corpus of BLOOM & \url{https://huggingface.co/bigscience-data} \\
mC4 & Multilingual pretraining corpus used for mT5 & \url{https://huggingface.co/datasets/mc4} \\
P3 & Multitask finetuning dataset with English data \& English prompts & \url{https://huggingface.co/datasets/bigscience/P3} \\
xP3 & Multitask finetuning dataset with multilingual data \& English prompts & \url{https://huggingface.co/datasets/bigscience/xP3} 
\\
xP3all & Same as xP3 with held-out evaluation sets & \url{https://huggingface.co/datasets/bigscience/xP3all}
\\
xP3mt & Same as xP3 with English \& multilingual machine-translated prompts & \url{https://huggingface.co/datasets/bigscience/xP3mt} 
\\
xP3megds & Processed version of xP3 for easy usage with Megatron-DeepSpeed & \url{https://huggingface.co/datasets/bigscience/xP3megds}
\\
xP3x & Extension of xP3 to 277 languages & \url{https://huggingface.co/datasets/Muennighoff/xP3x}
\\
\midrule
XGLM-7.5B & 7.5B parameter pretrained multilingual transformer & \url{https://huggingface.co/facebook/xglm-7.5B} \\
T0-11B & 11B parameter model finetuned on P3 & \url{https://huggingface.co/bigscience/t0} \\
mTk-Instruct-3.7B & 3.7B parameter multitask finetuned multilingual transformer & \url{https://huggingface.co/allenai/mtk-instruct-3b-def-pos} \\
mTk-Instruct-13B & 13B parameter multitask finetuned multilingual transformer & \url{https://huggingface.co/allenai/mtk-instruct-11b-def-pos} \\
\midrule
BLOOM-560M & 560M parameter model pretrained on ROOTS & \url{https://huggingface.co/bigscience/bloom-560m} \\
BLOOM-1.1B & 1.1B parameter model pretrained on ROOTS & \url{https://huggingface.co/bigscience/bloom-1b1} \\
BLOOM-1.7B & 1.7B parameter model pretrained on ROOTS & \url{https://huggingface.co/bigscience/bloom-1b7} \\
BLOOM-3B & 3B parameter model pretrained on ROOTS & \url{https://huggingface.co/bigscience/bloom-3b} \\
BLOOM-7.1B & 7.1B parameter model pretrained on ROOTS & \url{https://huggingface.co/bigscience/bloom-7b1} \\
BLOOM & 176B parameter model pretrained on ROOTS & \url{https://huggingface.co/bigscience/bloom} \\
\midrule
BLOOMZ-560M & 560M parameter model finetuned on xP3 & \url{https://huggingface.co/bigscience/bloomz-560m} 
\\
BLOOMZ-1.1B & 1.1B parameter model finetuned on xP3 & \url{https://huggingface.co/bigscience/bloomz-1b1} 
\\
BLOOMZ-1.7B & 1.7B parameter model finetuned on xP3 & \url{https://huggingface.co/bigscience/bloomz-1b7} 
\\
BLOOMZ-3B & 3B parameter model finetuned on xP3 & \url{https://huggingface.co/bigscience/bloomz-3b} 
\\
BLOOMZ-7.1B & 7.1B parameter model finetuned on xP3 & \url{https://huggingface.co/bigscience/bloomz-7b1} 
\\
BLOOMZ-7.1B-MT & 7.1B parameter model finetuned on xP3mt & 
\url{https://huggingface.co/bigscience/bloomz-7b1-mt} 
\\
BLOOMZ-7.1B-P3 & 7.1B parameter model finetuned on P3 & \url{https://huggingface.co/bigscience/bloomz-7b1-p3} 
\\
BLOOMZ & 176B parameter model finetuned on xP3 & \url{https://huggingface.co/bigscience/bloomz} 
\\
BLOOMZ-MT & 176B parameter model finetuned on xP3mt & \url{https://huggingface.co/bigscience/bloomz-mt} 
\\
BLOOMZ-P3 & 176B parameter model finetuned on P3 & \url{https://huggingface.co/bigscience/bloomz-p3} 
\\
\midrule
mT5-300M & 300M parameter model pretrained on a sampled version of mC4 & \url{https://huggingface.co/google/mt5-small} \\
mT5-580M & 580M parameter model pretrained on a sampled version of mC4 & \url{https://huggingface.co/google/mt5-base} \\
mT5-1.2B & 1.2B parameter model pretrained on a sampled version of mC4 & \url{https://huggingface.co/google/mt5-large} \\
mT5-3.7B & 3.7B parameter model pretrained on a sampled version of mC4 & \url{https://huggingface.co/google/mt5-xl} \\
mT5-13B & 13B parameter model pretrained on a sampled version of mC4 & \url{https://huggingface.co/google/mt5-xxl} \\
\midrule
mT0-300M & 300M parameter model finetuned on xP3 & \url{https://huggingface.co/bigscience/mt0-small} 
\\
mT0-580M & 580M parameter model finetuned on xP3 & \url{https://huggingface.co/bigscience/mt0-base} 
\\
mT0-1.2B & 1.2B parameter model finetuned on xP3 & \url{https://huggingface.co/bigscience/mt0-large} 
\\
mT0-3.7B & 3.7B parameter model finetuned on xP3 & \url{https://huggingface.co/bigscience/mt0-xl} 
\\
mT0-13B & 13B parameter model finetuned on xP3 & \url{https://huggingface.co/bigscience/mt0-xxl} 
\\
mT0-13B-MT & 13B parameter model finetuned on xP3mt & \url{https://huggingface.co/bigscience/mt0-xxl-mt} 
\\
mT0-13B-P3 & 13B parameter model finetuned on P3 & \url{https://huggingface.co/bigscience/mt0-xxl-p3} 
\\
    \bottomrule
    \end{tabular}}
    \caption{Links to all models \& datasets used as part of this work. BLOOMZ models have an additional repository containing the final optimizer states for training with Megatron-Deepspeed that can be found by appending ``-optimizer-states" to the respective URL. BLOOM(Z) models are released under the RAIL license, while mT5 / mT0 models are licensed under Apache 2.0}
    \label{tab:ckpts}
\end{table*}

\FloatBarrier

\section{ROOTS language contamination}
\label{sec:contamination}

\begin{figure*}[htbp]
    \centering
    \begin{center}
        \includegraphics[width=\textwidth]{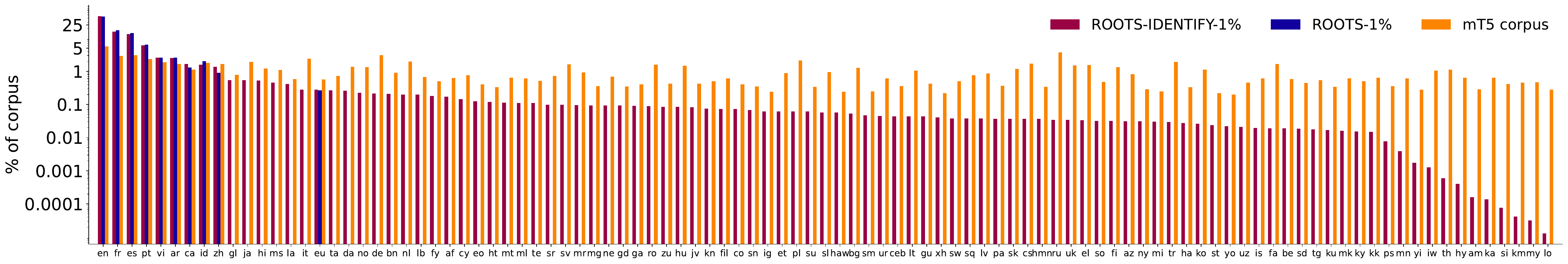}
        \caption{Language composition of ROOTS-IDENTIFY-1\%, ROOTS-1\% and the mT5 corpus. All mT5 languages are depicted. ROOTS-1\% is a random 1\% sample of ROOTS with its assigned meta-languages. ROOTS-IDENTIFY-1\% are the actual languages in ROOTS-1\% re-identified using \texttt{cld3}.}
        \label{fig:roots-identity-langs}
    \end{center}
\end{figure*}

While the BLOOM ROOTS corpus~\cite{laurencconbigscience} was collected from 46 natural languages and 13 programming languages, we find that sentences from the same document do not always belong to the collected (meta) language. Some sentences use languages like Russian or Japanese that were not the intentionally collected parts. This ``language contamination'' may stem from ``code-mixing'' or different languages being used in code comments. To investigate the extent of contamination, we randomly sample 1\% of the documents from ROOTS for a total of 51M documents. For each document, we use \texttt{cld3}\footnote{\url{https://github.com/google/cld3}}~\cite{xue2020mt5} to identify the languages used in each sentence and compare them with the meta language of the document. We summarize our results in Figure~\ref{fig:roots-identity-langs}. It shows that ROOTS contains unintentionally collected languages, such as Burmese (my: 0.00003\%), Thai (th: 0.006\%), Turkish (tr: 0.03\%), Greek (el: 0.03\%), Russian (ru: 0.03\%), Bulgarian (bg: 0.05\%),  Estonian (et: 0.06\%), Haitian (ht: 0.12\%), German (de: 0.21\%), Italian (it: 0.28\%) and Japanese (ja: 0.54\%). These ``unseen'' languages only have small sentence proportions in our subsample compared to English (en: 46.23\%), French (fr: 15.73\%) and Spanish (es: 13.38\%). Yet, they may help the language generalization of BLOOMZ models described in \S\ref{sec:langgen}. Japanese is mostly mixed in the meta English documents (47\%), meta Code documents (8\%) and meta Chinese documents (5\%). Meanwhile, Russian is mostly mixed in the meta English documents (52\%), meta Code documents (19\%) and meta French documents (11\%).

\FloatBarrier

\section{Code generations}
\label{sec:codegen}

\begin{figure*}[htbp]
    \centering
    \subfloat[\centering BLOOM]{{\includegraphics[width=0.45\textwidth]{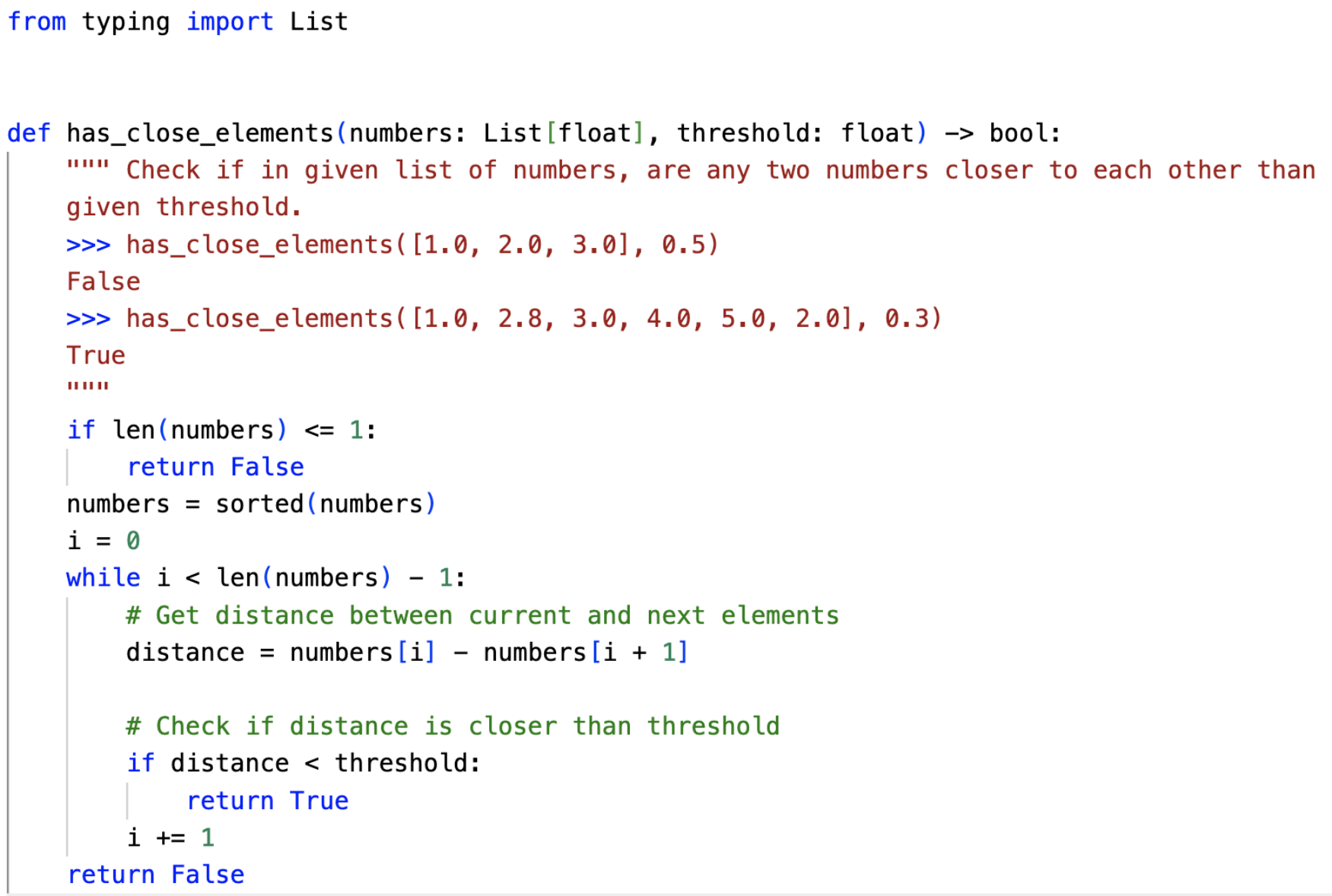}}}
    \qquad
    {\subfloat[\centering BLOOMZ]{\raisebox{0.85\height}{\includegraphics[width=0.45\textwidth]{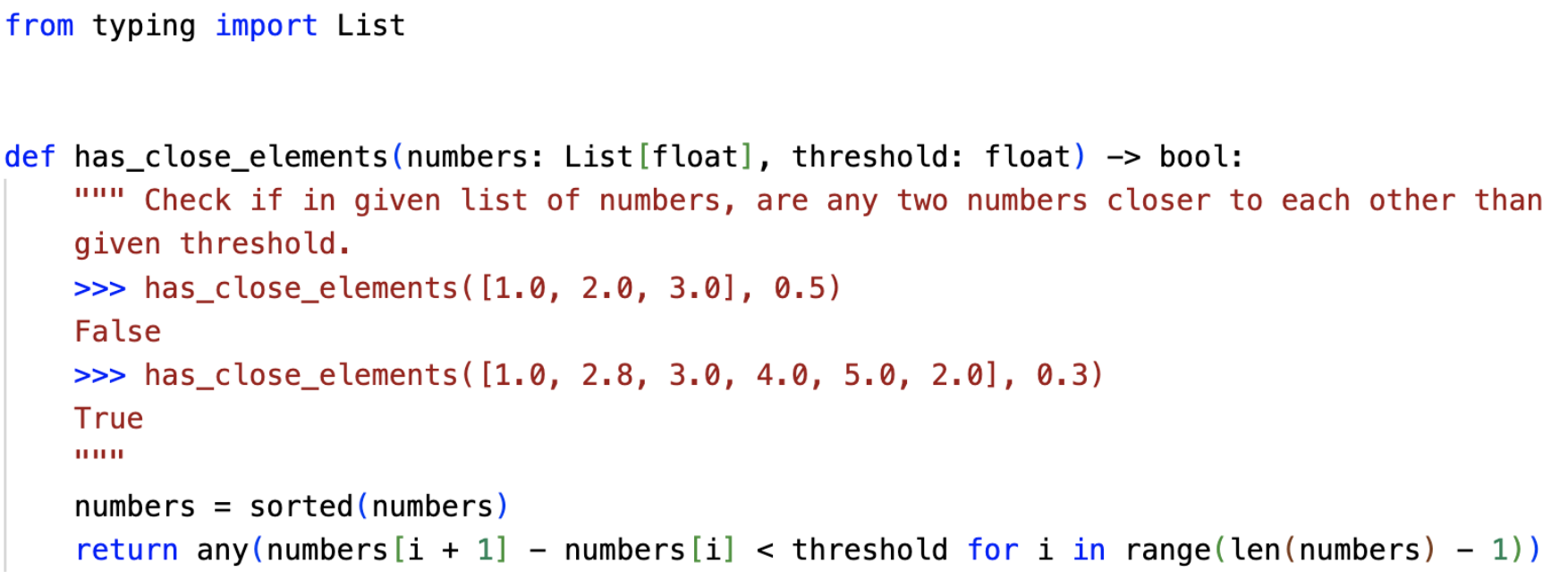}}}}
    \caption{Code generations of BLOOM and BLOOMZ on HumanEval. The model is prompted to generate after the final $"""$. The generation is stopped after an end-of-sequence token or a return statement followed by a newline.}
    \label{fig:codegen}
\end{figure*}

Table \ref{tab:codelen} provides statistics on code generations and code data. We find that BLOOM generates on average 70\% more characters and 17x more comments than BLOOMZ for a given problem from HumanEval. Figure \ref{fig:codegen} compares an example solution from BLOOM and BLOOMZ. While both solutions are correct, BLOOMZ is biased towards short and concise answers.

\begin{table}[htbp]
    \small
    \begin{center}
        \resizebox{0.5\linewidth}{!}{
            \begin{tabular}{l|cc|c}
\toprule
Data ($\rightarrow$) & \multicolumn{2}{c|}{HumanEval generations} & Fine-tuning data \\
& BLOOM & BLOOMZ & in xP3 (code data) \\
\midrule
Average characters & 247 & 144 & 531 \\
Average Python comments (\#) & 0.69 & 0.04 & 0.85 \\
\bottomrule
            \end{tabular}
        }
    \caption{Number of characters and comments for generations and fine-tuning data. For finetuning data, the statistics are computed for the targets that the model is tasked to generate, not the input.}
    \label{tab:codelen}
    \end{center}
\end{table}

\FloatBarrier

\section{Qualitative examples}
\label{sec:gen}

\begin{figure*}[htbp]
    \centering
    \subfloat[\centering English prompt]{{\includegraphics[width=0.45\textwidth]{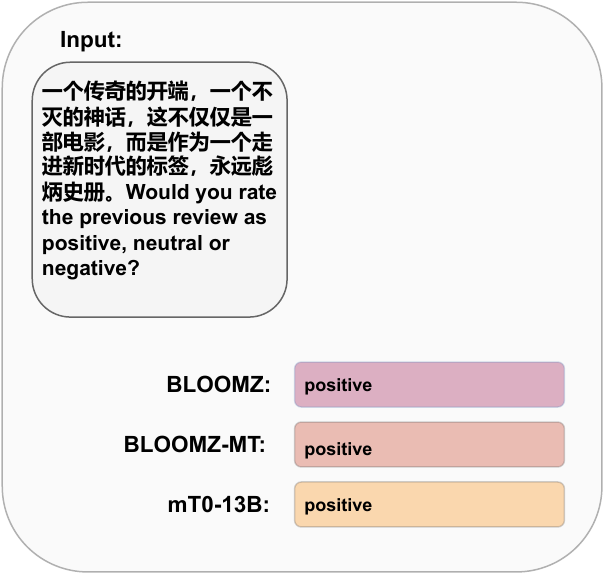}}}
    \qquad
    {\subfloat[\centering Non-English prompt]{{\includegraphics[width=0.45\textwidth]{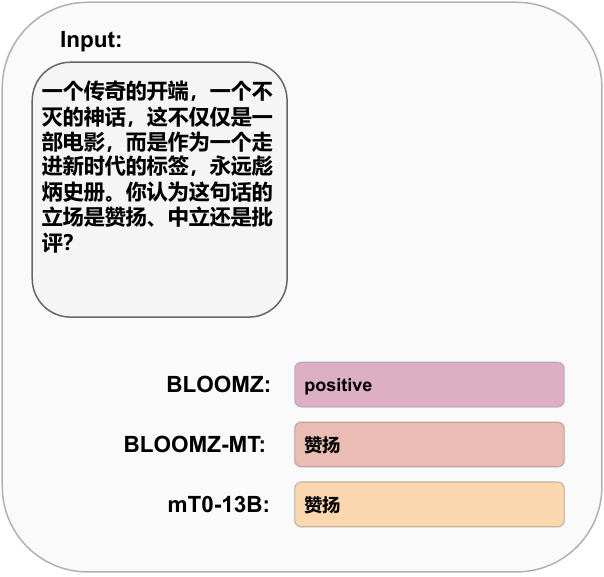}}}}
    \caption{Greedy generations for sentiment analysis, a task trained on. BLOOMZ and mT0-13B have not been trained on non-English prompts, but are still able to handle them. BLOOMZ, however, answers in English. The review is a five star review of Star Wars Episode IV.}
    \label{fig:sentgen}
\end{figure*}

\begin{figure*}[htbp]
    \centering
    \subfloat[\centering English prompt]{{\includegraphics[width=0.45\textwidth]{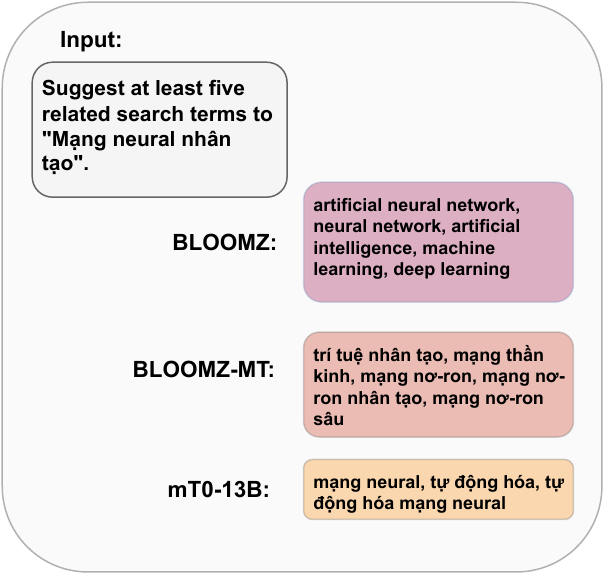}}}
    \qquad
    {\subfloat[\centering Non-English prompt]{{\includegraphics[width=0.45\textwidth]{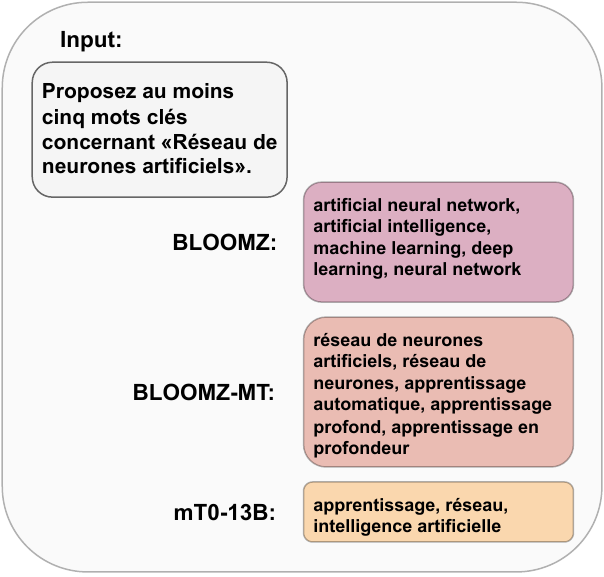}}}}
    \caption{Greedy generations for zero-shot query expansion, a task not trained on. The models sometimes fail to output at least five terms as requested in the prompt.}
    \label{fig:querygen}
\end{figure*}

\begin{figure*}[htbp]
    \centering
    \subfloat[\centering English prompt]{{\includegraphics[width=0.45\textwidth]{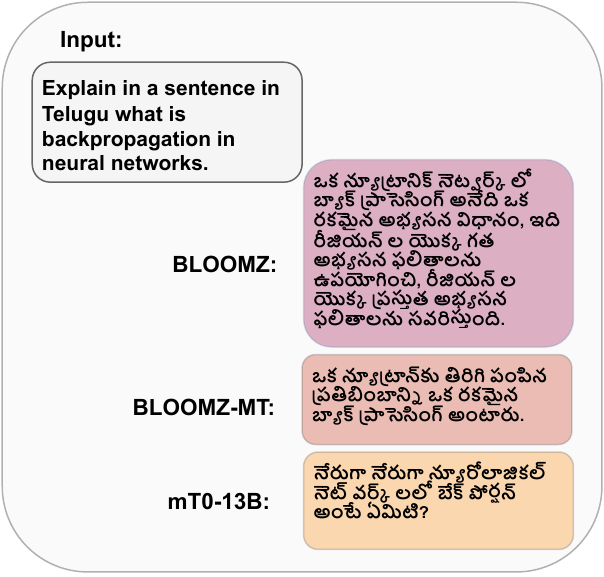}}}
    \qquad
    {\subfloat[\centering English prompt]{{\includegraphics[width=0.45\textwidth]{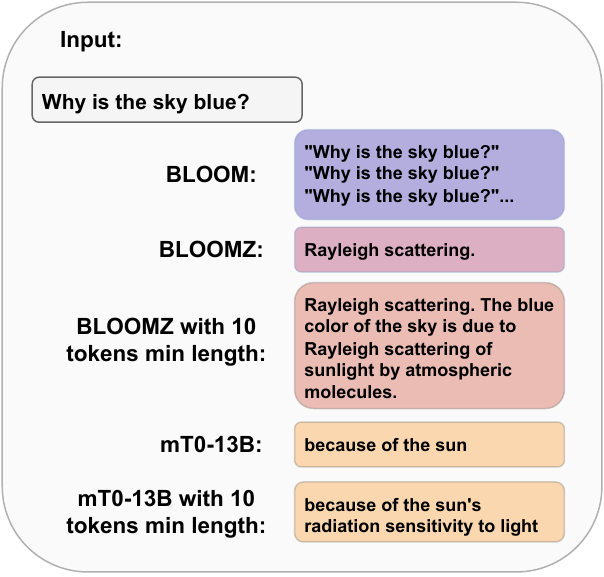}}}}
    \caption{Greedy generations on question answering, a task trained on. \textbf{Left:} Specifying the language in the prompt is an effective way to force the output language. \textbf{Right:} Setting a minimum token length as a generation hyperparameter is an effective way to force long generations. The output of BLOOM is shortened~(marked with $...$).}
    \label{fig:qgen}
\end{figure*}

\begin{figure*}[htbp]
    \centering
    \subfloat[\centering English prompt]{{\includegraphics[width=0.45\textwidth]{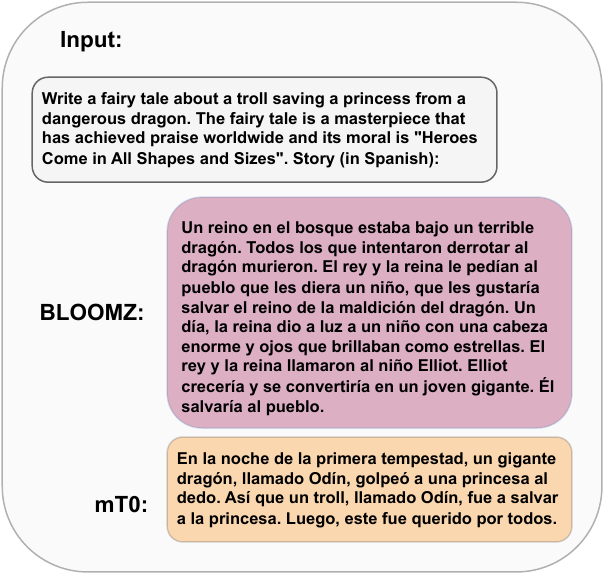}}}
    \qquad
    {\subfloat[\centering English prompt]{{\includegraphics[width=0.45\textwidth]{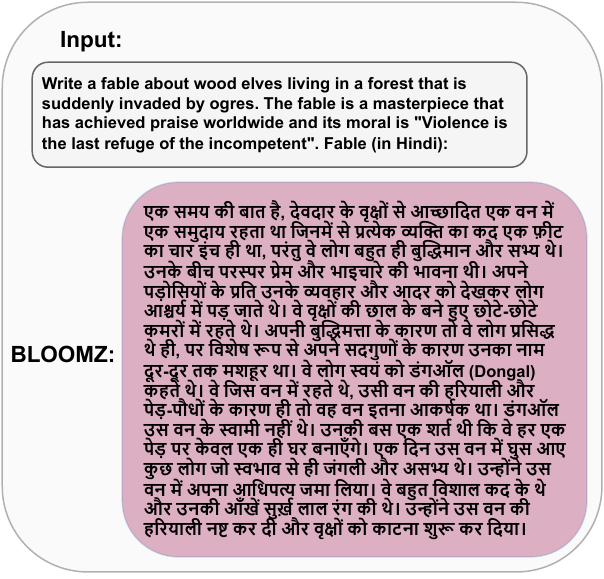}}}}
    \caption{Non-greedy fable generations given a moral, a task not trained on. The generations are cherry-picked from 16 outputs with no minimum length, a temperature of 0.9 and top $k$ of 40. \textbf{Left:} BLOOMZ generates an interesting fable with the desired moral. mT0 is significantly worse at writing stories likely due to its different pretraining objective. \textbf{Right:} BLOOMZ does not seem to understand the moral correctly.}
    \label{fig:fable}
\end{figure*}

\FloatBarrier

\section{Increasing generation length}
\label{sec:generationlength}

In \S\ref{sec:generation}, we found performance on generative tasks to worsen in later stages of training. To investigate this problem further, we study a 7.1 billion parameter BLOOM model that is finetuned for 13 billion tokens, which results in a low BLEU score of 0 and very short generations as shown in Table \ref{tab:genlength} (Default). We can solve this problem with two high-level strategies: \textbf{(a)} Reducing short tasks during finetuning and \textbf{(b)} Forcing a minimum generation length. 

For \textbf{(a)}, we do so by either early stopping, upweighting long tasks or adding new long tasks. As the majority of our finetuning data are single sentences, early stopping has the effect of finetuning on fewer short sentences. Upweighting long tasks is done by removing the loss normalization explained in \S\ref{sec:models}. This has the effect of each token getting equal weight regardless of the task, which upweights long tasks, as they have more tokens. Finally, for adding long tasks, we add tasks that require multi-sentence generations, such as generating an entire news article given a title. These long tasks collectively make up 10\% of finetuning data for this ablation. All three solutions result in longer average generations as shown in Table \ref{tab:genlength} and slightly better BLEU scores, albeit effects are still small.

For \textbf{(b)}, we force the model to generate a minimum number of tokens at inference. Our benchmarking task, MultiEURLEX~\cite{chalkidis2021multieurlex}, requires multi-sentence generations with an average target length of 1965 characters (about 491 tokens). By forcing the model to generate at least 768 tokens, we ensure that the generation is at least as long as the target. This boosts the BLEU score significantly to 9.05. This approach is thus an effective strategy to maintain long generations of good quality.

For our final models, we employ early stopping, adding of long tasks and recommend forcing a minimum generation length at inference for long generations. We do not upweight longer tasks, as it worsens accuracy on our NLU validation tasks by 10\%. The number of tokens our final models are fine-tuned for are displayed in Table~\ref{tab:earlystopping}.

\begin{table}[h]
    \small
    \begin{center}
        \resizebox{1\linewidth}{!}{
            \begin{tabular}{l|ccc}
\toprule
Model & Finetuning tokens & BLEU Score & Average generation length (characters) \\
\midrule
Default & 13 billion & 0.00 & 122 \\
\midrule
Early stopping & 6 billion & 0.00 & 155 \\
Upweight longer tasks & 13 billion & 0.06 & 364 \\
Add more long tasks & 13 billion & 0.06 & 136 \\
\midrule
Forcing 768 tokens at inference & 13 billion & 9.05 & 3072 \\
\bottomrule
            \end{tabular}
        }
    \caption{7.1 billion parameter BLOOMZ models with various modifications benchmarked on MultiEURLEX English-French translation~\cite{chalkidis2021multieurlex}. We benchmark three prompts on both English to French and French to English translation. We then take the median performance across the three prompts for each translation direction and average the two scores to arrive at the BLEU score reported.}
    \label{tab:genlength}
    \end{center}
\end{table}


\begin{table}[h]
    \small
    \begin{center}
            \begin{tabular}{l|cccccccccc}
\toprule
Model & mT0-300M & mT0-560M & mT0-1.2B & mT0-3.7B & mT0-13B \\
\midrule
Tokens & 4.62 & 4.62 & 4.62 & 1.85 & 1.29 \\
\midrule
Model & BLOOMZ-560M & BLOOMZ-1.1B & BLOOMZ-1.7B & BLOOMZ-3B & BLOOMZ-7.1B & BLOOMZ \\
\midrule
Tokens & 3.67 & 0.502 & 8.39 & 8.39 & 4.19 & 2.09 \\
\bottomrule
            \end{tabular}
    \caption{Tokens in billions that final models are finetuned for. We early-stop models based on validation performance. For -MT and -P3 variants we take the checkpoint after the same number of steps as for their default versions.}
    \label{tab:earlystopping}
    \end{center}
\end{table}

\FloatBarrier

\section{XNLI edit distances}
\label{sec:levenshtein}

\begin{table}[h]
    \small
    \begin{center}
        \resizebox{\linewidth}{!}{
            \begin{tabular}{lccc}
\toprule
Premise & Hypothesis & Lev. distance &
Label \\
\midrule
probably so probably so um-hum & probably yes so uh-huh & 13 & Entailment \\
equivalent to increasing national saving to 19 . & National savings are 18 now . & 34 & Neutral \\
The Inglethorps did not appear . & The Inglethorps were the first ones to turn up . & 26 & Contradiction \\
\bottomrule
            \end{tabular}
        }
    \caption{Three samples from the English XNLI split. To solve XNLI models need to classify whether the premise entails, is neutral to or contradicts the hypothesis. Samples are cherry-picked.}
    \label{tab:xnliex}
    \end{center}
\end{table}

As models are surprisingly capable of solving XNLI in languages they were never intentionally trained on (\S\ref{sec:langgen}), we investigate whether XNLI can be solved without any language understanding. To do so, we compute edit distances using the Levenshtein methodology~\cite{levenshtein1966binary} between premise and hypothesis. Table~\ref{tab:xnliex} shows three samples from the English XNLI and their edit distances. Our hypothesis is that entailment pairs generally need to cover similar content, and thus have similar distance. Contradiction pairs still need to cover similar content but differ in at least one major way. Meanwhile for neutral pairs, hypothesis and premise may be about completely different topics, hence they should have the highest distance. In Table~\ref{tab:levenshtein} we compute distances across all Thai, Turkish and Greek samples, three languages where we found language generalization to occur for BLOOMZ. Results confirm our hypothesis that distances are generally largest for neutral samples and smallest for entailment samples. However, the aggregate differences are very small with only a few edits difference. For example, Thai contradiction samples only have 2.5 edits more on average than entailment samples. Thus, comparing characters based on edit distance alone is likely not sufficient to fully explain the language generalization of models in \S\ref{sec:langgen}.

\begin{table}[h]
    \small
    \begin{center}
        \resizebox{0.5\linewidth}{!}{
            \begin{tabular}{lccc}
\toprule
Label ($\rightarrow$) & Entailment & Neutral &
Contradiction \\
Language ($\downarrow$) & & & \\
\midrule
Thai (th) & 79.08 & 82.64 & 81.52 \\
Turkish (tr) & 76.93 & 80.59 & 80.24 \\
Greek (el) & 90.90 & 95.10 & 93.93 \\
\bottomrule
            \end{tabular}
        }
    \caption{Levenshtein distances between hypothesis and premise averaged across samples from different XNLI labels. Each label has 830 samples per language subset.}
    \label{tab:levenshtein}
    \end{center}
\end{table}

\FloatBarrier

\section{Multilingual prompting in unseen languages}
\label{sec:multiunseen}

Table \ref{tab:promptlangl2} shows aggregate performances on languages not intentionally seen during pretraining nor finetuning for BLOOMZ and only seen during pretraining for mT0. For BLOOMZ, performance drops significantly when translating the prompts to the respective unseen languages. Unlike on translated prompts for seen languages (\S\ref{sec:multilingualprompting}), BLOOMZ-MT performs worse than BLOOMZ for machine-translated prompts in unseen languages. This is likely because BLOOMZ-MT has not been finetuned on prompts in these languages. For mT0 differences are less significant.

\input{promptlangl2.tex}

\FloatBarrier

\section{Ideas that did not work}
\label{sec:failures}

We list several experiments that did not improve over baseline results:

\paragraph{Non-causal} In a non-causal or prefix language model, the model attends bidirectionally over input tokens and only causally over target tokens. Given a pretrained causal decoder, other work found that multitask finetuning in a non-causal setup performed better than causal finetuning \cite{wang2022language,tay2022transcending}. However, in our experiments, non-causal finetuning did not improve over causal finetuning.

\paragraph{Special tokens} Instead of separating inputs and targets with a space, we experimented with special tokens. Using the end-of-sequence token as a separator or a completely new token that the model would learn during finetuning significantly worsened results. The models may need to train on more tokens, possibly even during pretraining, to learn these new special tokens \cite{zeng2022glm}.

\paragraph{Fixing prompts} PromptSource has been written with encoder-decoder models in mind, where inputs and targets are fed into different models. As a consequence, human-written prompts in PromptSource often lack separators between input and target. For our decoder models, we decided to separate them with a space. We additionally experimented with leaving them as is or rewriting a significant amount of prompts, but neither improved significantly over space separation.

\paragraph{BitFit} Previous work has shown bias-only finetuning~\cite{zaken2021bitfit} of large language models to be sufficient for strong downstream performance~\cite{logan2021cutting,hu2021lora,muennighoff2022sgpt,liu2022few,ding2022delta,muennighoff2022mteb}. We found multitask finetuning of only biases to perform 15 absolute percentage points worse on the average of held-out tasks for BLOOMZ-7.1B.

\FloatBarrier

\FloatBarrier

\section{Full results}
\label{sec:fullresults}

Table~\ref{tab:allres} shows all evaluation results on test datasets. Table~\ref{tab:valres} displays evaluation results on validation datasets which we use for checkpoint selection.



\eject \pdfpagewidth=77cm \pdfpageheight=80cm

\input{allres.tex}

\clearpage

\eject \pdfpagewidth=50cm \pdfpageheight=40cm

\input{valres.tex}

\clearpage
\eject \pdfpagewidth=21cm \pdfpageheight=29.7cm

\FloatBarrier

\section{Version control}

\textbf{V1 → V2:}
\begin{itemize}
    \item Added evaluation results for the validation datasets used for checkpoint selection (Appendix~\S\ref{sec:fullresults})
    \item Added a section on the effect on generation length (Appendix~\S\ref{sec:generationlength}) and rewrote parts of \S\ref{sec:generation}
    \item Added a mention of xP3x, the extension of xP3 to 277 languages in Appendix~\S\ref{sec:artifacts}
    \item Added an example of XNLI to Appendix~\S\ref{sec:levenshtein}
\end{itemize}

\FloatBarrier

\section{Prompts used}
\label{sec:prompts}
This section describes the prompts used for training and evaluation.
\clearpage
\input{prompt_appendix_nocite.tex}



\section*{Supplementary material}

In the following, dataset naming conventions follow those used in the Hugging Face datasets library. Since xP3 expands upon the P3 dataset employed by Sanh et al. (2022), we refer the reader to that work for example prompts from datasets that fall within P3. Here, we provide prompts curated for datasets that belong to xP3 but not to P3. The prompts provided are not exhaustive. Code will be released to provide a canonical reference. For each dataset considered, a dataset example is provided for additional context. Next, it is noted if the prompt does not match the original task formulation of the dataset. This is followed by a reference for the data, an input template and a target template. For prompts with predefined answer choices, these are also included. To provide examples of both human-translated and machine-translated prompts, samples of each kind are included for the XNLI ES dataset.

## Contents

# 1 Prompts

## 1.1 Simplification

### 1.1.1 GEM/BiSECT en

Dataset from Kim et al. (2021). Used in training.

**Data Example**

| Key | Value |
| --- | --- |
| gem_id | BiSECT_en-train-1 |
| source | To view any of the video clips belo... |
| target | If you want to watch one of the vid... |
| references | If you want to watch one of the vid... |

**Prompts**

Input Template:

```
Split and simplify the following sentence while retaining its full meaning:
{{source}}
 Simplified version:
```

Target Template:

```
{{target}}
```

Input Template:

```
{{source}}
The above sentence is very complicated. Please provide me a simplified synonymous
version consisting of multiple sentences:
```

Target Template:

```
{{target}}
```

Input Template:

```
{{source}}. This sentence is hard to understand. A simpler version with equivalent
meaning is the following:
```

Target Template:

```
{{target}}
```

### 1.1.2 GEM/BiSECT es

**Data Example**

| Key | Value |
| --- | --- |
| gem_id | BiSECT_es-train-1 |
| source | Al final de la Santa Misa , mientra... |
| target | Al finalizar la santa misa , mientr... |
| references | Al finalizar la santa misa , mientr... |

**Prompts**

Input Template:

```
{{source}}. Esta frase es difícil de entender. Una versión más simple con
significado equivalente es la siguiente:
```

Target Template:

```
{{target}}
```

Input Template:

```
Divida y simplifique la siguiente oración conservando su significado completo:
{{source}}
 Versión simplificada:
```

Target Template:

```
{{target}}
```

Input Template:

```
{{source}}
La frase anterior es muy complicada. Por favor, proporcione una versión sinónima
simplificada que consta de varias oraciones:
```

Target Template:

```
{{target}}
```

### 1.1.3 GEM/BiSECT fr

**Data Example**

**Prompts**

Input Template:

| Key | Value |
| --- | --- |
| gem_id | BiSECT_fr-train-1 |
| source | N'ayez pas peur de poser des questi... |
| target | Il ne faut pas avoir peur de poser ... |
| references | Il ne faut pas avoir peur de poser ... |

```
Divisez et simplifiez la phrase suivante tout en conservant son sens complet :
{{source}}
 Version simplifiée :
```

Target Template:

```
{{target}}
```

Input Template:

```
{{source}}
La phrase ci-dessus est très compliquée. Veuillez me fournir une version synonyme
simplifiée composée de plusieurs phrases :
```

Target Template:

```
{{target}}
```

Input Template:

```
{{source}}. Cette phrase est difficile à comprendre. Une version plus simple avec
une signification équivalente est la suivante :
```

Target Template:

```
{{target}}
```

## 1.2 Summarization

### 1.2.1 GEM/wiki_lingua en

Dataset from lad (2020). Used in training.

**Data Example**

**Prompts**

**Notes:** xsum DOC_write_summary_of_above template

| Key | Value |
| --- | --- |
| gem_id | `wikilingua_multilingual-train-42437...` |
| gem_parent_id | `wikilingua_multilingual-train-42437...` |
| source_language | `en` |
| target_language | `en` |
| source | `Go online and simply search "Decor ...` |
| target | `Take a quiz online to find your sty...` |
| references | `Take a quiz online to find your sty...` |

Input Template:

```
{{source}}

===

Write a summary of the text above in English :
```

Target Template:

```
{{target}}
```

---

**Notes:** xsum 'article_DOC_summary' template

Input Template:

```
Article in English: {{source}}

Summary in English:
```

Target Template:

```
{{target}}
```

---

**Notes:** xsum 'DOC_how_would_you_rephrase_few_words' template

Input Template:

```
{{source}}

How would you rephrase that briefly in English?
```

Target Template:

```
{{target}}
```

---

**Notes:** xsum 'DOC_tldr' template

Input Template:

```
{{source}}

TL;DR in English:
```

Target Template:

```
{{target}}
```

---

**Notes:** xsum 'read_below_DOC_write_abstract' template

Input Template:

```
First, read the English article below.

{{source}}

Now, please write a short abstract for it in English.
```

Target Template:

```
{{target}}
```

---

Input Template:

```
{{target}}

Given the above abstract, write an English article for it.
```

Target Template:

```
{{source}}
```

---

Input Template:

```
{{target}}

I'm interested in that, but I only have a few mins. Can you give me the first 500
characters of an article about that?
```

Target Template:

```
{{source[:500]}}
```

---

### 1.2.2 GEM/WIKI_LINGUA ES

**Data Example**

| Key | Value |
| --- | --- |
| gem_id | `wikilingua_multilingual-train-34808...` |
| gem_parent_id | `wikilingua_multilingual-train-34808...` |
| source_language | `es` |
| target_language | `es` |
| source | `Navega en la web y simplemente busc...` |
| target | `Haz un cuestionario en línea para e...` |
| references | `Haz un cuestionario en línea para e...` |

**Prompts**

**Notes:** xsum templates

Input Template:

```
{{source}}

===

Write a summary of the text above in Spanish:
```

Target Template:

```
{{target}}
```

**Notes:** xsum templates

Input Template:

```
First, read the Spanish article below.

{{source}}

Now, please write a short abstract for it in Spanish.
```

Target Template:

```
{{target}}
```

**Notes:** xsum templates

Input Template:

```
{{source}}

TL;DR in Spanish:
```

Target Template:

```
{{target}}
```

**Notes:** xsum templates

Input Template:

```
Article in Spanish: {{source}}

Summary in Spanish:
```

Target Template:

```
{{target}}
```

**Notes:** xsum templates

Input Template:

```
{{source}}
How would you rephrase that briefly in Spanish?
```

Target Template:

```
{{target}}
```

### 1.2.3 GEM/XLSUM BENGALI

Dataset from Hasan et al. (2021). Used in training.

**Data Example**

| Key | Value |
|---|---|
| gem_id | `xlsum_bengali-train-2` |
| url | `https://www.bbc.com/bengali/news-50...` |
| title | রাশিয়ায় ক্ষমতার ২০ বছর যেভাবে কেট... |
| target | ভ্লাদিমির পুতিন তাঁর ক্ষমতায় থাকার... |
| references | ভ্লাদিমির পুতিন তাঁর ক্ষমতায় থাকার... |
| text | গত ২০ বছরে তিনি রাশিয়ার প্রেসিডেন্... |

**Prompts**

Input Template:

একটি নিবন্ধের নীচের শিরোনাম এবং সারাংশ দেওয়া, একটি ছোট নিবন্ধ তৈরি করুন বা তাদের সাথে যেতে একটি দীর্ঘ নিবন্ধের শুরু করুন। শিরোনাম: {{title}}সারাংশ: {{target}}

Target Template:

`{{text[:500]}}`

Input Template:

বিষয়বস্তু: `{{text[:7000]}}`

Target Template:

`{{target}}`

Input Template:

ডক সংক্ষিপ্ত করার জন্য: `{{text[:8500]}}`

Target Template:

`{{target}}`

Input Template:

...`{{text[3000:3500]}}`

Target Template:

`{{text[5000:]}}`

Input Template:

শিরোনাম: `{{title}}`

Target Template:

`{{text[:7000]}}`

Input Template:

`{{text}}`

Target Template:

`{{title}}`

Input Template:

```
শিরোনাম: {{title}}
```

Target Template:

```
{{text[:7000]}}
```

Input Template:

```
{{title}}{{text[:5000]}}
```

Target Template:

```
{{target}}
```

Input Template:

```
{{text[:1000]}}
```

Target Template:

```
{{text[1000:5000]}}
```

### 1.2.4 GEM/xlsum english

**Data Example**

| Key | Value |
|---|---|
| gem_id | xlsum_english-train-2 |
| url | https://www.bbc.com/news/uk-scotlan... |
| title | Huge tidal turbine installed at Ork... |
| target | The massive tidal turbine AK1000 ha... |
| references | The massive tidal turbine AK1000 ha... |
| text | Atlantis Resources unveiled the mar... |

**Prompts**

Input Template:

```
Doc to summarize: {{text[:8500]}}\nSummary in the same language as the doc:
```

Target Template:

```
{{target}}
```

Input Template:

```
Content: {{text[:7000]}}\nThe previous content can be summarized as follows:
```

Target Template:

```
{{target}}
```

Input Template:

```
{{title}}\n{{text[:5000]}}\n\ntl;dr:
```

Target Template:

```
{{target}}
```

Input Template:

```
{{text}} \n\nGive me a good title for the article above.
```

Target Template:

```
{{title}}
```

Input Template:

```
Given the below title and summary of an article, generate a short article or the beginning of a long article to go along with them. Title: {{title}}\nSummary: {{target}}\nArticle (Max 500 characters):
```

Target Template:

```
{{text[:500]}}
```

Input Template:

```
Title: {{title}}\nGiven the above title of an imaginary article, imagine the article.\n
```

Target Template:

```
{{text[:7000]}}
```

Input Template:

```
Title: {{title}}\nGiven the above title of an imaginary article, imagine the
article.\n
```

Target Template:

```
{{text[:7000]}}
```

Input Template:

```
{{text[:1000]}}... Continue the article for another 4000 characters max:
```

Target Template:

```
{{text[1000:5000]}}
```

Input Template:

```
...{{text[3000:3500]}}... Write the rest of the article:
```

Target Template:

```
{{text[5000:]}}
```

### 1.3 Translation

#### 1.3.1 Helsinki-NLP/tatoeba_mt ben-eng

Dataset from Tiedemann (2020). Used in training.

**Data Example**

| Key | Value |
|---|---|
| sourceLang | `ben` |
| targetlang | `eng` |
| sourceString | `Tatoebaর অর্থ কী?` |
| targetString | `What does "Tatoeba" mean?` |

**Prompts**

Input Template:

```
Translate the following text from English to Bengali {{ targetString }}
```

Target Template:

```
{{ sourceString }}
```

Input Template:

```
Translate the following text from Bengali to English {{ sourceString }}
```

Target Template:

```
{{ targetString }}
```

### 1.3.2 Helsinki-NLP/tatoeba_mt eng-fra

**Data Example**

| Key | Value |
| --- | --- |
| sourceLang | eng |
| targetlang | fra |
| sourceString | Aah. Now I understand. |
| targetString | Ah! Maintenant, je comprends. |

**Prompts**

Input Template:

```
Translate the following text from French to English {{ targetString }}
```

Target Template:

```
{{ sourceString }}
```

Input Template:

```
Translate the following text from English to French {{ sourceString }}
```

Target Template:

```
{{ targetString }}
```

### 1.3.3 facebook/flores ben_Beng-eng_Latn

Dataset from NLLB (2022). Used in training.

**Data Example**

| Key | Value |
| --- | --- |
| id | 2 |
| URL | `https://en.wikinews.org/wiki/Scient...` |
| domain | `wikinews` |
| topic | `health` |
| has_image | 0 |
| has_hyperlink | 0 |
| sentence_ben_Beng | শীর্ষ গবেষকরা বলছেন, এটি নিম্ন-আয়ে... |
| sentence_eng_Latn | `Lead researchers say this may bring...` |

**Prompts**

Input Template:

```
{{sentence_ben_Beng}}
```

Target Template:

```
{{sentence_eng_Latn}}
```

Input Template:

```
A text in Bengali: {{sentence_ben_Beng}}
```

Target Template:

```
{{sentence_eng_Latn}}
```

Input Template:

```
{{sentence_ben_Beng}}
```

Target Template:

```
{{sentence_eng_Latn}}
```

### 1.3.4 facebook/flores ben_Beng-fra_Latn

**Data Example**

| Key | Value |
| --- | --- |
| id | 2 |
| URL | https://en.wikinews.org/wiki/Scient... |
| domain | wikinews |
| topic | health |
| has_image | 0 |
| has_hyperlink | 0 |
| sentence_ben_Beng | শীর্ষ গবেষকরা বলেছেন, এটি নিম্ন-আয়ে... |
| sentence_fra_Latn | Selon les chercheurs principaux, ce... |

**Prompts**

Input Template:

```
{{sentence_ben_Beng}}
```

Target Template:

```
{{sentence_fra_Latn}}
```

Input Template:

```
{{sentence_ben_Beng}}
```

Target Template:

```
{{sentence_fra_Latn}}
```

Input Template:

```
A text in Bengali: {{sentence_ben_Beng}}
```

Target Template:

```
{{sentence_fra_Latn}}
```

### 1.4 Program Synthesis

#### 1.4.1 Muennighoff/mbpp sanitized

Dataset from Austin et al. (2021). Used in training.

**Data Example**

**Prompts**

Input Template:

| Key | Value |
| --- | --- |
| source_file | `Benchmark Questions Verification V2...` |
| task_id | `3` |
| prompt | `Write a python function to identify...` |
| code | `import math` |
|  | `def is_not_prime(n):` |
|  | `    ...` |
| test_imports |  |
| test_list | `assert is_not_prime(2) == False;ass...` |

`{{ prompt }} Here is a solution in Python:`

Target Template:

`{{ code }}`

Note: the prompt does not correspond to the original task intended by the dataset authors.

Input Template:

`{{ prompt }} This can be solved in Python with the following code:`

Target Template:

`{{ code }}`

### 1.4.2 CODEPARROT/APPS ALL

Dataset from Hendrycks et al. (2021). Used in training.

**Data Example**

| Key | Value |
| --- | --- |
| problem_id | `1` |
| question | `Mikhail walks on a Cartesian plane....` |
| solutions | `["q=int(input())\n\nfor e in range(...` |
| input_output | `{` |
|  | `    "inputs": [` |
|  | `        "3\n2 2 3\n4 3 ...` |
| difficulty | `interview` |
| url | `https://codeforces.com/problemset/p...` |
| starter_code |  |

**Prompts**

Input Template:

```
Solve in Python:
{{ question }}
```

Target Template:

```
{{ solution }}
```

Input Template:

```
{{ question }}

Can you solve the above problem using Python?
```

Target Template:

```
{{ solution }}
```

Input Template:

```
I found an interesting problem on {{url}}:
{{ question }}

I tried it in Python, but could not do it. Can you solve it?
```

Target Template:

```
{{ solution }}
```

### 1.4.3 codeparrot/github-jupyter-text-code-pairs

**Data Example**

| Key | Value |
|---|---|
| markdown | Extract the dataset from the compre... |
| code | num_classes = 10 |
|  | np.random.seed(133... |
| path | machine-learning/deep-learning/udac... |
| repo_name | pk-ai/training |
| license | mit |

**Prompts**

Input Template:

```
"{{ markdown }}"
Please write code following the instructions in jupyter notebook style.
```

Target Template:

```
{{ code }}
```

Input Template:

```
I am working on the file "{{ path }}".
The first task is:
{{ markdown }}
Can you write Python code for it?
```

Target Template:

```
{{ code }}
```

Note: the prompt does not correspond to the original task intended by the dataset authors.

Input Template:

```
{{ markdown }}
```

Target Template:

```
{{ code }}
```

Note: the prompt does not correspond to the original task intended by the dataset authors.

Input Template:

```
{{ code }}
Given the above code, generate some markdown instructions for it.
```

Target Template:

```
{{ markdown }}
```

### 1.4.4 codeparrot/xlcost-text-to-code C++-program-level

Dataset from Zhu et al. (2022). Used in training.

**Data Example**

### Data Example

| Key  | Value                                      |
|------|--------------------------------------------|
| text | Check if a number can be represente...     |
| code | #include <bits/stdc++.h> NEW_LINE u...     |

**Prompts**

Input Template:

```
"{{ text }}"
Solution in C++:
```

Target Template:

```
{{ code_clean }}
```

Input Template:

```
"{{ text }}"
How can the above be solved in C++?
```

Target Template:

```
{{ code_clean }}
```

### 1.4.5 codeparrot/xlcost-text-to-code C-program-level

**Data Example**

| Key  | Value                                     |
|------|-------------------------------------------|
| text | Logarithm tricks for Competitive Pr...    |
| code | #include <stdio.h> NEW_LINE #includ...    |

**Prompts**

Input Template:

```
"{{ text }}"
Solution in C:
```

Target Template:

```
{{ code_clean }}
```

Input Template:

```
{{ text }}
How can the above be solved in C?
```

Target Template:

```
{{ code_clean }}
```

### 1.4.6 codeparrot/xlcost-text-to-code Csharp-program-level

**Data Example**

| Key  | Value                              |
|------|------------------------------------|
| text | Check if a number can be represente... |
| code | using System ; class GFG { static b... |

**Prompts**

Input Template:

```
"{{ text }}"
Solution in C#:
```

Target Template:

```
{{ code_clean }}
```

Input Template:

```
"{{ text }}"
How can the above be solved in C-Sharp?
```

Target Template:

```
{{ code_clean }}
```

### 1.4.7 codeparrot/xlcost-text-to-code Java-program-level

**Data Example**

| Key  | Value                                   |
|------|-----------------------------------------|
| text | Check if a number can be represente...  |
| code | import java . io . * ; class GFG { ... |

**Prompts**

Input Template:

```
"{{ text }}"
Solution in Java:
```

Target Template:

```
{{ code_clean }}
```

Input Template:

```
"{{ text }}"
How can the above be solved in Java?
```

Target Template:

```
{{ code_clean }}
```

### 1.4.8 codeparrot/xlcost-text-to-code Javascript-program-level

**Data Example**

| Key  | Value                                        |
|------|----------------------------------------------|
| text | Check if a number can be represente...       |
| code | function sumOfTwoCubes ( n ) { var ...       |

**Prompts**

Input Template:

```
"{{ text }}"
Solution in Javascript:
```

Target Template:

```
{{ code_clean }}
```

Input Template:

```
"{{ text }}"
How can the above be solved in JS?
```

Target Template:

```
{{ code_clean }}
```

### 1.4.9 codeparrot/xlcost-text-to-code PHP-program-level

**Data Example**

| Key | Value |
|---|---|
| text | Rearrange the array to maximize the... |
| code | < ? php function solve ( $ a , $ n ... |

**Prompts**

Input Template:

```
"{{ text }}"
Solution in php:
```

Target Template:

```
{{ code_clean }}
```

Input Template:

```
"{{ text }}"
How can the above be solved in PHP?
```

Target Template:

```
{{ code_clean }}
```

### 1.4.10 codeparrot/xlcost-text-to-code Python-program-level

**Data Example**

| Key  | Value                                  |
|------|----------------------------------------|
| text | Check if a number can be represente... |
| code | import math NEW_LINE def sumOfTwoCu... |

**Prompts**

Input Template:

```
"{{ text }}"
Solution in Python:
```

Target Template:

```
{{ code_clean }}
```

Input Template:

```
"{{ text }}"
How can the above be solved in Python?
```

Target Template:

```
{{ code_clean }}
```

### 1.4.11 NEURAL_CODE_SEARCH EVALUATION_DATASET

Dataset from hug (2018). Used in training.

**Data Example**

| Key                 | Value                                   |
|---------------------|-----------------------------------------|
| stackoverflow_id    | 4616095                                 |
| question            | How to get the build/version number...  |
| question_url        | https://stackoverflow.com/questions...  |
| question_author     | Fahad Ali Shaikh                        |
| question_author_url | https://stackoverflow.com/users/565...  |
| answer              | try { <br>    PackageInfo pInfo = this.ge... |
| answer_url          | https://stackoverflow.com/a/6593822     |
| answer_author       | plus-                                   |
| answer_author_url   | https://stackoverflow.com/users/709...  |
| examples            | 4130029;3398176;2320640                 |
| examples_url        | https://github.com/altanizio/Concei...  |

**Prompts**

Note: the prompt does not correspond to the original task intended by the dataset authors.

Input Template:

```
Description:
{{ question }}

Implementation:
```

Target Template:

```
{{ answer }}
```

Note: the prompt does not correspond to the original task intended by the dataset authors.

Input Template:

```
Given the following code:
{{ answer }}
Describe it:
```

Target Template:

```
{{ question }}
```

### 1.4.12 teven/code_contests

**Data Example**

| Key | Value |
|---|---|
| name | 1575_A. Another Sorting Problem |
| description | Andi and Budi were given an assignm... |
| source | 2 |
| difficulty | 7 |
| solution | #include <bits/stdc++.h> using name... |
| language | CPP |

**Prompts**

Input Template:

```
{{description}}
```

Target Template:

```
{{solution}}
```

Input Template:

```
Can you solve the below in {{language}}?
{{description}}
```

Target Template:

```
{{solution}}
```

Input Template:

```
{{description}}
The above is tricky. Write me a correct solution in {{language}}.
```

Target Template:

```
{{solution}}
```

Input Template:

```
{{description}}
Solve the task in {{language}}.
```

Target Template:

```
{{solution}}
```

Input Template:

```
{{description}}
Using {{language | lower}} can you solve the prior task?
```

Target Template:

```
{{solution}}
```

Input Template:

```
{{description}}
{{solution[:5]}}
```

Target Template:

```
{{solution[5:]}}
```

Input Template:

```
{{language}} solution for "{{description}}":
```

Target Template:

```
{{solution}}
```

## 1.5 Coreference Resolution

### 1.5.1 Muennighoff/xwinograd en

Dataset from Tikhonov and Ryabinin (2021). Used in evaluation.

**Data Example**

| Key | Value |
| --- | --- |
| sentence | The city councilmen refused the dem... |
| option1 | The city councilmen |
| option2 | the demonstrators |
| answer | 2 |

**Prompts**

Input Template:

```
{{sentence}}
Replace the _ in the above sentence with the correct option:
- {{option1}}
- {{option2}}
```

Target Template:

```
{% if answer == '1' %} {{option1}} {% else %} {{ option2 }} {% endif %}
```

Answer Choices Template:

```
{{option1}} ||| {{option2}}
```

Input Template:

```
Fill in the _ in the below sentence:
{{sentence}}

Choices:
- {{ option1 }}
- {{ option2 }}

Answer:
```

Target Template:

```
{% if answer == '1' %} {{option1}} {% else %} {{ option2 }} {% endif %}
```

Answer Choices Template:

```
{{option1}} ||| {{option2}}
```

Note: the prompt does not correspond to the original task intended by the dataset authors.

Input Template:

```
The _ in the sentence below refers to {{option1}}. True or False?
{{sentence}}
```

Target Template:

```
{{answer_choices[answer|int - 1]}}
```

Answer Choices Template:

```
True ||| False
```

Input Template:

```
{{ sentence }} In the previous sentence, does _ refer to {{ option1 }} or  {{ option2 }}?
```

Target Template:

```
{% if answer == '1' %} {{option1}} {% else %} {{ option2 }} {% endif %}
```

Answer Choices Template:

```
{{ option1 }} ||| {{ option2 }}
```

Input Template:

```
{{sentence}}
What does the _ in the above sentence refer to? {{ option1 }} or {{ option2 }}?
```

Target Template:

```
{% if answer == '1' %} {{option1}} {% else %} {{ option2 }} {% endif %}
```

Answer Choices Template:

```
{{option1}} ||| {{option2}}
```

Input Template:

```
In the sentence below, does the _ stand for {{answer_choices[0]}} or
{{answer_choices[1]}}?
{{sentence}}
```

Target Template:

```
{{answer_choices[answer | int - 1]}}
```

Answer Choices Template:

```
{{option1}} ||| {{option2}}
```

### 1.5.2 Muennighoff/xwinograd fr

**Data Example**

| Key | Value |
|---|---|
| sentence | La coupe n'entre pas dans la valise... |
| option1 | La coupe |
| option2 | la valise |
| answer | 2 |

**Prompts**

Input Template:

```
{{ sentence }} Dans la phrase précédente, _ fait-il référence à {{ option1 }} ou
{{ option2 }} ?
```

Target Template:

```
{% if answer == '1' %} {{option1}} {% else %} {{ option2 }} {% endif %}
```

Answer Choices Template:

```
{{ option1 }} ||| {{ option2 }}
```

Input Template:

```
Dans la phrase ci-dessous, le _ signifie-t-il {{answer_choices[0]}} ou {{answer_choices[1]}} ?
{{sentence}}
```

Target Template:

```
{{answer_choices[answer | int - 1]}}
```

Answer Choices Template:

```
{{option1}} ||| {{option2}}
```

Input Template:

```
{{sentence}}
Remplacez le _ dans la phrase ci-dessus par l'option correcte :
- {{option1}}
- {{option2}}
```

Target Template:

```
{% if answer == '1' %} {{option1}} {% else %} {{ option2 }} {% endif %}
```

Answer Choices Template:

```
{{option1}} ||| {{option2}}
```

Input Template:

```
{{sentence}}
À quoi le _ dans la phrase ci-dessus fait-il référence ? {{ option1 }} ou {{ option2 }} ?
```

Target Template:

```
{% if answer == '1' %} {{option1}} {% else %} {{ option2 }} {% endif %}
```

Answer Choices Template:

```
{{option1}} ||| {{option2}}
```

Input Template:

```
Le _ dans la phrase ci-dessous fait référence à {{option1}}. Vrai ou faux?
{{sentence}}
```

Target Template:

```
{{answer_choices[answer|int - 1]}}
```

Answer Choices Template:

```
Vrai ||| Faux
```

## 1.6 QUESTION ANSWERING MULTIPLE CHOICE

### 1.6.1 CLUE C3

Dataset from Sun et al. (2020). Used in training.

**Data Example**

| Key | Value |
|---|---|
| id | 1 |
| context | 男：足球比赛是明天上午八点开始吧?;女：因为天气不好，比赛改到后天下午... |
| question | 根据对话，可以知道什么? |
| choice | 今天天气不好;比赛时间变了;校长忘了时间 |
| answer | 比赛时间变了 |

**Prompts**

Input Template:

```
{% for statement in context %}
{{ statement }}
{% endfor %}
鉴于上面的对话/段落，问题 "{{question}}" 的答案是什么
```

Target Template:

```
{{ answer }}
```

Input Template:

```
段落: {% for statement in context %}
{{ statement }}
{% endfor %}
什么样的问题会引起 {{ answer }} 的回答响应？
```

Target Template:

```
{{ question }}
```

Input Template:

```
{% for statement in context %}
{{ statement }}
{% endfor %}
Given the dialogue / passage above, use the following options to answer the
question "{{question}}".
Options:
- {{ answer_choices | join('\n- ') }}
```

Target Template:

```
{{ answer }}
```

Answer Choices Template:

```
{{ choice | join(" ||| ") }}
```

Input Template:

```
{% for statement in context %}
{{ statement }}
{% endfor %}
鉴于上面的对话/段落，使用以下选项回答问题 "{{question}}"。
选项:
- {{ answer_choices | join('
- ') }}
```

Target Template:

```
{{ answer }}
```

Answer Choices Template:

```
{{ choice | join(" ||| ") }}
```

Input Template:

```
Passage: {% for statement in context %}
{{ statement }}
{% endfor %}
Question: "{{question}}"
Answer choices: {{ answer_choices[:-1] | join(', ') }}, or  {{ answer_choices[-1] }}?
```

Target Template:

```
{{ answer }}
```

Answer Choices Template:

```
{{ choice | join(" ||| ") }}
```

Note: the prompt does not correspond to the original task intended by the dataset authors.

Input Template:

```
Passage: {% for statement in context %}
{{ statement }}
{% endfor %}
What kind of question would elicit an answer response of {{ answer }}?
```

Target Template:

```
{{ question }}
```

Input Template:

```
段落: {% for statement in context %}
{{ statement }}
{% endfor %}
问题: "{{question}}"
答案选择: {{ answer_choices[:-1] | join(', ') }} 还是 {{ answer_choices[-1] }} ?
```

Target Template:

```
{{ answer }}
```

Answer Choices Template:

```
{{ choice | join(' ||| ') }}
```

Input Template:

```
{% for statement in context %}
{{ statement }}
{% endfor %}
Given the dialogue / passage above, what is the answer for the question
"{{question}}"
Answer choices: {{ answer_choices[:-1] | join(', ') }}, or  {{ answer_choices[-1] }}?
```

Target Template:

```
{{ answer }}
```

Answer Choices Template:

```
{{ choice | join(' ||| ') }}
```

Input Template:

```
{% for statement in context %}
{{ statement }}
{% endfor %}
鉴于上面的对话/段落，问题 "{{question}}" 的答案是什么
答案选择: {{ answer_choices[:-1] | join(', ') }} 还是 {{ answer_choices[-1] }} ?
```

Target Template:

```
{{ answer }}
```

Answer Choices Template:

```
{{ choice | join(' ||| ') }}
```

Note: the prompt does not correspond to the original task intended by the dataset authors.

Input Template:

```
{% for statement in context %}
{{ statement }}
{% endfor %}
Given the dialogue / passage above, what is the answer for the question
"{{question}}"
```

Target Template:

```
{{ answer }}
```

## 1.7 Question Answering Extractive

### 1.7.1 clue cmrc2018

Dataset from Cui et al. (2018). Used in training.

**Data Example**

| Key | Value |
| --- | --- |
| id | `TRAIN_186_QUERY_1` |
| context | 范廷颂枢机 (), 圣名保禄·若瑟 (), 是越南罗马天主教枢机。1963年... |
| question | 1990年，范廷颂担任什么职务？ |
| answers | `{'text': ['1990年被擢升为天主教河内总教区宗座署理'],...` |

**Prompts**

Input Template:

```
问: {{ question }}你能写一些上下文来回答这个问题吗？
```

Target Template:

```
{{ context }}
```

Note: the prompt does not correspond to the original task intended by the dataset authors.

Input Template:

```
Given this context "{{ context }}", generate a question that would return the
answer of "{{ answers['text'][0] }}".
```

Target Template:

```
{{ question }}
```

Input Template:

```
{{ context }}
{{ question }} 的答案在上面的段落中。它是什么？
```

Target Template:

```
{{ answers['text'][0] }}
```

Input Template:

```
In an exam, you are asked {{ question }}, and you are tasked to find the answer
from the following passage.
{{ context }}
What's the answer?
```

Target Template:

```
{{ answers['text'][0] }}
```

Input Template:

```
{{ context }}
The answer to {{ question }} is in the passage above. What is it?
```

Target Template:

```
{{ answers['text'][0] }}
```

Input Template:

```
Answer the question using the given context.
Question: {{ question }}
Context: {{ context }}
Answer:
```

Target Template:

```
{{ answers['text'][0] }}
```

Input Template:

```
Q: {{ question }} Can you write some context to answer the question?
```

Target Template:

```
{{ context }}
```

Input Template:

```
{{ context[:answers["answer_start"][0]-5] }}... How would you continue the prior
text to answer "{{ question }}"?
```

Target Template:

```
{{ context[answers["answer_start"][0]-5:] }}
```

Input Template:

```
{{ context[:answers["answer_start"][0]-5] }}... 你将如何继续前面的文本来回答 "{{ question }}"？
```

Target Template:

```
{{ context[answers["answer_start"][0]-5:] }}
```

Input Template:

```
使用给定的上下文回答问题。
问题: {{ question }}
上下文: {{ context }}
答案:
```

Target Template:

```
{{ answers['text'][0] }}
```

Input Template:

```
在考试中，你被问到 {{ question }}，你的任务是从以下段落中找到答案。
{{ context }}
答案是什么？
```

Target Template:

```
{{ answers['text'][0] }}
```

Input Template:

```
给定这个上下文 "{{ context }}"，生成一个返回 "{{ answers['text'][0] }}" 答案的问题。
```

Target Template:

```
{{ question }}
```

### 1.7.2 CLUE DRCD

Dataset from Xu et al. (2020). Used in training.

**Data Example**

| Key | Value |
|---|---|
| id | 1001-10-2 |
| context | 2010年引進的廣州快速公交運輸系統，屬世界第二大快速公交系統，日常載... |
| question | 從哪一天開始在廣州市𝔽騎摩托車會被𝔽收？ |
| answers | {'text': ['2007年1月16日'], 'answer_st... |

**Prompts**

Input Template:

```
{{ context }}
{{ question }} 的答案在上面的段落中。它是什麼？
```

Target Template:

```
{{ answers['text'][0] }}
```

Input Template:

```
Answer the question using the given context.
Question: {{ question }}
Context: {{ context }}
Answer:
```

Target Template:

```
{{ answers['text'][0] }}
```

Input Template:

```
{{context[:answers["answer_start"]-5]}}... 你将如何继续前面的文本来回答 "{{question}}" ?
```

Target Template:

```
{{context[answers["answer_start"]-5:]}}
```

Input Template:

```
在考试中，你被问到 {{ question }}，你的任务是找到回答问题的段落。写这样一段话：
```

Target Template:

```
{{ context }}
```

Input Template:

```
{{ context }}
The answer to {{ question }} is in the passage above. What is it?
```

Target Template:

```
{{ answers['text'][0] }}
```

Input Template:

```
在考试中，你被问到 {{ question }}，你的任务是从以下段落中找到答案。
{{ context }}
答案是什么？
```

Target Template:

```
{{ answers['text'][0] }}
```

Input Template:

```
给定这个上下文 "{{ context }}"，生成一个返回 "{{ answers['text'][0] }}" 答案的问题。
```

Target Template:

```
{{ question }}
```

Input Template:

```
{{context[:answers["answer_start"]-5]}}... How would you continue the prior text to answer "{{ question }}"?
```

Target Template:

```
{{context[answers["answer_start"]-5:]}}
```

Input Template:

```
使用给定的上下文回答问题。
问题：{{ question }}
上下文：{{ context }}
答案：
```

Target Template:

```
{{ answers['text'][0] }}
```

Note: the prompt does not correspond to the original task intended by the dataset authors.

Input Template:

```
Given this context "{{ context }}", generate a question that would return the
answer of "{{ answers['text'][0] }}".
```

Target Template:

```
{{ question }}
```

Input Template:

```
In an exam, you are asked {{ question }}, and you are tasked to find the answer
from the following passage.
{{ context }}
What's the answer?
```

Target Template:

```
{{ answers['text'][0] }}
```

Input Template:

```
In an exam, you are asked {{ question }}, and you are tasked to find a passage
answering the question. Write such a passage:
```

Target Template:

```
{{ context }}
```

### 1.7.3 MLQA MLQA.VI.VI

Dataset from Lewis et al. (2019). Used in training.

**Data Example**

| Key | Value |
|---|---|
| context | Thành phố Miêu Lật tiếng Trung:苗栗市,... |
| question | Miaoli có tỷ lệ cao loại người nào? |
| answers | {'answer_start': [311], 'text': ['K... |
| id | 2f0d6ff162619164bb113c0cadbcca06a50... |

**Prompts**

Input Template:

```
{{ context[:answers.answer_start[0]-5]}} ... Tiếp tục ở trên, sao cho nó trả lời
"{{question}}":
```

Target Template:

```
{{ context[answers.answer_start[0]-5:]}}
```

Input Template:

```
{{context}}

Với sự tham chiếu đến ngữ cảnh trên, {{question}}
```

Target Template:

```
{{answers.text[0]}}
```

Input Template:

```
{{context}}
H: {{question}}
Đề cập đến đoạn văn trên, câu trả lời đúng cho câu hỏi đã cho trong ngôn ngữ của
đoạn văn là
```

Target Template:

```
{{answers["text"][0]}}
```

Input Template:

```
Câu hỏi: {{question}}
 Ngữ cảnh: {{context}}
 Câu trả lời từ ngữ cảnh:
```

Target Template:

```
{{answers.text[0]}}
```

Input Template:

```
Tham khảo đoạn văn dưới đây và sau đó trả lời câu hỏi sau đó bằng ngôn ngữ tương
tự như đoạn văn:

Đoạn: {{context}}

Câu hỏi: {{question}}
```

Target Template:

```
{{answers["text"][0]}}
```

Input Template:

```
Tôi đã tìm thấy một văn bản trả lời "{{question}}" bằng {{answers.text[0]}}. Nó
bắt đầu bằng "{{ context[:10] }}". Bạn có thể tiếp tục nó không?
```

Target Template:

```
{{ context[10:] }}
```

Input Template:

```
Đọc đoạn văn sau và sau đó trả lời câu hỏi tiếp theo bằng cách trích một phần đúng
trong đoạn văn:
{{context}}
{{question}}
```

Target Template:

```
{{answers.text[0]}}
```

Input Template:

```
D: {{context}}

H: {{question}}

A:
```

Target Template:

```
{{answers["text"][0]}}
```

### 1.7.4  MLQA MLQA.ZH.ZH

**Data Example**

| Key | Value |
|---|---|
| context | 楚河州包括有整个楚河河谷及邻近的山脉与峡谷。河谷的黑土非常肥沃，而且被… |
| question | 哪水体有助土地如此多产？ |
| answers | {'answer_start': [36], 'text': ['楚河… |
| id | 1aee17dd937cc1043e3ff47c38396541fc3… |

**Prompts**

Input Template:

```
阅读下面的短文，然后从短文中选出正确的部分来回答下面的问题：
{{context}}
{{question}}
```

Target Template:

```
{{answers.text[0]}}
```

Input Template:

```
{{ context[:answers.answer_start[0]-5]}}... 继续上述操作，使其回答"{{question}}":
```

Target Template:

```
{{ context[answers.answer_start[0]-5:]}}
```

Input Template:

```
D: {{context}}

问: {{question}}

答:
```

Target Template:

```
{{answers["text"][0]}}
```

Input Template:

我找到了一个用 `{{answers.text[0]}}` 回答 "`{{answers.text[0]}}`" 的文本。它以 "`{{ context[:10] }}`" 开头。可以继续吗？

Target Template:

`{{ context[10:] }}`

Input Template:

问题：`{{question}}`
上下文：`{{context}}`
从上下文中回答：

Target Template:

`{{answers.text[0]}}`

Input Template:

参考下面的段落，然后用与段落相同的语言回答问题：

段落：`{{context}}`

问题：`{{question}}`

Target Template:

`{{answers["text"][0]}}`

Input Template:

`{{context}}`

参考上述上下文，`{{question}}`

Target Template:

`{{answers.text[0]}}`

Input Template:

```
{{context}}
问：{{question}}
参考上面的段落，用该段落的语言对给定问题的正确答案是
```

Target Template:

```
{{answers["text"][0]}}
```

### 1.7.5 XQUAD XQUAD.VI

Dataset from Artetxe et al. (2019). Used in training.

**Data Example**

| Key | Value |
| --- | --- |
| id | 56beb4343aeaaa14008c925c |
| context | Đội thủ của Panthers chỉ thua 308 đ... |
| question | Jared Allen có bao nhiêu lần vật ng... |
| answers | {'text': ['136'], 'answer_start': [... |

**Prompts**

Input Template:

```
{{context}}

Với sự tham chiếu đến ngữ cảnh trên, {{question}}
```

Target Template:

```
{{answers.text[0]}}
```

Input Template:

```
Đưa ra câu trả lời {{answers.text[0]}} cho {{question}}, hãy viết một văn bản giải
thích điều này. Câu trả lời phải bắt đầu ở số ký tự {{answers.answer_start[0]}}.
Văn bản:
```

Target Template:

```
{{context}}
```

Input Template:

```
{{question}} Rõ ràng là {{answers.text[0]}}. Bạn có thể cung cấp cho tôi một số
bối cảnh?
```

Target Template:

```
{{context}}
```

Input Template:

```
{{context}}

H: {{question}}

Đề cập đến đoạn văn trên, câu trả lời chính xác cho câu hỏi được đưa ra là
```

Target Template:

```
{{answers["text"][0]}}
```

Input Template:

```
{{context}}

H: {{question}}

A:
```

Target Template:

```
{{answers["text"][0]}}
```

Input Template:

```
Đọc đoạn văn sau và trả lời câu hỏi sau:
{{context}}
{{question}}
```

Target Template:

```
{{answers.text[0]}}
```

Input Template:

```
Tham khảo đoạn văn dưới đây và trả lời câu hỏi sau:

Đoạn: {{context}}

Câu hỏi: {{question}}
```

Target Template:

```
{{answers["text"][0]}}
```

Input Template:

```
{{context}}

Từ đoạn văn trên, một câu hỏi hợp lý với "{{answers["text"][0]}}" như câu trả lời
sẽ là:
```

Target Template:

```
{{question}}
```

Input Template:

```
{{context}}

Tạo câu hỏi từ đoạn văn trên:
```

Target Template:

```
{{question}}
```

### 1.7.6 XQUAD XQUAD.ZH

**Data Example**

| Key | Value |
|---|---|
| id | 56beb4343aeaaa14008c925c |
| context | 黑豹队的防守只丢了 308分，在联赛中排名第六，同时也以 24 次拦截... |
| question | 贾里德在职业生涯中有多少次擒杀？ |
| answers | {'text': ['136 次'], 'answer_start':... |

**Prompts**

Input Template:

阅读下面的短文，回答下面的问题：
`{{context}}`
`{{question}}`

Target Template:

`{{answers.text[0]}}`

Input Template:

`{{context}}`

问：`{{question}}`

参考上面的段落，给定问题的正确答案是

Target Template:

`{{answers["text"][0]}}`

Input Template:

参考下面的短文，回答下列问题：

段落：`{{context}}`

问题：`{{question}}`

Target Template:

`{{answers["text"][0]}}`

Input Template:

`{{context}}`

从上面的段落中，一个以"`{{answers["text"][0]}}`"为答案的合理问题将是：

Target Template:

`{{question}}`

Input Template:

```
{{context}}

从上面的段落中产生一个问题：
```

Target Template:

```
{{question}}
```

Input Template:

```
{{context}}

参考上述上下文，{{question}}
```

Target Template:

```
{{answers.text[0]}}
```

Input Template:

```
{{context}}

问：{{question}}

答：
```

Target Template:

```
{{answers["text"][0]}}
```

## 1.8 Topic Classification

### 1.8.1 CLUE CSL

**Data Example**

| Key | Value |
| --- | --- |
| idx | 1 |
| corpus_id | 2565 |
| abst | 针对核函数参数选择的重要性,提出了粒子群(PSO)模式搜索算法来搜索最... |
| label | -1 |
| keyword | 模式搜索;支持向量机;核参数选取 |

**Prompts**

Input Template:

```
After John wrote the abstract "{{abst}}", he wrote these keywords "{{ keyword |
join(', ') }}". Do you think his choice of keywords was correct? Answer {{
answer_choices[1]}} or {{ answer_choices[0]}}.
```

Target Template:

```
{{ answer_choices[label] }}
```

Answer Choices Template:

```
no ||| yes
```

Input Template:

```
Do these keywords "{{ keyword | join(', ') }}" represent key concepts in the
abstract "{{ abst }}"?
```

Target Template:

```
{{ answer_choices[label] }}
```

Answer Choices Template:

```
no ||| yes
```

Note: the prompt does not correspond to the original task intended by the dataset authors.

Input Template:

```
Given the abstract {{abst}}, list out {{ keyword | length }} keywords for it.
```

Target Template:

```
{% if label == 1 %}
{{ keyword | join(', ') }}
{% endif %}
```

Input Template:

```
一位学者使用 "{{ keyword | join(', ') }}" 作为搜索词。你认为搜索引擎会返回摘要
"{{abst}}" 吗？回答 {{ answer_choices[1] }} 或 {{ answer_choices[0] }}。
```

Target Template:

```
{{ answer_choices[label] }}
```

Answer Choices Template:

```
不 ||| 是的
```

Input Template:

```
给定抽象{{abst}}，列出{{ keyword | length }} 关键字。
```

Target Template:

```
{% if label == 1 %}
{{ keyword | join(', ') }}
{% endif %}
```

Input Template:

```
写一篇关于 "{{ keyword | join(', ') }}" 的摘要：
```

Target Template:

```
{% if label == 1 %} {{abst}} {% endif %}
```

Answer Choices Template:

```
不 ||| 是的
```

Input Template:

```
在约翰写完摘要 "{{abst}}" 之后，他写了这些关键字 "{{ keyword | join(', ') }}"。你认为他选择的关键词是正确的吗？回答 {{ answer_choices[1]}} 或 {{ answer_choices[0]}}。
```

Target Template:

```
{{ answer_choices[label] }}
```

Answer Choices Template:

```
不 ||| 是的
```

Input Template:

这些关键字 "`{{ keyword | join(', ') }}`" 是否代表抽象 "`{{ abst }}`" 中的关键概念？

Target Template:

`{{ answer_choices[label] }}`

Answer Choices Template:

`不 ||| 是的`

Input Template:

```
A scholar used "{{ keyword | join(', ') }}" as search terms. Do you think the
search engine would return the abstract "{{abst}}"? Answer {{ answer_choices[1] }}
or {{ answer_choices[0] }}.
```

Target Template:

`{{ answer_choices[label] }}`

Answer Choices Template:

`no ||| yes`

Input Template:

`Write an abstract about "{{ keyword | join(', ') }}":`

Target Template:

`{% if label == 1 %} {{abst}} {% endif %}`

Answer Choices Template:

`no ||| yes`

### 1.8.2 CLUE TNEWS

**Data Example**

**Prompts**

Input Template:

| Key | Value |
|---|---|
| sentence | 买套房不香吗？为什么会有人愿花600万买部手机？ |
| label | -1 |
| idx | 1 |

```
将标题 "{{ sentence }}" 分为以下主题:
- {{ answer_choices | join('\n- ') }}
主题:
```

Target Template:

```
{{ answer_choices[label] }}
```

Answer Choices Template:

```
故事 ||| 文化 ||| 娱乐 ||| 运动的 ||| 金融 ||| 房地产 ||| 车 ||| 教育 ||| 技术 |||
军队 ||| 旅行 ||| 世界新闻 ||| 股票 ||| 农业 ||| 游戏
```

Input Template:

```
Classify the title "{{ sentence }}" into the following topics:
- {{ answer_choices | join('\n- ') }}
Topic:
```

Target Template:

```
{{ answer_choices[label] }}
```

Answer Choices Template:

```
story ||| culture ||| entertainment ||| sports ||| finance ||| real estate ||| car
||| education ||| tech ||| military ||| travel ||| world news ||| stock |||
agriculture ||| game
```

Input Template:

```
Given the topics of {{answer_choices[:-1] | join(', ') }}, and {{
answer_choices[-1] }}, specify which of them best represents the following
sentence:
{{ sentence }}

Best:
```

Target Template:

```
{{ answer_choices[label] }}
```

Answer Choices Template:

```
story ||| culture ||| entertainment ||| sports ||| finance ||| real estate ||| car ||| education ||| tech ||| military ||| travel ||| world news ||| stock ||| agriculture ||| game
```

Input Template:

```
以下新闻标题 "{{ sentence }}" 属于什么主题？{{ answer_choices[0] | capitalize }}, {{ answer_choices[1:-1] | join(', ') }} 还是 {{ answer_choices[-1] }}？
```

Target Template:

```
{{ answer_choices[label] }}
```

Answer Choices Template:

```
故事 ||| 文化 ||| 娱乐 ||| 运动的 ||| 金融 ||| 房地产 ||| 车 ||| 教育 ||| 技术 ||| 军队 ||| 旅行 ||| 世界新闻 ||| 股票 ||| 农业 ||| 游戏
```

Input Template:

```
鉴于 {{answer_choices[:-1] | join(', ') }} 和 {{ answer_choices[-1] }}, 指定它们中的哪一个最能代表以下句子:
{{ sentence }}

最佳:
```

Target Template:

```
{{ answer_choices[label] }}
```

Answer Choices Template:

```
故事 ||| 文化 ||| 娱乐 ||| 运动的 ||| 金融 ||| 房地产 ||| 车 ||| 教育 ||| 技术 ||| 军队 ||| 旅行 ||| 世界新闻 ||| 股票 ||| 农业 ||| 游戏
```

Input Template:

```
What topic does the following news title "{{ sentence }}" belong to? {{ answer_choices[0] | capitalize }}, {{ answer_choices[1:-1] | join(', ') }}, or {{ answer_choices[-1] }}?
```

Target Template:

```
{{ answer_choices[label] }}
```

Answer Choices Template:

```
story ||| culture ||| entertainment ||| sports ||| finance ||| real estate ||| car
||| education ||| tech ||| military ||| travel ||| world news ||| stock |||
agriculture ||| game
```

## 1.9 CODE MISC.

### 1.9.1 CODEPARROT/CODECOMPLEX CODEPARROT–CODECOMPLEX

**Data Example**

| Key | Value |
|---|---|
| src | `import java.util.Scanner; public ...` |
| complexity | `linear` |
| problem | `1197_B. Pillars` |
| from | `CODEFORCES` |

**Prompts**

Input Template:

```
{{ code }} What is the time complexity of the previous code?
```

Target Template:

```
{{ complexity }}
```

Note: the prompt does not correspond to the original task intended by the dataset authors.

Input Template:

```
Identify the time complexity of the following code as constant, linear, quadratic,
cubic, log(n), nlog(n) or NP-hard. {{ code }} Complexity:
```

Target Template:

```
{{ complexity }}
```

Note: the prompt does not correspond to the original task intended by the dataset authors.

Input Template:

```
{{ code }} Which one is the correct time complexity of the code snippet: constant,
linear, quadratic, cubic, log(n), nlog(n) or NP-hard?
```

Target Template:

```
{{ complexity }}
```

### 1.9.2 GREAT_CODE

Dataset from Hellendoorn et al. (2020). Used in training.

**Data Example**

| Key | Value |
| --- | --- |
| id | 1 |
| source_tokens | #NEWLINE#;def test_get_params(;self... |
| has_bug | True |
| error_location | 76 |
| repair_candidates | 2;76;4;11;18;22;30;40;48;58;66;80;8... |
| bug_kind | 1 |
| bug_kind_name | VARIABLE_MISUSE |
| repair_targets | 4;11;18;22;30;40;48;58;66;80;88;103... |
| edges | [{'before_index': 1, 'after_index':... |
| provenances | {'datasetProvenance': {'datasetName... |

**Prompts**

Note: the prompt does not correspond to the original task intended by the dataset authors.

Input Template:

```
{% set mask = 'def <FUNC_NAME> (' %}
{% set indent = '    ' %}
{% set ns = namespace(indent_size=0, result=[], masked=false, target='') %}
{% for token in source_tokens  %}
    {% if ns.masked is false and token.startswith('def') %}
            {% set ns.target = token.split('def ')[1][:-1] %}
            {% set token = mask %}
            {% set ns.masked = true %}
    {% endif%}
    {% if token== '#INDENT#' %}
        {% set ns.indent_size = ns.indent_size + 1 %}
        {% set ns.result = ns.result + [indent * ns.indent_size] %}
    {% elif token == '#NEWLINE#' %}
         {% set ns.result = ns.result + ["\n"] %}
    {% elif token == '#UNINDENT#' %}
       {% set ns.indent_size = ns.indent_size - 1 %}
    {% else %}
        {% if not loop.first and loop.previtem == '#NEWLINE#' %}
             {% set ns.result = ns.result + [indent * ns.indent_size] %}
        {% endif %}
        {% set ns.result = ns.result + [token | replace('\\n', '\n'), " "] %}
    {% endif %}
{% endfor %}
{{ns.result | join("") | replace(" . ", ".") | replace(" , ", ", ")  | replace("( ", "(") | replace(" )", ")") | replace("[ ", "[") | replace(" ]", "]")}}

What is the function name?
```

Target Template:

```
{{ ns.target }}
```

Note: the prompt does not correspond to the original task intended by the dataset authors.

Input Template:

```
{% set result = "" %}
{% set indent = '    ' %}
{% set ns = namespace(indent_size=0, line_number=0, buggy_line=0, bug_location=0, bug_len=0, result=[], result_lines=[]) %}
{% set fixed_token = source_tokens[repair_targets[0]] %}
{% set buggy_line_content = "" %}
{% set fixed_buggy_line_content = "" %}

{% if has_bug and (repair_targets | length > 0) %}
    {% for token in source_tokens  %}
        {% if loop.index0 == error_location %}
                {% set ns.buggy_line = ns.line_number %}
                {% set ns.bug_location = (ns.result | join("") | length) %}
                {% set ns.bug_len = (token | length) %}
        {% endif%}
        {% if token== '#INDENT#' %}
            {% set ns.indent_size = ns.indent_size + 1 %}
            {% set ns.result = ns.result + [indent * ns.indent_size] %}
        {% elif token == '#NEWLINE#' %}
                {% set ns.result_lines = ns.result_lines + [ns.result | join("")] %}
                {% set ns.result = [] %}
                {% set ns.line_number = ns.line_number + 1 %}
        {% elif token == '#UNINDENT#' %}
             {% set ns.indent_size = ns.indent_size - 1 %}
        {% else %}
                {% if not loop.first and loop.previtem == '#NEWLINE#' %}
                    {% set ns.result = ns.result + [indent * ns.indent_size] %}
                {% endif %}
                {% set ns.result = ns.result + [token | replace('\\n', '\n'), " "] %}
        {% endif %}
    {% endfor %}
    {% set ns.result_lines = ns.result_lines + [ns.result | join("")] %}
    {% set result = ns.result_lines | join("\n") %}
    {{result | replace(" . ", ".") | replace(" , ", ", ")  | replace("( ", "(") | replace(" )", ")") | replace("[ ", "[") | replace(" ]", "]")}}

    {% set buggy_line_content = ns.result_lines[ns.buggy_line] | trim | replace(" . ", ".") | replace(" , ", ", ")  | replace("( ", "(") | replace(" )", ")") | replace("[ ", "[") | replace(" ]", "]") %}
    {% set fixed_buggy_line_content = (ns.result_lines[ns.buggy_line][:ns.bug_location] + fixed_token + ns.result_lines[ns.buggy_line][ns.bug_location + ns.bug_len:]) | trim | replace(" . ", ".") | replace(" , ", ", ")  | replace("( ", "(") | replace(" )", ")") | replace("[ ", "[")  | replace(" ]", "]")%}

    Fix the buggy line: {{buggy_line_content}}
```

Target Template:

```
{{fixed_buggy_line_content}}
{% endif %}
```

Input Template:

```
{% set mask = '<MASK>' %}
{% set indent = '    ' %}
{% set ns = namespace(indent_size=0, result=[]) %}

{% if has_bug %}
    {% for token in source_tokens  %}
        {% if loop.index0 == error_location %}
                {% set token = mask %}
        {% endif%}
        {% if token== '#INDENT#' %}
            {% set ns.indent_size = ns.indent_size + 1 %}
            {% set ns.result = ns.result + [indent * ns.indent_size] %}
        {% elif token == '#NEWLINE#' %}
              {% set ns.result = ns.result + ["\n"] %}
        {% elif token == '#UNINDENT#' %}
            {% set ns.indent_size = ns.indent_size - 1 %}
        {% else %}
              {% if not loop.first and loop.previtem == '#NEWLINE#' %}
                  {% set ns.result = ns.result + [indent * ns.indent_size] %}
              {% endif %}
              {% set ns.result = ns.result + [token | replace('\\n', '\n'), " "] %}
        {% endif %}
    {% endfor %}
    {{ns.result | join("") | replace(" . ", ".") | replace(" , ", ", ") | replace("( ", "(") | replace(" )", ")") | replace("[ ", "[") | replace(" ]", "]")}}

    Given the code above, what is a proper replacement for {{mask}}?
```

Target Template:

```
{{source_tokens[repair_targets[0]]}}
{% endif %}
```

Note: the prompt does not correspond to the original task intended by the dataset authors.

Input Template:

```
{% set indent = '    ' %}
{% set ns = namespace(indent_size=0, result=[]) %}
{% for token in source_tokens  %}
    {% if token== '#INDENT#' %}
        {% set ns.indent_size = ns.indent_size + 1 %}
        {% set ns.result = ns.result + [indent * ns.indent_size] %}
    {% elif token == '#NEWLINE#' %}
          {% set ns.result = ns.result + ["\n"] %}
    {% elif token == '#UNINDENT#' %}
        {% set ns.indent_size = ns.indent_size - 1 %}
    {% else %}
          {% if not loop.first and loop.previtem == '#NEWLINE#' %}
              {% set ns.result = ns.result + [indent * ns.indent_size] %}
```

```
            {% endif %}
            {% set ns.result = ns.result + [token | replace('\\n', '\n'), " "] %}
        {% endif %}
{% endfor %}
{{ns.result | join("") | replace(" . ", ".") | replace(" , ", ", ")  | replace("( ", "(") | replace(" )", ")") | replace("[ ", "[") | replace(" ]", "]")}}

Is there a bug in the code above?
```

Target Template:

```
{{ {True: "Yes", False: "No"}[has_bug] }}
```

Answer Choices Template:

```
Yes ||| No
```

Note: the prompt does not correspond to the original task intended by the dataset authors.

Input Template:

```
{% set mask = '<MASK>' %}
{% set indent = '    ' %}
{% set ns = namespace(indent_size=0, result=[]) %}

{% if has_bug %}
    {% for token in source_tokens  %}
        {% if loop.index0 == error_location %}
                {% set token = mask %}
        {% endif%}
        {% if token== '#INDENT#' %}
            {% set ns.indent_size = ns.indent_size + 1 %}
            {% set ns.result = ns.result + [indent * ns.indent_size] %}
        {% elif token == '#NEWLINE#' %}
                {% set ns.result = ns.result + ["\n"] %}
        {% elif token == '#UNINDENT#' %}
             {% set ns.indent_size = ns.indent_size - 1 %}
        {% else %}
                {% if not loop.first and loop.previtem == '#NEWLINE#' %}
                    {% set ns.result = ns.result + [indent * ns.indent_size] %}
                {% endif %}
                {% set ns.result = ns.result + [token | replace('\\n', '\n'), " "] %}
        {% endif %}
    {% endfor %}
    {{ns.result | join("") | replace(" . ", ".") | replace(" , ", ", ")  | replace("( ", "(") | replace(" )", ")") | replace("[ ", "[") | replace(" ]", "]")}}

    Given the code above, what is a proper replacement for {{mask}}? Choose among: {{answer_choices | join(", ")}}
```

Target Template:

```
{{source_tokens[repair_targets[0]]}}
{% endif %}
```

Answer Choices Template:

```
{% if has_bug %}     {% set nss = namespace(choices=[]) %}     {% for i in
repair_candidates  %}          {% set nss.choices = nss.choices + [source_tokens[(i
| int)]] %}     {% endfor %}     {{nss.choices | unique | join(" ||| ")}} {% endif
%}
```

### 1.9.3 TEVEN/CODE_DOCSTRING_CORPUS TOP_LEVEL

**Data Example**

| Key    | Value                          |
|--------|--------------------------------|
| desc   | 'XXX22: This has to be present' |
| decl   | def XXX11():                   |
| bodies |     pass   |

**Prompts**

Input Template:

```
Complete the below
{{decl}}
    '''{{desc | replace('
', '
    ')}}'''
```

Target Template:

```
{{bodies}}
```

Input Template:

```
I wrote the below code

{{bodies}}

What's a good function header?
```

Target Template:

```
{{decl}}
```

Input Template:

```
{{decl}}
```

Target Template:

```
"""{{desc | replace('
', '
    ') | replace("'", '')}}"""
{{bodies}}
```

## 1.10 Word Sense Disambiguation

### 1.10.1 pasinit/xlwic xlwic__en__zh

Dataset from Raganato et al. (2020). Used in training.

**Data Example**

| Key | Value |
| --- | --- |
| id | EN_1 |
| context__1 | We like to summer in the Mediterran... |
| context__2 | We summered in Kashmir. |
| target__word | summer |
| pos | V |
| target__word__location__1 | {'char_start': 11, 'char_end': 17} |
| target__word__location__2 | {'char_start': 3, 'char_end': 11} |
| language | EN |
| label | 1 |

**Prompts**

Input Template:

```
第 1 句: {{context_1}}
句子 2: {{context_2}}

确定单词 "{{target_word}}" 在两个句子中的使用是否相同。是还是不是？
```

Target Template:

```
{% if label != -1%}
{{answer_choices[label]}}
{% endif %}
```

Answer Choices Template:

```
不 ||| 是的
```

Input Template:

```
家庭作业

判断 "{{target_word}}" 这个词在以下两个句子中的含义是否相同。回答是或否。
{{context_1}}
{{context_2}}
```

Target Template:

```
{% if label != -1%}
{{answer_choices[label]}}
{% endif %}
```

Answer Choices Template:

```
不 ||| 是的
```

Input Template:

```
{{context_1}}
{{context_2}}
{{target_word}} 的类似意义？
```

Target Template:

```
{% if label != -1%}
{{answer_choices[label]}}
{% endif %}
```

Answer Choices Template:

```
不 ||| 是的
```

Input Template:

```
"{{target_word}}" 这个词有多种含义。第 1 句和第 2 句的意思相同吗？是还是不是？

第 1 句: {{context_1}}
句子 2: {{context_2}}
```

Target Template:

```
{% if label != -1%}
{{answer_choices[label]}}
{% endif %}
```

Answer Choices Template:

不 ||| 是的

Input Template:

```
确定以下两个句子中是否以相同的方式使用了单词 "{{target_word}}"。
{{context_1}}
{{context_2}}
```

Target Template:

```
{% if label != -1%}
{{answer_choices[label]}}
{% endif %}
```

Answer Choices Template:

不 ||| 是的

Input Template:

```
{{context_1}}
{{context_2}}
问题："{{target_word}}" 这个词在上面两个句子中的含义是否相同？
```

Target Template:

```
{% if label != -1%}
{{answer_choices[label]}}
{% endif %}
```

Answer Choices Template:

不 ||| 是的

Input Template:

```
"{{target_word}}" 这个词在这两个句子中是否具有相同的含义？是的，不是吗？
{{context_1}}
{{context_2}}
```

Target Template:

```
{% if label != -1%}
{{answer_choices[label]}}
{% endif %}
```

Answer Choices Template:

不 ||| 是的

Input Template:

```
"{{target_word}}" 这个词在这两个句子中是否具有相同的含义？
{{context_1}}
{{context_2}}
```

Target Template:

```
{% if label != -1%}
{{answer_choices[label]}}
{% endif %}
```

Answer Choices Template:

不 ||| 是的

Input Template:

```
句子 A: {{context_1}}
句子 B: {{context_2}}

"{{target_word}}" 在句子 A 和 B 中具有相似的含义。对还是错？
```

Target Template:

```
{% if label != -1%}
{{answer_choices[label]}}
{% endif %}
```

Answer Choices Template:

错误的 ||| 真的

Input Template:

```
{{context_1}}
{{context_2}}
问题："{{target_word}}" 这个词在上面两个句子中的含义是否相同？是的，不是吗？
```

Target Template:

```
{% if label != -1%}
{{answer_choices[label]}}
{% endif %}
```

Answer Choices Template:

```
不 ||| 是的
```

### 1.10.2 pasinit/xlwic xlwic_fr_fr

**Data Example**

| Key | Value |
| --- | --- |
| id | FR_1 |
| context_1 | L' éclaircie est généralement une co... |
| context_2 | Améliorations utiles. |
| target_word | amélioration |
| pos | N |
| target_word_location_1 | {'char_start': 41, 'char_end': 53} |
| target_word_location_2 | {'char_start': 0, 'char_end': 13} |
| language | FR |
| label | 1 |

**Prompts**

Input Template:

```
{{context_1}}
{{context_2}}
Sens similaire de {{target_word}} ?
```

Target Template:

```
{% if label != -1%}
{{answer_choices[label]}}
{% endif %}
```

Answer Choices Template:

```
Non ||| Oui
```

Input Template:

```
Devoirs

Décidez si le mot "{{target_word}}" est utilisé avec le même sens dans les deux
phrases suivantes. Répondez par oui ou non.
{{context_1}}
{{context_2}}
```

Target Template:

```
{% if label != -1%}
{{answer_choices[label]}}
{% endif %}
```

Answer Choices Template:

```
Non ||| Oui
```

Input Template:

```
Le mot "{{target_word}}" a-t-il le même sens dans ces deux phrases ? Oui Non?
{{context_1}}
{{context_2}}
```

Target Template:

```
{% if label != -1%}
{{answer_choices[label]}}
{% endif %}
```

Answer Choices Template:

```
Non ||| Oui
```

Input Template:

```
{{context_1}}
{{context_2}}
Question : Le mot '{{target_word}}' est-il utilisé dans le même sens dans les deux phrases ci-dessus ? Oui Non?
```

Target Template:

```
{% if label != -1%}
{{answer_choices[label]}}
{% endif %}
```

Answer Choices Template:

```
Non ||| Oui
```

Input Template:

```
Le mot "{{target_word}}" a plusieurs significations. A-t-il le même sens dans les phrases 1 et 2 ? Oui ou non?

Phrase 1 : {{context_1}}
Phrase 2 : {{context_2}}
```

Target Template:

```
{% if label != -1%}
{{answer_choices[label]}}
{% endif %}
```

Answer Choices Template:

```
Non ||| Oui
```

Input Template:

```
Phrase 1 : {{context_1}}
Phrase 2 : {{context_2}}

Déterminez si le mot "{{target_word}}" est utilisé dans le même sens dans les deux phrases. Oui ou non?
```

Target Template:

```
{% if label != -1%}
{{answer_choices[label]}}
{% endif %}
```

Answer Choices Template:

```
Non ||| Oui
```

Input Template:

```
Le mot "{{target_word}}" a-t-il le même sens dans ces deux phrases ?
{{context_1}}
{{context_2}}
```

Target Template:

```
{% if label != -1%}
{{answer_choices[label]}}
{% endif %}
```

Answer Choices Template:

```
Non ||| Oui
```

Input Template:

```
Phrase A : {{context_1}}
Phrase B : {{context_2}}

"{{target_word}}" a une signification similaire dans les phrases A et B. Vrai ou
faux ?
```

Target Template:

```
{% if label != -1%}
{{answer_choices[label]}}
{% endif %}
```

Answer Choices Template:

```
Faux ||| Vrai
```

Input Template:

```
Déterminez si le mot '{{target_word}}' est utilisé de la même manière dans les
deux phrases ci-dessous.
{{context_1}}
{{context_2}}
```

Target Template:

```
{% if label != -1%}
{{answer_choices[label]}}
{% endif %}
```

Answer Choices Template:

```
Non ||| Oui
```

Input Template:

```
{{context_1}}
{{context_2}}
Question : Le mot '{{target_word}}' est-il utilisé dans le même sens dans les deux
phrases ci-dessus ?
```

Target Template:

```
{% if label != -1%}
{{answer_choices[label]}}
{% endif %}
```

Answer Choices Template:

```
Non ||| Oui
```

## 1.11 Paraphrase Identification

### 1.11.1 PAWS-X EN

Dataset from Yang et al. (2019). Used in training.

**Data Example**

| Key | Value |
| --- | --- |
| id | 2 |
| sentence1 | The NBA season of 1975 -- 76 was th... |
| sentence2 | The 1975 -- 76 season of the Nation... |
| label | 1 |

**Prompts**

**Notes:** Generalized prompt format, task_description-input.

Input Template:

```
Determine if the following two sentences paraphrase each other or not.
Sent 1: {{sentence1}}
Sent 2: {{sentence2}}
```

Target Template:

```
{{answer_choices[label]}}
```

Answer Choices Template:

```
No ||| Yes
```

**Notes:** Natural question.

Input Template:

```
Sentence 1: {{sentence1}}
Sentence 2: {{sentence2}}
Question: Do Sentence 1 and Sentence 2 express the same meaning? Yes or No?
```

Target Template:

```
{{answer_choices[label]}}
```

Answer Choices Template:

```
No ||| Yes
```

**Notes:** Generalized prompt format, context-question without any label.

Input Template:

```
{{sentence1}}
Is that a paraphrase of the following sentence?
{{sentence2}}?
```

Target Template:

```
{{answer_choices[label]}}
```

Answer Choices Template:

```
No ||| Yes
```

**Notes:** Natural Question without label.

Input Template:

```
Sentence 1: {{sentence1}}
Sentence 2: {{sentence2}}
Question: Can we rewrite Sentence 1 to Sentence 2?
```

Target Template:

```
{{answer_choices[label]}}
```

Answer Choices Template:

```
No ||| Yes
```

**Notes:** Generalized prompt format, context-question.

Input Template:

```
{{sentence1}}
Is that a paraphrase of the following sentence?
{{sentence2}}?
Yes or No.
```

Target Template:

```
{{answer_choices[label]}}
```

Answer Choices Template:

```
No ||| Yes
```

**Notes:** Concatenation of sentence 1 and sentence 2.

Input Template:

```
Sentence 1: {{sentence1}}
Sentence 2: {{sentence2}}
Question: Does Sentence 1 paraphrase Sentence 2? Yes or No?
```

Target Template:

```
{{answer_choices[label]}}
```

Answer Choices Template:

```
No ||| Yes
```

Note: the prompt does not correspond to the original task intended by the dataset authors.

**Notes:** Create a generative paraphrase task.

Input Template:

```
{% if label == 1 %}
Paraphrase the sentence: {{sentence1}}
```

Target Template:

```
{{sentence2}}
{% endif %}
```

**Notes:** Concatenation of sentence 1 and sentence 2 without any label.

Input Template:

```
Sentence 1: {{sentence1}}
Sentence 2: {{sentence2}}
Question: Does Sentence 1 paraphrase Sentence 2?
```

Target Template:

```
{{answer_choices[label]}}
```

Answer Choices Template:

```
No ||| Yes
```

**Notes:** Natural question without label.

Input Template:

```
Sentence 1: {{sentence1}}
Sentence 2: {{sentence2}}
Question: Do Sentence 1 and Sentence 2 express the same meaning?
```

Target Template:

```
{{answer_choices[label]}}
```

Answer Choices Template:

```
No ||| Yes
```

Prompt from Brown et al. (2020) **Notes:** ANLI prompt format from Table G7 in the GPT3 paper Brown et al. (2020)

Input Template:

```
{{sentence1}} Question: {{sentence2}} True or False?
```

Target Template:

```
{{answer_choices[label]}}
```

Answer Choices Template:

```
False ||| True
```

**Notes:** Natural Question.

Input Template:

```
Sentence 1: {{sentence1}}
Sentence 2: {{sentence2}}
Question: Can we rewrite Sentence 1 to Sentence 2? Yes or No?
```

Target Template:

```
{{answer_choices[label]}}
```

Answer Choices Template:

```
No ||| Yes
```

Prompt from Brown et al. (2020) **Notes:** ANLI prompt format from Table G7 in the GPT3 paper Brown et al. (2020). Additionally added task information without any label.

Input Template:

```
{{sentence1}} Question: {{sentence2}} Paraphrase or not?
```

Target Template:

```
{{answer_choices[label]}}
```

Answer Choices Template:

```
No ||| Yes
```

### 1.11.2 PAWS-X ES

**Data Example**

| Key | Value |
| --- | --- |
| id | 2 |
| sentence1 | La temporada de la NBA de 1975: 76 ... |
| sentence2 | La temporada 1975 - 76 de la Asocia... |
| label | 1 |

**Prompts**

Input Template:

```
Oración 1: {{sentence1}}
Oración 2: {{sentence2}}
Pregunta: ¿La oración 1 parafrasea la oración 2? ¿Si o no?
```

Target Template:

```
{{answer_choices[label]}}
```

Answer Choices Template:

```
No ||| Sí
```

Input Template:

```
{{sentence1}} Pregunta: {{sentence2}} ¿Parafrasear o no?
```

Target Template:

```
{{answer_choices[label]}}
```

Answer Choices Template:

```
No ||| Sí
```

Input Template:

```
{{sentence1}}
¿Es una paráfrasis de la siguiente oración?
{{sentence2}}?
Si o no.
```

Target Template:

```
{{answer_choices[label]}}
```

Answer Choices Template:

```
No ||| Sí
```

Input Template:

```
Oración 1: {{sentence1}}
Oración 2: {{sentence2}}
Pregunta: ¿La Oración 1 y la Oración 2 expresan el mismo significado?
```

Target Template:

```
{{answer_choices[label]}}
```

Answer Choices Template:

```
No ||| Sí
```

Input Template:

```
{% if label == 1 %}
Parafrasea la oración: {{sentence1}}
```

Target Template:

```
{{sentence2}}
{% endif %}
```

Input Template:

```
{{sentence1}} Pregunta: {{sentence2}} ¿Verdadero o falso?
```

Target Template:

```
{{answer_choices[label]}}
```

Answer Choices Template:

```
Falso ||| Verdadero
```

Input Template:

```
Oración 1: {{sentence1}}
Oración 2: {{sentence2}}
Pregunta: ¿La oración 1 parafrasea la oración 2?
```

Target Template:

```
{{answer_choices[label]}}
```

Answer Choices Template:

```
No ||| Sí
```

Input Template:

```
Determina si las siguientes dos oraciones se parafrasean entre sí o no.
Enviado 1: {{sentence1}}
Enviado 2: {{sentence2}}
```

Target Template:

```
{{answer_choices[label]}}
```

Answer Choices Template:

```
No ||| Sí
```

Input Template:

```
Oración 1: {{sentence1}}
Oración 2: {{sentence2}}
Pregunta: ¿La Oración 1 y la Oración 2 expresan el mismo significado? ¿Si o no?
```

Target Template:

```
{{answer_choices[label]}}
```

Answer Choices Template:

```
No ||| Sí
```

Input Template:

```
Oración 1: {{sentence1}}
Oración 2: {{sentence2}}
Pregunta: ¿Podemos reescribir la Oración 1 a la Oración 2? ¿Si o no?
```

Target Template:

```
{{answer_choices[label]}}
```

Answer Choices Template:

```
No ||| Sí
```

Input Template:

```
{{sentence1}}
¿Es una paráfrasis de la siguiente oración?
{{sentence2}}?
```

Target Template:

```
{{answer_choices[label]}}
```

Answer Choices Template:

```
No ||| Sí
```

Input Template:

```
Oración 1: {{sentence1}}
Oración 2: {{sentence2}}
Pregunta: ¿Podemos reescribir la Oración 1 a la Oración 2?
```

Target Template:

```
{{answer_choices[label]}}
```

Answer Choices Template:

```
No ||| Sí
```

### 1.12 Sentence Completion

#### 1.12.1 xcopa vi

Dataset from Ponti et al. (2020). Used in evaluation.

**Data Example**

| Key | Value |
|---|---|
| premise | Cô gái tìm thấy con bọ trong ngũ cốc... |
| choice1 | Cô đổ sữa vào bát. |
| choice2 | Cô mất cảm giác ngon miệng. |
| question | effect |
| label | 1 |
| idx | 1 |
| changed | False |

**Prompts**

Input Template:

```
{{ premise }}

Tôi đang lưỡng lự giữa hai lựa chọn. Giúp tôi chọn nguyên nhân {% if question ==
"cause" %} có khả năng xảy ra cao hơn: {% else %} effect: {% endif %}
- {{choice1}}
- {{choice2}}
```

Target Template:

```
{% if label != -1 %} {{ answer_choices[label] }} {% endif %}
```

Answer Choices Template:

```
{{choice1}} ||| {{choice2}}
```

Input Template:

```
{{ premise }}

Lựa chọn tốt nhất là gì?
- {{choice1}}
- {{choice2}}

Chúng tôi đang tìm kiếm {% if question == "cause" %} một nguyên nhân {% else %}
một ảnh hưởng {% endif %}
```

Target Template:

```
{% if label != -1 %} {{answer_choices[label]}} {% endif %}
```

Answer Choices Template:

```
{{choice1}} ||| {{choice2}}
```

Input Template:

```
{{ premise }} {% if question == "cause" %} Điều này xảy ra vì ... {% else %} Do đó
... {% endif %}
Giúp tôi chọn tùy chọn hợp lý hơn:
- {{choice1}}
- {{choice2}}
```

Target Template:

```
{% if label != -1 %} {{ answer_choices[label] }} {% endif %}
```

Answer Choices Template:

```
{{choice1}} ||| {{choice2}}
```

Input Template:

```
"{{ answer_choices[0] }}" hay "{{ answer_choices[1] }}"? {{ premise }} {% if
question == "cause" %} bởi vì {% else %} nên {% endif %}
```

Target Template:

```
{% if label != -1 %} {{ answer_choices[label] }} {% endif %}
```

Answer Choices Template:

```
{{choice1 }} ||| {{choice2}}
```

Input Template:

```
{{ premise }}

Chọn nguyên nhân {% if question == "cause" %} hợp lý nhất: {% else %} effect: {%
endif %}
- {{choice1}}
- {{choice2}}
```

Target Template:

```
{% if label != -1 %} {{ answer_choices[label] }} {% endif %}
```

Answer Choices Template:

```
{{choice1}} ||| {{choice2}}
```

### 1.12.2 XCOPA ZH

**Data Example**

| Key | Value |
|---|---|
| premise | 这个女孩在麦片粥中发现了一个虫子。 |
| choice1 | 她向碗里倒了牛奶。 |
| choice2 | 她没了食欲。 |
| question | effect |
| label | 1 |
| idx | 1 |
| changed | False |

**Prompts**

Input Template:

```
{{ premise }} {% if question == "cause" %} 这是因为... {% else %} 结果... {% endif
%}
帮助我选择更合理的选项:
- {{choice1}}
- {{choice2}}
```

Target Template:

```
{% if label != -1 %}{{ answer_choices[label] }}{%endif%}
```

Answer Choices Template:

```
{{choice1}} ||| {{choice2}}
```

Input Template:

```
{{ premise }}

选择最合理的 {% if question == "cause" %} 原因: {% else %} 效果: {% endif %}
- {{choice1}}
- {{choice2}}
```

Target Template:

```
{% if label != -1 %}{{ answer_choices[label] }}{%endif%}
```

Answer Choices Template:

```
{{choice1}} ||| {{choice2}}
```

Input Template:

```
"{{ answer_choices[0] }}" 还是 "{{ answer_choices[1] }}"? {{ premise }} {% if question == "cause" %} 因为 {% else %} 所以 {% endif %}
```

Target Template:

```
{% if label != -1 %}{{ answer_choices[label] }}{% endif %}
```

Answer Choices Template:

```
{{choice1 }} ||| {{choice2}}
```

Input Template:

```
{{ premise }}

最好的选择是什么？
- {{choice1}}
- {{choice2}}

我们正在寻找 {% if question == "cause" %} 一个原因 {% else %} 一个结果 {% endif %}
```

Target Template:

```
{% if label != -1 %}{{answer_choices[label]}}{%endif%}
```

Answer Choices Template:

```
{{choice1}} ||| {{choice2}}
```

Input Template:

```
{{ premise }}
我在两个选项之间犹豫不决。帮我选择更有可能的 {% if question == "cause" %} 原因：{% else %} 效果：{% endif %}
- {{choice1}}
- {{choice2}}
```

Target Template:

```
{% if label != -1 %}{{ answer_choices[label] }}{%endif%}
```

Answer Choices Template:

```
{{choice1}} ||| {{choice2}}
```

Input Template:

```
{{ premise }}
我正在考虑两个选项。请帮我最有可能的{% if question == "cause" %}导因：{% else %}后果：{% endif %}
- {{choice1}}
- {{choice2}}
```

Target Template:

```
{% if label != -1 %}{{ answer_choices[label] }}{%endif%}
```

Answer Choices Template:

```
{{choice1}} ||| {{choice2}}
```

Input Template:

```
{{ premise }} {% if question == "cause" %}这个会发生是因为... {% else %}结果是... {% endif %}
帮我挑选合适的选项:
- {{choice1}}
- {{choice2}}
```

Target Template:

```
{% if label != -1 %}{{ answer_choices[label] }}{%endif%}
```

Answer Choices Template:

```
{{choice1}} ||| {{choice2}}
```

**Notes:** Adapted from Perez et al. (2021) and Schick and Schütze (2020).

Input Template:

```
"{{ answer_choices[0] }}" 还是"{{ answer_choices[1] }}"? {{ premise }} {% if question == "cause" %}因为{% else %}所以{% endif %}
```

Target Template:

```
{% if label != -1 %}{{ answer_choices[label] }}{% endif %}
```

Answer Choices Template:

```
{{choice1}} ||| {{choice2}}
```

Input Template:

```
{{ premise }}

哪个是最好的答案?
- {{choice1}}
- {{choice2}}

我们正在考虑{% if question == "cause" %}起因{% else %}后果 {% endif %}
```

Target Template:

```
{% if label != -1 %}{{answer_choices[label]}}{%endif%}
```

Answer Choices Template:

```
{{choice1}} ||| {{choice2}}
```

Input Template:

```
{{ premise }}

请选择最贴切的答案:  {% if question == "cause" %}导因:{% else %}结果: {% endif %}
- {{choice1}}
- {{choice2}}
```

Target Template:

```
{% if label != -1 %}{{ answer_choices[label] }}{%endif%}
```

Answer Choices Template:

```
{{choice1}} ||| {{choice2}}
```

## 1.13 Natural Language Inference

### 1.13.1 XNLI EN

Dataset from Conneau et al. (2018). Used in evaluation.

**Data Example**

| Key | Value |
|---|---|
| premise | you know during the season and i gu... |
| hypothesis | You lose the things to the followin... |
| label | 0 |

**Prompts**

**Notes:** Sanh et al. (2022)

Input Template:

```
Take the following as truth: {{premise}}
Then the following statement: "{{hypothesis}}" is {{"true"}}, {{"false"}}, or
{{"inconclusive"}}?
```

Target Template:

```
{{ answer_choices[label] }}
```

Answer Choices Template:

```
True ||| Inconclusive ||| False
```

**Notes:** Sanh et al. (2022)

Input Template:

```
{{premise}}

Question: Does this imply that "{{hypothesis}}"? Yes, no, or maybe?
```

Target Template:

```
{{answer_choices[label]}}
```

Answer Choices Template:

```
Yes ||| Maybe ||| No
```

**Notes:** Same as reported in Figure G7 of Brown et al. (2020), except that there is no task identifying tokens like "anli R1: ".

Input Template:

```
{{premise}}
Question: {{hypothesis}} True, False, or Neither?
```

Target Template:

```
{{ answer_choices[label] }}
```

Answer Choices Template:

```
True ||| Neither ||| False
```

**Notes:** Sanh et al. (2022)

Input Template:

```
Given that {{premise}} Does it follow that {{hypothesis}} Yes, no, or maybe?
```

Target Template:

```
{{ answer_choices[label] }}
```

Answer Choices Template:

```
Yes ||| Maybe ||| No
```

**Notes:** Adapted from the BoolQ prompts in Schick and Schütze (2020).

Input Template:

```
{{premise}} Based on the previous passage, is it true that "{{hypothesis}}"? Yes, no, or maybe?
```

Target Template:

```
{{ answer_choices[label] }}
```

Answer Choices Template:

```
Yes ||| Maybe ||| No
```

**Notes:** Webson and Pavlick (2021)

Input Template:

```
Given {{premise}} Is it guaranteed true that "{{hypothesis}}"? Yes, no, or maybe?
```

Target Template:

```
{{ answer_choices[label] }}
```

Answer Choices Template:

```
Yes ||| Maybe ||| No
```

**Notes:** Webson and Pavlick (2021)

Input Template:

```
Given {{premise}} Should we assume that "{{hypothesis}}" is true? Yes, no, or maybe?
```

Target Template:

```
{{ answer_choices[label] }}
```

Answer Choices Template:

```
Yes ||| Maybe ||| No
```

**Notes:** Sanh et al. (2022)

Input Template:

```
Given that {{premise}} Therefore, it must be true that "{{hypothesis}}"? Yes, no, or maybe?
```

Target Template:

```
{{ answer_choices[label] }}
```

Answer Choices Template:

```
Yes ||| Maybe ||| No
```

**Notes:** Webson and Pavlick (2021)

Input Template:

```
Suppose {{premise}} Can we infer that "{{hypothesis}}"? Yes, no, or maybe?
```

Target Template:

```
{{ answer_choices[label] }}
```

Answer Choices Template:

```
Yes ||| Maybe ||| No
```

**Notes:** Webson and Pavlick (2021)

Input Template:

```
{{premise}} Are we justified in saying that "{{hypothesis}}"? Yes, no, or maybe?
```

Target Template:

```
{{ answer_choices[label] }}
```

Answer Choices Template:

```
Yes ||| Maybe ||| No
```

**Notes:** Sanh et al. (2022)

Input Template:

```
{{premise}} Based on that information, is the claim: "{{hypothesis}}" {{"true"}},
{{"false"}}, or {{"inconclusive"}}?
```

Target Template:

```
{{ answer_choices[label] }}
```

Answer Choices Template:

```
True ||| Inconclusive ||| False
```

**Notes:** Sanh et al. (2022)

Input Template:

```
{{premise}}

Keeping in mind the above text, consider: {{hypothesis}} Is this {{"always"}},
{{"sometimes"}}, or {{"never"}} correct?
```

Target Template:

```
{{ answer_choices[label] }}
```

Answer Choices Template:

```
Always ||| Sometimes ||| Never
```

**Notes:** Sanh et al. (2022)

Input Template:

```
Suppose it's true that {{premise}} Then, is "{{hypothesis}}" {{"always"}},
{{"sometimes"}}, or {{"never"}} true?
```

Target Template:

```
{{ answer_choices[label] }}
```

Answer Choices Template:

```
Always ||| Sometimes ||| Never
```

**Notes:** Sanh et al. (2022)

Input Template:

```
Assume it is true that {{premise}}

Therefore, "{{hypothesis}}" is {{"guaranteed"}}, {{"possible"}}, or
{{"impossible"}}?
```

Target Template:

```
{{ answer_choices[label] }}
```

Answer Choices Template:

```
Guaranteed ||| Possible ||| Impossible
```

**Notes:** Adapted from Williams et al. (2018) instructions to crowdsourcing workers.

Input Template:

```
{{premise}} Using only the above description and what you know about the world,
"{{hypothesis}}" is definitely correct, incorrect, or inconclusive?
```

Target Template:

```
{{ answer_choices[label] }}
```

Answer Choices Template:

```
Correct ||| Inconclusive ||| Incorrect
```

### 1.13.2 XNLI ES

**Data Example**

| Key | Value |
|---|---|
| premise | Usted sabe durante la temporada y s... |
| hypothesis | Pierdes las cosas al siguiente nive... |
| label | 0 |

**Prompts**

#### 1.13.2.1 Human-translated prompts

**Notes:** Same as reported in Figure G7 of Brown et al. (2020), except that there is no task identifying tokens like "anli R1: ".

Input Template:

```
{{premise}}
Pregunta: {{hypothesis}} Verdadero, Falso, o Ninguno?
```

Target Template:

```
{{ answer_choices[label] }}
```

Answer Choices Template:

```
Verdadero ||| Ninguno ||| Falso
```

**Notes:** Webson and Pavlick (2021)

Input Template:

```
Supongamos {{premise}} Podemos inferir que "{{hypothesis}}"? Si, no, o tal vez?
```

Target Template:

```
{{ answer_choices[label] }}
```

Answer Choices Template:

```
Sí ||| Tal vez ||| No
```

**Notes:** Webson and Pavlick (2021)

Input Template:

```
{{premise}} Estamos justificados en decir que "{{hypothesis}}"? Si, no, o tal vez?
```

Target Template:

```
{{ answer_choices[label] }}
```

Answer Choices Template:

```
Sí ||| Tal vez ||| No
```

**Notes:** Sanh et al. (2022)

Input Template:

```
Supongamos que es cierto que {{premise}}

por lo tanto, "{{hypothesis}}" es {{"garantizado"}}, {{"posible"}}, o
{{"imposible"}}?
```

Target Template:

```
{{ answer_choices[label] }}
```

Answer Choices Template:

```
Garantizado ||| Posible ||| Imposible
```

**Notes:** Adapted from Williams et al. (2018) instructions to crowdsourcing workers.

Input Template:

```
{{premise}} Usando solo la descripción anterior y lo que sabe sobre el mundo,
"{{hypothesis}}" es definitivamente correcto, incorrecto o no concluyente?
```

Target Template:

```
{{ answer_choices[label] }}
```

Answer Choices Template:

```
Correcto ||| No concluyente ||| Incorrecto
```

### 1.13.2.2 Machine-translated prompts

Input Template:

```
{{premise}} ¿Estamos justificados al decir que &quot;{{hypothesis}}&quot;? ¿Sí, no
o tal vez?
```

Target Template:

```
{{ answer_choices[label] }}
```

Answer Choices Template:

```
Sí ||| Quizás ||| No
```

Input Template:

```
{{premise}} Pregunta: {{hypothesis}} ¿Verdadero, falso o ninguno?
```

Target Template:

```
{{ answer_choices[label] }}
```

Answer Choices Template:

```
Verdadero ||| Ninguno de los dos ||| Falso
```

Input Template:

```
{{premise}} Usando solo la descripción anterior y lo que sabe sobre el mundo,
&quot;{{hypothesis}}&quot; es definitivamente correcta, incorrecta o no
concluyente.
```

Target Template:

```
{{ answer_choices[label] }}
```

Answer Choices Template:

```
Correcto ||| Poco concluyente ||| Incorrecto
```

Input Template:

```
Supongamos {{premise}} ¿Podemos inferir que &quot;{{hypothesis}}&quot;? ¿Sí, no o
tal vez?
```

Target Template:

```
{{ answer_choices[label] }}
```

Answer Choices Template:

```
Sí ||| Quizás ||| No
```

Input Template:

```
Supongamos que es cierto que {{premise}} Por lo tanto, &quot;{{hypothesis}}&quot;
es {{"guaranteed"}}, {{"possible"}} o {{"impossible"}}.
```

Target Template:

```
{{ answer_choices[label] }}
```

Answer Choices Template:

```
garantizado ||| Posible ||| Imposible
```


    
    
\end{document}

%% file: promptlangl1.tex
\begin{table}[htbp]
    \begin{center}
        \resizebox{1\linewidth}{!}{
            \begin{tabular}{@{}ll|cc|cc@{}}
\toprule
Task & Prompt & \multicolumn{4}{c}{Average accuracy} \\
     &        & \small{BLOOMZ} & \small{BLOOMZ-MT} & \small{mT0-13B} & \small{mT0-13B-MT} \\
\midrule
XNLI & EN & \textbf{52.99} & 49.01 & 48.24 & \textbf{51.29} \\
 & MT & 37.56 & \textbf{41.16} & 39.31 & \textbf{41.66} \\
 & HT & 40.4 & \textbf{43.88} & 44.95 & \textbf{46.87} \\
\midrule
XCOPA & EN & 72.52 & \textbf{73.24} & \textbf{81.4} & 80.36 \\
 & MT & 70.04 & \textbf{71.84} & \textbf{81.16} & 79.64 \\
\midrule
XStoryCloze & EN & \textbf{81.73} & 81.39 & 81.99 & \textbf{82.3} \\
 & MT & 80.89 & \textbf{81.76} & \textbf{83.37} & 82.86 \\
\midrule
XWinograd & EN & \textbf{60.07} & 59.15 & 70.49 & \textbf{73.24} \\
 & MT & 58.48 & \textbf{60.14} & 66.89 & \textbf{72.33} \\
\bottomrule
            \end{tabular}
        }
    \caption{Comparison between EN (English), MT (machine-translated) and HT (human-translated) prompts for 176B BLOOMZ and 13B mT0 models finetuned on either only English or English and machine-translated multilingual prompts (-MT).}
    \label{tab:promptlangl1}
    \end{center}
\end{table}

%% file: contributions.tex
{\setlength{\parindent}{0cm}
\textbf{Niklas Muennighoff} evaluated all models, created xP3 and wrote most of the paper.

\textbf{Niklas Muennighoff, Thomas Wang, Lintang Sutawika, Adam Roberts and Hailey Schoelkopf} wrote the training and evaluation code.

\textbf{Niklas Muennighoff and Adam Roberts} trained the models.

\textbf{Niklas Muennighoff, Teven Le Scao, Hailey Schoelkopf, Zheng-Xin Yong, Thomas Wang, Khalid Almubarak, Alham Fikri Aji, M Saiful Bari and Zaid Alyafeai} contributed prompts or datasets.

\textbf{Lintang Sutawika, Stella Biderman, Zheng-Xin Yong, Khalid Almubarak, M Saiful Bari and Albert Webson} initiated the project.

\textbf{Sheng Shen} conducted the contamination analysis.

\textbf{Samuel Albanie} wrote the prompt appendix.

\textbf{Thomas Wang and Zheng-Xin Yong} converted checkpoints.

\textbf{Colin Raffel, Thomas Wang, Teven Le Scao, M Saiful Bari, Edward Raff and Dragomir Radev} advised the project.

\textbf{Niklas Muennighoff, Lintang Sutawika, Teven Le Scao, Colin Raffel, Stella Biderman, Alham Fikri Aji, Adam Roberts, Samuel Albanie, Sheng Shen, M Saiful Bari, Albert Webson, Xiangru Tang, Dragomir Radev and Edward Raff} contributed to the paper.
}

%% file: promptlangl2.tex
\begin{table}[htbp]
    \small
    \begin{center}
            \begin{tabular}{ll|cc|cc}
\toprule
Task & Prompt & \multicolumn{4}{c}{Average accuracy} \\
     &        & BLOOMZ & BLOOMZ-MT & mT0-13B & mT0-13B-MT \\
\midrule
XNLI & EN & \textbf{45.65} & 43.2 & 48.52 & \textbf{51.33} \\
 & MT & \textbf{36.48} & 35.67 & \textbf{41.86} & 39.78 \\
\midrule
XCOPA & EN & \textbf{54.27} & 53.67 & \textbf{72.67} & 71.6 \\
 & MT & \textbf{53.2} & 53.0 & \textbf{71.57} & 70.87\\
\midrule
XStoryCloze & EN & \textbf{61.59} & 61.36 & 79.31 & \textbf{80.13} \\
 & MT & \textbf{60.5} & 59.91 & 80.21 & \textbf{80.28}\\
\midrule
XWinograd & EN & \textbf{55.98} & 54.54 & 70.81 & \textbf{72.0} \\
 & MT & \textbf{53.11} & 52.46 & 67.86 & \textbf{70.45} \\
\bottomrule
            \end{tabular}
    \caption{Comparison between EN (English) and MT (machine-translated) prompts for 176B BLOOMZ and 13B mT0 models finetuned on either only English or English and machine-translated multilingual prompts (-MT). For BLOOMZ the evaluation languages averaged are never intentionally seen, such as Japanese and Russian for XWinograd (see Figure \ref{fig:langgen}). For mT0 the evaluation languages are only seen during pretraining.}
    \label{tab:promptlangl2}
    \end{center}
\end{table}

%% file: allres.tex
\begin{table*}[ht]
    \centering
    \begin{minipage}{\pdfpagewidth}
    \resizebox{0.95\textwidth}{!}{
    \footnotesize
\begin{tabular}{llllll|c|cccccc|c|cc|ccccccc|cccccccccc}
\toprule
\multicolumn{6}{c|}{} & \multicolumn{7}{c|}{Pretrained} & \multicolumn{20}{c}{Pretrained + Multitask finetuned} \\
\cmidrule{7-13} \cmidrule{14-33}
Task & Dataset & Config & Split & Prompt & Metric & XGLM-7.5B & BLOOM-560M & BLOOM-1.1B & BLOOM-1.7B & BLOOM-3B & BLOOM-7.1B & BLOOM & T0-11B & mTk-Instruct-3.7B & mTk-Instruct-13B & mT0-300M & mT0-560M & mT0-1.2B & mT0-3.7B & mT0-13B & mT0-13B-MT & mT0-13B-P3 & BLOOMZ-560M & BLOOMZ-1.1B & BLOOMZ-1.7B & BLOOMZ-3B & BLOOMZ-7.1B & BLOOMZ-7.1B-MT & BLOOMZ-7.1B-P3 & BLOOMZ & BLOOMZ-MT & BLOOMZ-P3 \\
\midrule
\midrule
Coref. resolution & winogrande & xl & validation & EN & Median acc. & 49.25 & 49.88 & 50.99 & 49.57 & 49.96 & 49.41 & 48.62 & 60.46 & 50.99 & 52.33 & 49.57 & 51.62 & 50.51 & 52.01 & 62.27 & 62.51 & 56.91 & 49.80 & 51.07 & 50.75 & 51.78 & 55.41 & 55.88 & 51.78 & 58.41 & 58.64 & 55.64\\
Coref. resolution & winogrande & xl & validation & EN & Max acc. & 50.12 & 50.99 & 51.62 & 50.91 & 51.46 & 50.91 & 49.64 & 63.61 & 51.14 & 54.54 & 50.51 & 53.28 & 51.78 & 52.49 & 63.38 & 62.67 & 58.56 & 52.41 & 52.33 & 51.14 & 53.67 & 55.80 & 56.51 & 54.06 & 59.27 & 59.98 & 57.06\\
Coref. resolution & xwinograd & en & test & EN & Median acc. & 50.88 & 50.62 & 51.10 & 50.67 & 50.97 & 50.15 & 50.28 & 62.75 & 52.22 & 52.77 & 50.11 & 51.01 & 52.30 & 57.94 & 79.91 & 81.33 & 59.87 & 50.24 & 50.15 & 52.09 & 54.84 & 60.09 & 59.31 & 52.26 & 67.87 & 64.73 & 59.74\\
Coref. resolution & xwinograd & en & test & EN & Max acc. & 51.61 & 51.53 & 51.57 & 51.66 & 51.70 & 50.71 & 51.27 & 70.71 & 53.12 & 60.82 & 51.31 & 51.40 & 54.80 & 61.89 & 81.29 & 83.31 & 70.71 & 51.01 & 50.49 & 56.34 & 59.23 & 66.02 & 65.76 & 53.72 & 69.08 & 69.33 & 60.65\\
\bottomrule
            \end{tabular}
            }
    \caption{Evaluation results. Results per prompt can be found at \url{https://huggingface.co/datasets/bigscience/evaluation-results}
    \label{tab:allres}
    }
\end{minipage}
\end{table*}

%% file: valres.tex
\begin{table*}[ht]
    \centering
    \begin{minipage}{\pdfpagewidth}
    \resizebox{0.9\textwidth}{!}{
    \footnotesize
            \begin{tabular}{llllll|ccccc|ccccccc}
\toprule
Task & Dataset & Config & Split & Prompt & Metric & mT0-300M & mT0-560M & mT0-1.2B & mT0-3.7B & mT0-13B & BLOOMZ-560M & BLOOMZ-1.1B & BLOOMZ-1.7B & BLOOMZ-3B & BLOOMZ-7.1B & BLOOMZ \\
\midrule
\midrule
Extractive QA & craigslist\_bargains & bargains & validation & EN & Median acc. & 30.49 & 23.95 & 22.61 & 39.61 & 25.96 & 38.94 & 47.99 & 28.14 & 22.86 & 46.48 & 26.47\\
Extractive QA & craigslist\_bargains & bargains & validation & EN & Max acc. & 49.41 & 28.14 & 31.32 & 50.92 & 40.54 & 72.53 & 72.36 & 46.90 & 31.32 & 60.47 & 51.76\\
\midrule
Grammar Correction & blimp\_adjunct & island & validation & EN & Median acc. & 50.40 & 51.60 & 51.80 & 53.80 & 55.10 & 51.60 & 52.30 & 50.60 & 49.20 & 49.90 & 49.80\\
Grammar Correction & blimp\_adjunct & island & validation & EN & Max acc. & 50.90 & 57.00 & 58.00 & 59.10 & 56.80 & 77.10 & 60.90 & 62.30 & 59.90 & 57.60 & 51.60\\
Grammar Correction & glue & cola & validation & EN & Median acc. & 30.97 & 38.26 & 56.57 & 35.19 & 45.83 & 31.26 & 57.81 & 31.16 & 31.35 & 33.27 & 44.58\\
Grammar Correction & glue & cola & validation & EN & Max acc. & 64.33 & 51.01 & 62.80 & 47.17 & 58.29 & 41.71 & 67.98 & 46.40 & 65.39 & 56.86 & 63.37\\
\midrule
Multiple-Choice QA & aqua\_rat & raw & validation & EN & Median acc. & 27.95 & 25.20 & 24.80 & 20.47 & 16.14 & 19.29 & 22.83 & 22.05 & 22.44 & 24.41 & 27.56\\
\bottomrule
            \end{tabular}
            }
    \caption{Evaluation results on validation datasets used for checkpoint selection. Results of the chosen checkpoint are shown. Results per prompt can be found at \url{https://huggingface.co/datasets/bigscience/evaluation-results}
    \label{tab:valres}
    }
\end{minipage}
\end{table*}

%% file: prompt_appendix_nocite.tex
\nocite{Artetxe:etal:2019}
\nocite{DBLP:conf/iclr/HellendoornSSMB20}
\nocite{DBLP:journals/corr/abs-2001-07676}
\nocite{N18-1101}
\nocite{austin2021program}
\nocite{cmrc2018-dataset}
\nocite{conneau2018xnli}
\nocite{goyal2021flores}
\nocite{gpt3}
\nocite{guzman2019two}
\nocite{hasan-etal-2021-xl}
\nocite{hendrycksapps2021}
\nocite{huggingface:dataset}
\nocite{kim-etal-2021-bisect}
\nocite{ladhak-wiki-2020}
\nocite{lewis2019mlqa}
\nocite{nllb2022}
\nocite{pawsx2019emnlp}
\nocite{ponti2020xcopa}
\nocite{raganato-etal-2020-xl-wic}
\nocite{roemmele2011choice}
\nocite{sanh2022multitask}
\nocite{sun2020investigating}
\nocite{tiedemann-2020-tatoeba}
\nocite{tikhonov2021heads}
\nocite{true-zero-shot}
\nocite{webson-pavlick-2021}
\nocite{xu2020clue}
\nocite{zhu2022xlcost}

%% file: acl2023.bbl
\begin{thebibliography}{119}
\expandafter\ifx\csname natexlab\endcsname\relax\def\natexlab#1{#1}\fi

\bibitem[{hug(2018)}]{huggingface:dataset}
 2018.
\newblock \href {http://arxiv.org/abs/1908.09804} {Neural code search
  evaluation dataset}.
\newblock page arXiv:1908.09804 [cs.SE].

\bibitem[{lad(2020)}]{ladhak-wiki-2020}
 2020.
\newblock \href {https://arxiv.org/abs/2010.03093} {Wikilingua: A new benchmark
  dataset for multilingual abstractive summarization}.
\newblock \emph{arXiv preprint arXiv:2010.03093}.

\bibitem[{Allal et~al.(2023)Allal, Li, Kocetkov, Mou, Akiki, Ferrandis,
  Muennighoff, Mishra, Gu, Dey et~al.}]{allal2023santacoder}
Loubna~Ben Allal, Raymond Li, Denis Kocetkov, Chenghao Mou, Christopher Akiki,
  Carlos~Munoz Ferrandis, Niklas Muennighoff, Mayank Mishra, Alex Gu, Manan
  Dey, et~al. 2023.
\newblock Santacoder: don't reach for the stars!
\newblock \emph{arXiv preprint arXiv:2301.03988}.

\bibitem[{Aribandi et~al.(2021)Aribandi, Tay, Schuster, Rao, Zheng, Mehta,
  Zhuang, Tran, Bahri, Ni, Gupta, Hui, Ruder, and Metzler}]{ExT5}
Vamsi Aribandi, Yi~Tay, Tal Schuster, Jinfeng Rao, Huaixiu~Steven Zheng,
  Sanket~Vaibhav Mehta, Honglei Zhuang, Vinh~Q. Tran, Dara Bahri, Jianmo Ni,
  Jai~Prakash Gupta, Kai Hui, Sebastian Ruder, and Donald Metzler. 2021.
\newblock \href {http://arxiv.org/abs/2111.10952} {Ext5: Towards extreme
  multi-task scaling for transfer learning}.
\newblock \emph{CoRR}, abs/2111.10952.

\bibitem[{Artetxe et~al.(2019)Artetxe, Ruder, and Yogatama}]{Artetxe:etal:2019}
Mikel Artetxe, Sebastian Ruder, and Dani Yogatama. 2019.
\newblock \href {http://arxiv.org/abs/1910.11856} {On the cross-lingual
  transferability of monolingual representations}.
\newblock \emph{CoRR}, abs/1910.11856.

\bibitem[{Austin et~al.(2021)Austin, Odena, Nye, Bosma, Michalewski, Dohan,
  Jiang, Cai, Terry, Le et~al.}]{austin2021program}
Jacob Austin, Augustus Odena, Maxwell Nye, Maarten Bosma, Henryk Michalewski,
  David Dohan, Ellen Jiang, Carrie Cai, Michael Terry, Quoc Le, et~al. 2021.
\newblock Program synthesis with large language models.
\newblock \emph{arXiv preprint arXiv:2108.07732}.

\bibitem[{Bach et~al.(2022)Bach, Sanh, Yong, Webson, Raffel, Nayak, Sharma,
  Kim, Bari, Fevry, Alyafeai, Dey, Santilli, Sun, Ben-David, Xu, Chhablani,
  Wang, Fries, Al-shaibani, Sharma, Thakker, Almubarak, Tang, Tang, Jiang, and
  Rush}]{promptsource}
Stephen~H. Bach, Victor Sanh, Zheng-Xin Yong, Albert Webson, Colin Raffel,
  Nihal~V. Nayak, Abheesht Sharma, Taewoon Kim, M~Saiful Bari, Thibault Fevry,
  Zaid Alyafeai, Manan Dey, Andrea Santilli, Zhiqing Sun, Srulik Ben-David,
  Canwen Xu, Gunjan Chhablani, Han Wang, Jason~Alan Fries, Maged~S.
  Al-shaibani, Shanya Sharma, Urmish Thakker, Khalid Almubarak, Xiangru Tang,
  Xiangru Tang, Mike Tian-Jian Jiang, and Alexander~M. Rush. 2022.
\newblock \href {http://arxiv.org/abs/2202.01279} {Promptsource: An integrated
  development environment and repository for natural language prompts}.

\bibitem[{Biderman et~al.(2023)Biderman, Schoelkopf, Anthony, Bradley, O'Brien,
  Hallahan, Khan, Purohit, Prashanth, Raff et~al.}]{biderman2023pythia}
Stella Biderman, Hailey Schoelkopf, Quentin Anthony, Herbie Bradley, Kyle
  O'Brien, Eric Hallahan, Mohammad~Aflah Khan, Shivanshu Purohit, USVSN~Sai
  Prashanth, Edward Raff, et~al. 2023.
\newblock Pythia: A suite for analyzing large language models across training
  and scaling.
\newblock \emph{arXiv preprint arXiv:2304.01373}.

\bibitem[{Black et~al.(2022)Black, Biderman, Hallahan, Anthony, Gao, Golding,
  He, Leahy, McDonell, Phang et~al.}]{black2022gpt}
Sid Black, Stella Biderman, Eric Hallahan, Quentin Anthony, Leo Gao, Laurence
  Golding, Horace He, Connor Leahy, Kyle McDonell, Jason Phang, et~al. 2022.
\newblock Gpt-neox-20b: An open-source autoregressive language model.
\newblock \emph{arXiv preprint arXiv:2204.06745}.

\bibitem[{Black et~al.(2021)Black, Gao, Wang, Leahy, and
  Biderman}]{black2021gpt}
Sid Black, Leo Gao, Phil Wang, Connor Leahy, and Stella Biderman. 2021.
\newblock Gpt-neo: Large scale autoregressive language modeling with
  mesh-tensorflow.
\newblock \emph{If you use this software, please cite it using these metadata},
  58.

\bibitem[{Bommasani et~al.(2021)Bommasani, Hudson, Adeli, Altman, Arora, von
  Arx, Bernstein, Bohg, Bosselut, Brunskill
  et~al.}]{bommasani2021opportunities}
Rishi Bommasani, Drew~A Hudson, Ehsan Adeli, Russ Altman, Simran Arora, Sydney
  von Arx, Michael~S Bernstein, Jeannette Bohg, Antoine Bosselut, Emma
  Brunskill, et~al. 2021.
\newblock On the opportunities and risks of foundation models.
\newblock \emph{arXiv preprint arXiv:2108.07258}.

\bibitem[{Brown et~al.(2020)Brown, Mann, Ryder, Subbiah, Kaplan, Dhariwal,
  Neelakantan, Shyam, Sastry, Askell, Agarwal, Herbert-Voss, Krueger, Henighan,
  Child, Ramesh, Ziegler, Wu, Winter, Hesse, Chen, Sigler, Litwin, Gray, Chess,
  Clark, Berner, McCandlish, Radford, Sutskever, and Amodei}]{gpt3}
Tom Brown, Benjamin Mann, Nick Ryder, Melanie Subbiah, Jared~D Kaplan, Prafulla
  Dhariwal, Arvind Neelakantan, Pranav Shyam, Girish Sastry, Amanda Askell,
  Sandhini Agarwal, Ariel Herbert-Voss, Gretchen Krueger, Tom Henighan, Rewon
  Child, Aditya Ramesh, Daniel Ziegler, Jeffrey Wu, Clemens Winter, Chris
  Hesse, Mark Chen, Eric Sigler, Mateusz Litwin, Scott Gray, Benjamin Chess,
  Jack Clark, Christopher Berner, Sam McCandlish, Alec Radford, Ilya Sutskever,
  and Dario Amodei. 2020.
\newblock \href
  {https://proceedings.neurips.cc/paper/2020/file/1457c0d6bfcb4967418bfb8ac142f64a-Paper.pdf}
  {Language models are few-shot learners}.
\newblock In \emph{Advances in Neural Information Processing Systems},
  volume~33, pages 1877--1901. Curran Associates, Inc.

\bibitem[{Cahyawijaya et~al.(2023)Cahyawijaya, Lovenia, Yu, Chung, and
  Fung}]{cahyawijaya2023instruct}
Samuel Cahyawijaya, Holy Lovenia, Tiezheng Yu, Willy Chung, and Pascale Fung.
  2023.
\newblock Instruct-align: Teaching novel languages with to llms through
  alignment-based cross-lingual instruction.
\newblock \emph{arXiv preprint arXiv:2305.13627}.

\bibitem[{Chalkidis et~al.(2021)Chalkidis, Fergadiotis, and
  Androutsopoulos}]{chalkidis2021multieurlex}
Ilias Chalkidis, Manos Fergadiotis, and Ion Androutsopoulos. 2021.
\newblock Multieurlex--a multi-lingual and multi-label legal document
  classification dataset for zero-shot cross-lingual transfer.
\newblock \emph{arXiv preprint arXiv:2109.00904}.

\bibitem[{Chen et~al.(2021)Chen, Tworek, Jun, Yuan, Pinto, Kaplan, Edwards,
  Burda, Joseph, Brockman et~al.}]{chen2021evaluating}
Mark Chen, Jerry Tworek, Heewoo Jun, Qiming Yuan, Henrique Ponde de~Oliveira
  Pinto, Jared Kaplan, Harri Edwards, Yuri Burda, Nicholas Joseph, Greg
  Brockman, et~al. 2021.
\newblock Evaluating large language models trained on code.
\newblock \emph{arXiv preprint arXiv:2107.03374}.

\bibitem[{Chowdhery et~al.(2022)Chowdhery, Narang, Devlin, Bosma, Mishra,
  Roberts, Barham, Chung, Sutton, Gehrmann et~al.}]{chowdhery2022palm}
Aakanksha Chowdhery, Sharan Narang, Jacob Devlin, Maarten Bosma, Gaurav Mishra,
  Adam Roberts, Paul Barham, Hyung~Won Chung, Charles Sutton, Sebastian
  Gehrmann, et~al. 2022.
\newblock Palm: Scaling language modeling with pathways.
\newblock \emph{arXiv preprint arXiv:2204.02311}.

\bibitem[{Chung et~al.(2023)Chung, Constant, Garcia, Roberts, Tay, Narang, and
  Firat}]{chung2023unimax}
Hyung~Won Chung, Noah Constant, Xavier Garcia, Adam Roberts, Yi~Tay, Sharan
  Narang, and Orhan Firat. 2023.
\newblock Unimax: Fairer and more effective language sampling for large-scale
  multilingual pretraining.
\newblock \emph{arXiv preprint arXiv:2304.09151}.

\bibitem[{Chung et~al.(2022)Chung, Hou, Longpre, Zoph, Tay, Fedus, Li, Wang,
  Dehghani, Brahma, Webson, Gu, Dai, Suzgun, Chen, Chowdhery, Narang, Mishra,
  Yu, Zhao, Huang, Dai, Yu, Petrov, Chi, Dean, Devlin, Roberts, Zhou, Le, and
  Wei}]{chung2022scaling}
Hyung~Won Chung, Le~Hou, Shayne Longpre, Barret Zoph, Yi~Tay, William Fedus,
  Eric Li, Xuezhi Wang, Mostafa Dehghani, Siddhartha Brahma, Albert Webson,
  Shixiang~Shane Gu, Zhuyun Dai, Mirac Suzgun, Xinyun Chen, Aakanksha
  Chowdhery, Sharan Narang, Gaurav Mishra, Adams Yu, Vincent Zhao, Yanping
  Huang, Andrew Dai, Hongkun Yu, Slav Petrov, Ed~H. Chi, Jeff Dean, Jacob
  Devlin, Adam Roberts, Denny Zhou, Quoc~V. Le, and Jason Wei. 2022.
\newblock \href {https://arxiv.org/abs/2210.11416} {Scaling
  instruction-finetuned language models}.
\newblock \emph{arXiv preprint arXiv:2210.11416}.

\bibitem[{Conneau et~al.(2019)Conneau, Khandelwal, Goyal, Chaudhary, Wenzek,
  Guzm{\'a}n, Grave, Ott, Zettlemoyer, and Stoyanov}]{conneau2019unsupervised}
Alexis Conneau, Kartikay Khandelwal, Naman Goyal, Vishrav Chaudhary, Guillaume
  Wenzek, Francisco Guzm{\'a}n, Edouard Grave, Myle Ott, Luke Zettlemoyer, and
  Veselin Stoyanov. 2019.
\newblock Unsupervised cross-lingual representation learning at scale.
\newblock \emph{arXiv preprint arXiv:1911.02116}.

\bibitem[{Conneau et~al.(2018)Conneau, Rinott, Lample, Williams, Bowman,
  Schwenk, and Stoyanov}]{conneau2018xnli}
Alexis Conneau, Ruty Rinott, Guillaume Lample, Adina Williams, Samuel~R.
  Bowman, Holger Schwenk, and Veselin Stoyanov. 2018.
\newblock Xnli: Evaluating cross-lingual sentence representations.
\newblock In \emph{Proceedings of the 2018 Conference on Empirical Methods in
  Natural Language Processing}. Association for Computational Linguistics.

\bibitem[{Cui et~al.(2018)Cui, Liu, Xiao, Chen, Ma, Che, Wang, and
  Hu}]{cmrc2018-dataset}
Yiming Cui, Ting Liu, Li~Xiao, Zhipeng Chen, Wentao Ma, Wanxiang Che, Shijin
  Wang, and Guoping Hu. 2018.
\newblock A span-extraction dataset for chinese machine reading comprehension.
\newblock \emph{arXiv preprint arXiv:1810.07366}.

\bibitem[{Dettmers et~al.(2023)Dettmers, Pagnoni, Holtzman, and
  Zettlemoyer}]{dettmers2023qlora}
Tim Dettmers, Artidoro Pagnoni, Ari Holtzman, and Luke Zettlemoyer. 2023.
\newblock Qlora: Efficient finetuning of quantized llms.
\newblock \emph{arXiv preprint arXiv:2305.14314}.

\bibitem[{Devlin et~al.(2018)Devlin, Chang, Lee, and
  Toutanova}]{devlin2018bert}
Jacob Devlin, Ming-Wei Chang, Kenton Lee, and Kristina Toutanova. 2018.
\newblock Bert: Pre-training of deep bidirectional transformers for language
  understanding.
\newblock \emph{arXiv preprint arXiv:1810.04805}.

\bibitem[{Dhole et~al.(2021)Dhole, Gangal, Gehrmann, Gupta, Li, Mahamood,
  Mahendiran, Mille, Srivastava, Tan et~al.}]{dhole2021nl}
Kaustubh~D Dhole, Varun Gangal, Sebastian Gehrmann, Aadesh Gupta, Zhenhao Li,
  Saad Mahamood, Abinaya Mahendiran, Simon Mille, Ashish Srivastava, Samson
  Tan, et~al. 2021.
\newblock Nl-augmenter: A framework for task-sensitive natural language
  augmentation.
\newblock \emph{arXiv preprint arXiv:2112.02721}.

\bibitem[{Ding et~al.(2022)Ding, Qin, Yang, Wei, Yang, Su, Hu, Chen, Chan, Chen
  et~al.}]{ding2022delta}
Ning Ding, Yujia Qin, Guang Yang, Fuchao Wei, Zonghan Yang, Yusheng Su,
  Shengding Hu, Yulin Chen, Chi-Min Chan, Weize Chen, et~al. 2022.
\newblock Delta tuning: A comprehensive study of parameter efficient methods
  for pre-trained language models.
\newblock \emph{arXiv preprint arXiv:2203.06904}.

\bibitem[{Ennen et~al.(2023)Ennen, Hsu, Hsu, Liu, Wu, Liao, Lin, Shiu, and
  Ma}]{ennen2023extending}
Philipp Ennen, Po-Chun Hsu, Chan-Jan Hsu, Chang-Le Liu, Yen-Chen Wu, Yin-Hsiang
  Liao, Chin-Tung Lin, Da-Shan Shiu, and Wei-Yun Ma. 2023.
\newblock Extending the pre-training of bloom for improved support of
  traditional chinese: Models, methods and results.
\newblock \emph{arXiv preprint arXiv:2303.04715}.

\bibitem[{Fan et~al.(2021)Fan, Bhosale, Schwenk, Ma, El-Kishky, Goyal, Baines,
  Celebi, Wenzek, Chaudhary et~al.}]{fan2021beyond}
Angela Fan, Shruti Bhosale, Holger Schwenk, Zhiyi Ma, Ahmed El-Kishky,
  Siddharth Goyal, Mandeep Baines, Onur Celebi, Guillaume Wenzek, Vishrav
  Chaudhary, et~al. 2021.
\newblock Beyond english-centric multilingual machine translation.
\newblock \emph{J. Mach. Learn. Res.}, 22(107):1--48.

\bibitem[{Fried et~al.(2022)Fried, Aghajanyan, Lin, Wang, Wallace, Shi, Zhong,
  Yih, Zettlemoyer, and Lewis}]{fried2022incoder}
Daniel Fried, Armen Aghajanyan, Jessy Lin, Sida Wang, Eric Wallace, Freda Shi,
  Ruiqi Zhong, Wen-tau Yih, Luke Zettlemoyer, and Mike Lewis. 2022.
\newblock Incoder: A generative model for code infilling and synthesis.
\newblock \emph{arXiv preprint arXiv:2204.05999}.

\bibitem[{Fries et~al.(2022)Fries, Weber, Seelam, Altay, Datta, Garda, Kang,
  Su, Kusa, Cahyawijaya et~al.}]{fries2022bigbio}
Jason~Alan Fries, Leon Weber, Natasha Seelam, Gabriel Altay, Debajyoti Datta,
  Samuele Garda, Myungsun Kang, Ruisi Su, Wojciech Kusa, Samuel Cahyawijaya,
  et~al. 2022.
\newblock Bigbio: A framework for data-centric biomedical natural language
  processing.
\newblock \emph{arXiv preprint arXiv:2206.15076}.

\bibitem[{Gao et~al.(2021)Gao, Tow, Biderman, Black, DiPofi, Foster, Golding,
  Hsu, McDonell, Muennighoff, Phang, Reynolds, Tang, Thite, Wang, Wang, and
  Zou}]{eval-harness}
Leo Gao, Jonathan Tow, Stella Biderman, Sid Black, Anthony DiPofi, Charles
  Foster, Laurence Golding, Jeffrey Hsu, Kyle McDonell, Niklas Muennighoff,
  Jason Phang, Laria Reynolds, Eric Tang, Anish Thite, Ben Wang, Kevin Wang,
  and Andy Zou. 2021.
\newblock \href {https://doi.org/10.5281/zenodo.5371628} {A framework for
  few-shot language model evaluation}.

\bibitem[{Goyal et~al.(2021{\natexlab{a}})Goyal, Du, Ott, Anantharaman, and
  Conneau}]{goyal2021larger}
Naman Goyal, Jingfei Du, Myle Ott, Giri Anantharaman, and Alexis Conneau.
  2021{\natexlab{a}}.
\newblock Larger-scale transformers for multilingual masked language modeling.
\newblock \emph{arXiv preprint arXiv:2105.00572}.

\bibitem[{Goyal et~al.(2021{\natexlab{b}})Goyal, Gao, Chaudhary, Chen, Wenzek,
  Ju, Krishnan, Ranzato, Guzm'{a}n, and Fan}]{goyal2021flores}
Naman Goyal, Cynthia Gao, Vishrav Chaudhary, Peng-Jen Chen, Guillaume Wenzek,
  Da~Ju, Sanjana Krishnan, Marc'Aurelio Ranzato, Francisco Guzm'{a}n, and
  Angela Fan. 2021{\natexlab{b}}.
\newblock The flores-101 evaluation benchmark for low-resource and multilingual
  machine translation.

\bibitem[{Guzm'{a}n et~al.(2019)Guzm'{a}n, Chen, Ott, Pino, Lample, Koehn,
  Chaudhary, and Ranzato}]{guzman2019two}
Francisco Guzm'{a}n, Peng-Jen Chen, Myle Ott, Juan Pino, Guillaume Lample,
  Philipp Koehn, Vishrav Chaudhary, and Marc'Aurelio Ranzato. 2019.
\newblock Two new evaluation datasets for low-resource machine translation:
  Nepali-english and sinhala-english.

\bibitem[{Hasan et~al.(2021)Hasan, Bhattacharjee, Islam, Mubasshir, Li, Kang,
  Rahman, and Shahriyar}]{hasan-etal-2021-xl}
Tahmid Hasan, Abhik Bhattacharjee, Md.~Saiful Islam, Kazi Mubasshir, Yuan-Fang
  Li, Yong-Bin Kang, M.~Sohel Rahman, and Rifat Shahriyar. 2021.
\newblock \href {https://aclanthology.org/2021.findings-acl.413} {{XL}-sum:
  Large-scale multilingual abstractive summarization for 44 languages}.
\newblock In \emph{Findings of the Association for Computational Linguistics:
  ACL-IJCNLP 2021}, pages 4693--4703, Online. Association for Computational
  Linguistics.

\bibitem[{Hellendoorn et~al.(2020)Hellendoorn, Sutton, Singh, Maniatis, and
  Bieber}]{DBLP:conf/iclr/HellendoornSSMB20}
Vincent~J. Hellendoorn, Charles Sutton, Rishabh Singh, Petros Maniatis, and
  David Bieber. 2020.
\newblock \href {https://openreview.net/forum?id=B1lnbRNtwr} {Global relational
  models of source code}.
\newblock In \emph{8th International Conference on Learning Representations,
  {ICLR} 2020, Addis Ababa, Ethiopia, April 26-30, 2020}. OpenReview.net.

\bibitem[{Hendrycks et~al.(2021)Hendrycks, Basart, Kadavath, Mazeika, Arora,
  Guo, Burns, Puranik, He, Song, and Steinhardt}]{hendrycksapps2021}
Dan Hendrycks, Steven Basart, Saurav Kadavath, Mantas Mazeika, Akul Arora,
  Ethan Guo, Collin Burns, Samir Puranik, Horace He, Dawn Song, and Jacob
  Steinhardt. 2021.
\newblock Measuring coding challenge competence with apps.
\newblock \emph{NeurIPS}.

\bibitem[{Hoffmann et~al.(2022)Hoffmann, Borgeaud, Mensch, Buchatskaya, Cai,
  Rutherford, Casas, Hendricks, Welbl, Clark et~al.}]{hoffmann2022training}
Jordan Hoffmann, Sebastian Borgeaud, Arthur Mensch, Elena Buchatskaya, Trevor
  Cai, Eliza Rutherford, Diego de~Las Casas, Lisa~Anne Hendricks, Johannes
  Welbl, Aidan Clark, et~al. 2022.
\newblock Training compute-optimal large language models.
\newblock \emph{arXiv preprint arXiv:2203.15556}.

\bibitem[{Honovich et~al.(2022)Honovich, Scialom, Levy, and
  Schick}]{honovich2022unnatural}
Or~Honovich, Thomas Scialom, Omer Levy, and Timo Schick. 2022.
\newblock Unnatural instructions: Tuning language models with (almost) no human
  labor.
\newblock \emph{arXiv preprint arXiv:2212.09689}.

\bibitem[{Hu et~al.(2021)Hu, Shen, Wallis, Allen-Zhu, Li, Wang, Wang, and
  Chen}]{hu2021lora}
Edward~J Hu, Yelong Shen, Phillip Wallis, Zeyuan Allen-Zhu, Yuanzhi Li, Shean
  Wang, Lu~Wang, and Weizhu Chen. 2021.
\newblock Lora: Low-rank adaptation of large language models.
\newblock \emph{arXiv preprint arXiv:2106.09685}.

\bibitem[{Iyer et~al.(2022)Iyer, Lin, Pasunuru, Mihaylov, Simig, Yu, Shuster,
  Wang, Liu, Koura et~al.}]{iyer2022opt}
Srinivasan Iyer, Xi~Victoria Lin, Ramakanth Pasunuru, Todor Mihaylov,
  D{\'a}niel Simig, Ping Yu, Kurt Shuster, Tianlu Wang, Qing Liu, Punit~Singh
  Koura, et~al. 2022.
\newblock Opt-iml: Scaling language model instruction meta learning through the
  lens of generalization.
\newblock \emph{arXiv preprint arXiv:2212.12017}.

\bibitem[{Kim et~al.(2021)Kim, Maddela, Kriz, Xu, and
  Callison-Burch}]{kim-etal-2021-bisect}
Joongwon Kim, Mounica Maddela, Reno Kriz, Wei Xu, and Chris Callison-Burch.
  2021.
\newblock \href {https://aclanthology.org/2021.emnlp-main.500} {{B}i{SECT}:
  Learning to split and rephrase sentences with bitexts}.
\newblock In \emph{Proceedings of the 2021 Conference on Empirical Methods in
  Natural Language Processing}, pages 6193--6209, Online and Punta Cana,
  Dominican Republic. Association for Computational Linguistics.

\bibitem[{Komatsuzaki et~al.(2022)Komatsuzaki, Puigcerver, Lee-Thorp, Ruiz,
  Mustafa, Ainslie, Tay, Dehghani, and Houlsby}]{komatsuzaki2022sparse}
Aran Komatsuzaki, Joan Puigcerver, James Lee-Thorp, Carlos~Riquelme Ruiz, Basil
  Mustafa, Joshua Ainslie, Yi~Tay, Mostafa Dehghani, and Neil Houlsby. 2022.
\newblock Sparse upcycling: Training mixture-of-experts from dense checkpoints.
\newblock \emph{arXiv preprint arXiv:2212.05055}.

\bibitem[{Kosec et~al.(2021)Kosec, Fu, and Krell}]{kosec2021packing}
Matej Kosec, Sheng Fu, and Mario~Michael Krell. 2021.
\newblock Packing: Towards 2x nlp bert acceleration.
\newblock \emph{arXiv preprint arXiv:2107.02027}.

\bibitem[{Lample and Conneau(2019)}]{lample2019}
Guillaume Lample and Alexis Conneau. 2019.
\newblock \href {https://doi.org/10.48550/ARXIV.1901.07291} {Cross-lingual
  language model pretraining}.

\bibitem[{Lauren{\c{c}}on et~al.(2022)Lauren{\c{c}}on, Saulnier, Wang, Akiki,
  del Moral, Le~Scao, Von~Werra, Mou, Ponferrada, Nguyen
  et~al.}]{laurencconbigscience}
Hugo Lauren{\c{c}}on, Lucile Saulnier, Thomas Wang, Christopher Akiki,
  Albert~Villanova del Moral, Teven Le~Scao, Leandro Von~Werra, Chenghao Mou,
  Eduardo~Gonz{\'a}lez Ponferrada, Huu Nguyen, et~al. 2022.
\newblock The bigscience roots corpus: A 1.6 tb composite multilingual dataset.
\newblock In \emph{Thirty-sixth Conference on Neural Information Processing
  Systems Datasets and Benchmarks Track}.

\bibitem[{Levenshtein et~al.(1966)}]{levenshtein1966binary}
Vladimir~I Levenshtein et~al. 1966.
\newblock Binary codes capable of correcting deletions, insertions, and
  reversals.
\newblock In \emph{Soviet physics doklady}, volume~10, pages 707--710. Soviet
  Union.

\bibitem[{Lewis et~al.(2019)Lewis, Oguz, Rinott, Riedel, and
  Schwenk}]{lewis2019mlqa}
Patrick Lewis, Barlas Oguz, Ruty Rinott, Sebastian Riedel, and Holger Schwenk.
  2019.
\newblock Mlqa: Evaluating cross-lingual extractive question answering.
\newblock \emph{arXiv preprint arXiv:1910.07475}.

\bibitem[{Li et~al.(2023)Li, Allal, Zi, Muennighoff, Kocetkov, Mou, Marone,
  Akiki, Li, Chim et~al.}]{li2023starcoder}
Raymond Li, Loubna~Ben Allal, Yangtian Zi, Niklas Muennighoff, Denis Kocetkov,
  Chenghao Mou, Marc Marone, Christopher Akiki, Jia Li, Jenny Chim, et~al.
  2023.
\newblock Starcoder: may the source be with you!
\newblock \emph{arXiv preprint arXiv:2305.06161}.

\bibitem[{Lin et~al.(2021)Lin, Mihaylov, Artetxe, Wang, Chen, Simig, Ott,
  Goyal, Bhosale, Du et~al.}]{lin2021few}
Xi~Victoria Lin, Todor Mihaylov, Mikel Artetxe, Tianlu Wang, Shuohui Chen,
  Daniel Simig, Myle Ott, Naman Goyal, Shruti Bhosale, Jingfei Du, et~al. 2021.
\newblock Few-shot learning with multilingual language models.
\newblock \emph{arXiv preprint arXiv:2112.10668}.

\bibitem[{Liu et~al.(2022)Liu, Tam, Muqeeth, Mohta, Huang, Bansal, and
  Raffel}]{liu2022few}
Haokun Liu, Derek Tam, Mohammed Muqeeth, Jay Mohta, Tenghao Huang, Mohit
  Bansal, and Colin Raffel. 2022.
\newblock Few-shot parameter-efficient fine-tuning is better and cheaper than
  in-context learning.
\newblock \emph{arXiv preprint arXiv:2205.05638}.

\bibitem[{Liu et~al.(2023)Liu, Zhou, Jiang, Dou, and Lin}]{liu2023zero}
Qian Liu, Fan Zhou, Zhengbao Jiang, Longxu Dou, and Min Lin. 2023.
\newblock From zero to hero: Examining the power of symbolic tasks in
  instruction tuning.
\newblock \emph{arXiv preprint arXiv:2304.07995}.

\bibitem[{Liu et~al.(2020)Liu, Gu, Goyal, Li, Edunov, Ghazvininejad, Lewis, and
  Zettlemoyer}]{liu2020multilingual}
Yinhan Liu, Jiatao Gu, Naman Goyal, Xian Li, Sergey Edunov, Marjan
  Ghazvininejad, Mike Lewis, and Luke Zettlemoyer. 2020.
\newblock Multilingual denoising pre-training for neural machine translation.
\newblock \emph{Transactions of the Association for Computational Linguistics},
  8:726--742.

\bibitem[{Logan et~al.(2021)Logan, Bala{\v{z}}evi{\'c}, Wallace, Petroni,
  Singh, and Riedel}]{logan2021cutting}
Robert~L Logan, Ivana Bala{\v{z}}evi{\'c}, Eric Wallace, Fabio Petroni, Sameer
  Singh, and Sebastian Riedel. 2021.
\newblock Cutting down on prompts and parameters: Simple few-shot learning with
  language models.
\newblock \emph{arXiv preprint arXiv:2106.13353}.

\bibitem[{Longpre et~al.(2023{\natexlab{a}})Longpre, Hou, Vu, Webson, Chung,
  Tay, Zhou, Le, Zoph, Wei et~al.}]{longpre2023flan}
Shayne Longpre, Le~Hou, Tu~Vu, Albert Webson, Hyung~Won Chung, Yi~Tay, Denny
  Zhou, Quoc~V Le, Barret Zoph, Jason Wei, et~al. 2023{\natexlab{a}}.
\newblock The flan collection: Designing data and methods for effective
  instruction tuning.
\newblock \emph{arXiv preprint arXiv:2301.13688}.

\bibitem[{Longpre et~al.(2023{\natexlab{b}})Longpre, Yauney, Reif, Lee,
  Roberts, Zoph, Zhou, Wei, Robinson, Mimno et~al.}]{longpre2023pretrainer}
Shayne Longpre, Gregory Yauney, Emily Reif, Katherine Lee, Adam Roberts, Barret
  Zoph, Denny Zhou, Jason Wei, Kevin Robinson, David Mimno, et~al.
  2023{\natexlab{b}}.
\newblock A pretrainer's guide to training data: Measuring the effects of data
  age, domain coverage, quality, \& toxicity.
\newblock \emph{arXiv preprint arXiv:2305.13169}.

\bibitem[{Min et~al.(2021)Min, Lewis, Zettlemoyer, and
  Hajishirzi}]{min2021metaicl}
Sewon Min, Mike Lewis, Luke Zettlemoyer, and Hannaneh Hajishirzi. 2021.
\newblock Metaicl: Learning to learn in context.
\newblock \emph{arXiv preprint arXiv:2110.15943}.

\bibitem[{Mishra et~al.(2021{\natexlab{a}})Mishra, Khashabi, Baral, and
  Hajishirzi}]{mishra2021cross}
Swaroop Mishra, Daniel Khashabi, Chitta Baral, and Hannaneh Hajishirzi.
  2021{\natexlab{a}}.
\newblock Cross-task generalization via natural language crowdsourcing
  instructions.
\newblock \emph{arXiv preprint arXiv:2104.08773}.

\bibitem[{Mishra et~al.(2021{\natexlab{b}})Mishra, Khashabi, Baral, and
  Hajishirzi}]{mishra2021}
Swaroop Mishra, Daniel Khashabi, Chitta Baral, and Hannaneh Hajishirzi.
  2021{\natexlab{b}}.
\newblock \href {http://arxiv.org/abs/2104.08773} {Natural instructions:
  Benchmarking generalization to new tasks from natural language instructions}.
\newblock \emph{CoRR}, abs/2104.08773.

\bibitem[{Muennighoff(2022)}]{muennighoff2022sgpt}
Niklas Muennighoff. 2022.
\newblock Sgpt: Gpt sentence embeddings for semantic search.
\newblock \emph{arXiv preprint arXiv:2202.08904}.

\bibitem[{Muennighoff et~al.(2023)Muennighoff, Rush, Barak, Scao, Piktus, Tazi,
  Pyysalo, Wolf, and Raffel}]{muennighoff2023scaling}
Niklas Muennighoff, Alexander~M. Rush, Boaz Barak, Teven~Le Scao, Aleksandra
  Piktus, Nouamane Tazi, Sampo Pyysalo, Thomas Wolf, and Colin Raffel. 2023.
\newblock \href {http://arxiv.org/abs/2305.16264} {Scaling data-constrained
  language models}.

\bibitem[{Muennighoff et~al.(2022)Muennighoff, Tazi, Magne, and
  Reimers}]{muennighoff2022mteb}
Niklas Muennighoff, Nouamane Tazi, Lo{\"\i}c Magne, and Nils Reimers. 2022.
\newblock \href {https://doi.org/10.48550/ARXIV.2210.07316} {Mteb: Massive text
  embedding benchmark}.
\newblock \emph{arXiv preprint arXiv:2210.07316}.

\bibitem[{{NLLB Team} et~al.(2022){NLLB Team}, Costa-jussà, Cross, Çelebi,
  Elbayad, Heafield, Heffernan, Kalbassi, Lam, Licht, Maillard, Sun, Wang,
  Wenzek, Youngblood, Akula, Barrault, Gonzalez, Hansanti, Hoffman, Jarrett,
  Sadagopan, Rowe, Spruit, Tran, Andrews, Ayan, Bhosale, Edunov, Fan, Gao,
  Goswami, Guzmán, Koehn, Mourachko, Ropers, Saleem, Schwenk, and
  Wang}]{nllb2022}
{NLLB Team}, Marta~R. Costa-jussà, James Cross, Onur Çelebi, Maha Elbayad,
  Kenneth Heafield, Kevin Heffernan, Elahe Kalbassi, Janice Lam, Daniel Licht,
  Jean Maillard, Anna Sun, Skyler Wang, Guillaume Wenzek, Al~Youngblood, Bapi
  Akula, Loic Barrault, Gabriel~Mejia Gonzalez, Prangthip Hansanti, John
  Hoffman, Semarley Jarrett, Kaushik~Ram Sadagopan, Dirk Rowe, Shannon Spruit,
  Chau Tran, Pierre Andrews, Necip~Fazil Ayan, Shruti Bhosale, Sergey Edunov,
  Angela Fan, Cynthia Gao, Vedanuj Goswami, Francisco Guzmán, Philipp Koehn,
  Alexandre Mourachko, Christophe Ropers, Safiyyah Saleem, Holger Schwenk, and
  Jeff Wang. 2022.
\newblock \href {https://doi.org/10.48550/ARXIV.2207.04672} {No language left
  behind: Scaling human-centered machine translation}.
\newblock \emph{arXiv preprint 2207.04672}.

\bibitem[{Ouyang et~al.(2022)Ouyang, Wu, Jiang, Almeida, Wainwright, Mishkin,
  Zhang, Agarwal, Slama, Ray et~al.}]{ouyang2022training}
Long Ouyang, Jeff Wu, Xu~Jiang, Diogo Almeida, Carroll~L Wainwright, Pamela
  Mishkin, Chong Zhang, Sandhini Agarwal, Katarina Slama, Alex Ray, et~al.
  2022.
\newblock Training language models to follow instructions with human feedback.
\newblock \emph{arXiv preprint arXiv:2203.02155}.

\bibitem[{Papineni et~al.(2002)Papineni, Roukos, Ward, and
  Zhu}]{papineni2002bleu}
Kishore Papineni, Salim Roukos, Todd Ward, and Wei-Jing Zhu. 2002.
\newblock Bleu: a method for automatic evaluation of machine translation.
\newblock In \emph{Proceedings of the 40th annual meeting of the Association
  for Computational Linguistics}, pages 311--318.

\bibitem[{Patel et~al.(2022)Patel, Li, Rasooli, Constant, Raffel, and
  Callison-Burch}]{patel2022bidirectional}
Ajay Patel, Bryan Li, Mohammad~Sadegh Rasooli, Noah Constant, Colin Raffel, and
  Chris Callison-Burch. 2022.
\newblock Bidirectional language models are also few-shot learners.
\newblock \emph{arXiv preprint arXiv:2209.14500}.

\bibitem[{Perez et~al.(2021{\natexlab{a}})Perez, Kiela, and
  Cho}]{perez2021true}
Ethan Perez, Douwe Kiela, and Kyunghyun Cho. 2021{\natexlab{a}}.
\newblock True few-shot learning with language models.
\newblock \emph{Advances in Neural Information Processing Systems},
  34:11054--11070.

\bibitem[{Perez et~al.(2021{\natexlab{b}})Perez, Kiela, and
  Cho}]{true-zero-shot}
Ethan Perez, Douwe Kiela, and Kyunghyun Cho. 2021{\natexlab{b}}.
\newblock \href {http://arxiv.org/abs/2105.11447} {True few-shot learning with
  language models}.
\newblock \emph{CoRR}, abs/2105.11447.

\bibitem[{Phang et~al.(2020)Phang, Calixto, Htut, Pruksachatkun, Liu, Vania,
  Kann, and Bowman}]{phang2020english}
Jason Phang, Iacer Calixto, Phu~Mon Htut, Yada Pruksachatkun, Haokun Liu, Clara
  Vania, Katharina Kann, and Samuel~R Bowman. 2020.
\newblock English intermediate-task training improves zero-shot cross-lingual
  transfer too.
\newblock \emph{arXiv preprint arXiv:2005.13013}.

\bibitem[{Ponti et~al.(2020)Ponti, Glavas, Majewska, Liu, Vuli'{c}, and
  Korhonen}]{ponti2020xcopa}
Edoardo~M. Ponti, Goran Glavas, Olga Majewska, Qianchu Liu, Ivan Vuli'{c}, and
  Anna Korhonen. 2020.
\newblock \href {https://ducdauge.github.io/files/xcopa.pdf} {{XCOPA: A}
  multilingual dataset for causal commonsense reasoning}.
\newblock \emph{arXiv preprint}.

\bibitem[{Radford et~al.(2019)Radford, Wu, Child, Luan, Amodei, Sutskever
  et~al.}]{radford2019language}
Alec Radford, Jeffrey Wu, Rewon Child, David Luan, Dario Amodei, Ilya
  Sutskever, et~al. 2019.
\newblock Language models are unsupervised multitask learners.
\newblock \emph{OpenAI blog}, 1(8):9.

\bibitem[{Rae et~al.(2021)Rae, Borgeaud, Cai, Millican, Hoffmann, Song,
  Aslanides, Henderson, Ring, Young et~al.}]{rae2021scaling}
Jack~W Rae, Sebastian Borgeaud, Trevor Cai, Katie Millican, Jordan Hoffmann,
  Francis Song, John Aslanides, Sarah Henderson, Roman Ring, Susannah Young,
  et~al. 2021.
\newblock Scaling language models: Methods, analysis \& insights from training
  gopher.
\newblock \emph{arXiv preprint arXiv:2112.11446}.

\bibitem[{Raffel et~al.(2020)Raffel, Shazeer, Roberts, Lee, Narang, Matena,
  Zhou, Li, Liu et~al.}]{raffel2020exploring}
Colin Raffel, Noam Shazeer, Adam Roberts, Katherine Lee, Sharan Narang, Michael
  Matena, Yanqi Zhou, Wei Li, Peter~J Liu, et~al. 2020.
\newblock Exploring the limits of transfer learning with a unified text-to-text
  transformer.
\newblock \emph{J. Mach. Learn. Res.}, 21(140):1--67.

\bibitem[{Raganato et~al.(2020)Raganato, Pasini, Camacho-Collados, and
  Pilehvar}]{raganato-etal-2020-xl-wic}
Alessandro Raganato, Tommaso Pasini, Jose Camacho-Collados, and Mohammad~Taher
  Pilehvar. 2020.
\newblock Xl-wic: A multilingual benchmark for evaluating semantic
  contextualization.
\newblock In \emph{Proceedings of the 2020 Conference on Empirical Methods in
  Natural Language Processing (EMNLP)}, pages 7193--7206.

\bibitem[{Roberts et~al.(2022)Roberts, Chung, Levskaya, Mishra, Bradbury,
  Andor, Narang, Lester, Gaffney, Mohiuddin, Hawthorne, Lewkowycz, Salcianu,
  van Zee, Austin, Goodman, Soares, Hu, Tsvyashchenko, Chowdhery, Bastings,
  Bulian, Garcia, Ni, Chen, Kenealy, Clark, Lee, Garrette, Lee-Thorp, Raffel,
  Shazeer, Ritter, Bosma, Passos, Maitin-Shepard, Fiedel, Omernick, Saeta,
  Sepassi, Spiridonov, Newlan, and Gesmundo}]{roberts2022t5x}
Adam Roberts, Hyung~Won Chung, Anselm Levskaya, Gaurav Mishra, James Bradbury,
  Daniel Andor, Sharan Narang, Brian Lester, Colin Gaffney, Afroz Mohiuddin,
  Curtis Hawthorne, Aitor Lewkowycz, Alex Salcianu, Marc van Zee, Jacob Austin,
  Sebastian Goodman, Livio~Baldini Soares, Haitang Hu, Sasha Tsvyashchenko,
  Aakanksha Chowdhery, Jasmijn Bastings, Jannis Bulian, Xavier Garcia, Jianmo
  Ni, Andrew Chen, Kathleen Kenealy, Jonathan~H. Clark, Stephan Lee, Dan
  Garrette, James Lee-Thorp, Colin Raffel, Noam Shazeer, Marvin Ritter, Maarten
  Bosma, Alexandre Passos, Jeremy Maitin-Shepard, Noah Fiedel, Mark Omernick,
  Brennan Saeta, Ryan Sepassi, Alexander Spiridonov, Joshua Newlan, and Andrea
  Gesmundo. 2022.
\newblock \href {https://arxiv.org/abs/2203.17189} {Scaling up models and data
  with $\texttt{t5x}$ and $\texttt{seqio}$}.
\newblock \emph{arXiv preprint arXiv:2203.17189}.

\bibitem[{Roemmele et~al.(2011)Roemmele, Bejan, and
  Gordon}]{roemmele2011choice}
Melissa Roemmele, Cosmin~Adrian Bejan, and Andrew~S Gordon. 2011.
\newblock \href
  {https://people.ict.usc.edu/~gordon/publications/AAAI-SPRING11A.PDF} {Choice
  of plausible alternatives: An evaluation of commonsense causal reasoning}.
\newblock In \emph{2011 AAAI Spring Symposium Series}.

\bibitem[{Sanh et~al.(2022)Sanh, Webson, Raffel, Bach, Sutawika, Alyafeai,
  Chaffin, Stiegler, Le~Scao, Raja et~al.}]{sanh2022multitask}
Victor Sanh, Albert Webson, Colin Raffel, Stephen Bach, Lintang Sutawika, Zaid
  Alyafeai, Antoine Chaffin, Arnaud Stiegler, Teven Le~Scao, Arun Raja, et~al.
  2022.
\newblock Multitask prompted training enables zero-shot task generalization.
\newblock In \emph{The Tenth International Conference on Learning
  Representations}.

\bibitem[{Scao et~al.(2022{\natexlab{a}})Scao, Fan, Akiki, Pavlick, Ili{\'c},
  Hesslow, Castagn{\'e}, Luccioni, Yvon, Gall{\'e} et~al.}]{scao2022bloom}
Teven~Le Scao, Angela Fan, Christopher Akiki, Ellie Pavlick, Suzana Ili{\'c},
  Daniel Hesslow, Roman Castagn{\'e}, Alexandra~Sasha Luccioni, Fran{\c{c}}ois
  Yvon, Matthias Gall{\'e}, et~al. 2022{\natexlab{a}}.
\newblock Bloom: A 176b-parameter open-access multilingual language model.
\newblock \emph{arXiv preprint arXiv:2211.05100}.

\bibitem[{Scao and Rush(2021)}]{scao2021many}
Teven~Le Scao and Alexander~M Rush. 2021.
\newblock How many data points is a prompt worth?
\newblock \emph{arXiv preprint arXiv:2103.08493}.

\bibitem[{Scao et~al.(2022{\natexlab{b}})Scao, Wang, Hesslow, Saulnier, Bekman,
  Bari, Bideman, Elsahar, Muennighoff, Phang et~al.}]{scao2022language}
Teven~Le Scao, Thomas Wang, Daniel Hesslow, Lucile Saulnier, Stas Bekman,
  M~Saiful Bari, Stella Bideman, Hady Elsahar, Niklas Muennighoff, Jason Phang,
  et~al. 2022{\natexlab{b}}.
\newblock What language model to train if you have one million gpu hours?
\newblock \emph{arXiv preprint arXiv:2210.15424}.

\bibitem[{Schick and Sch{\"u}tze(2020)}]{schick2020exploiting}
Timo Schick and Hinrich Sch{\"u}tze. 2020.
\newblock Exploiting cloze questions for few shot text classification and
  natural language inference.
\newblock \emph{arXiv preprint arXiv:2001.07676}.

\bibitem[{Schick and Sch{\"{u}}tze(2020)}]{DBLP:journals/corr/abs-2001-07676}
Timo Schick and Hinrich Sch{\"{u}}tze. 2020.
\newblock \href {http://arxiv.org/abs/2001.07676} {Exploiting cloze questions
  for few-shot text classification and natural language inference}.
\newblock \emph{CoRR}, abs/2001.07676.

\bibitem[{Scialom et~al.(2022)Scialom, Chakrabarty, and
  Muresan}]{scialom2022continual}
Thomas Scialom, Tuhin Chakrabarty, and Smaranda Muresan. 2022.
\newblock Continual-t0: Progressively instructing 50+ tasks to language models
  without forgetting.
\newblock \emph{arXiv preprint arXiv:2205.12393}.

\bibitem[{Shen et~al.(2023)Shen, Hou, Zhou, Du, Longpre, Wei, Chung, Zoph,
  Fedus, Chen et~al.}]{shen2023flan}
Sheng Shen, Le~Hou, Yanqi Zhou, Nan Du, Shayne Longpre, Jason Wei, Hyung~Won
  Chung, Barret Zoph, William Fedus, Xinyun Chen, et~al. 2023.
\newblock Flan-moe: Scaling instruction-finetuned language models with sparse
  mixture of experts.
\newblock \emph{arXiv preprint arXiv:2305.14705}.

\bibitem[{Shliazhko et~al.(2022)Shliazhko, Fenogenova, Tikhonova, Mikhailov,
  Kozlova, and Shavrina}]{shliazhko2022mgpt}
Oleh Shliazhko, Alena Fenogenova, Maria Tikhonova, Vladislav Mikhailov,
  Anastasia Kozlova, and Tatiana Shavrina. 2022.
\newblock mgpt: Few-shot learners go multilingual.
\newblock \emph{arXiv preprint arXiv:2204.07580}.

\bibitem[{Smith et~al.(2022)Smith, Patwary, Norick, LeGresley, Rajbhandari,
  Casper, Liu, Prabhumoye, Zerveas, Korthikanti et~al.}]{smith2022using}
Shaden Smith, Mostofa Patwary, Brandon Norick, Patrick LeGresley, Samyam
  Rajbhandari, Jared Casper, Zhun Liu, Shrimai Prabhumoye, George Zerveas,
  Vijay Korthikanti, et~al. 2022.
\newblock Using deepspeed and megatron to train megatron-turing nlg 530b, a
  large-scale generative language model.
\newblock \emph{arXiv preprint arXiv:2201.11990}.

\bibitem[{Soltan et~al.(2022)Soltan, Ananthakrishnan, FitzGerald, Gupta, Hamza,
  Khan, Peris, Rawls, Rosenbaum, Rumshisky, Prakash, Sridhar, Triefenbach,
  Verma, Tur, and Natarajan}]{saleh2022}
Saleh Soltan, Shankar Ananthakrishnan, Jack FitzGerald, Rahul Gupta, Wael
  Hamza, Haidar Khan, Charith Peris, Stephen Rawls, Andy Rosenbaum, Anna
  Rumshisky, Chandana~Satya Prakash, Mukund Sridhar, Fabian Triefenbach, Apurv
  Verma, Gokhan Tur, and Prem Natarajan. 2022.
\newblock \href {https://doi.org/10.48550/ARXIV.2208.01448} {Alexatm 20b:
  Few-shot learning using a large-scale multilingual seq2seq model}.

\bibitem[{Srivastava et~al.(2022)Srivastava, Rastogi, Rao, Shoeb, Abid, Fisch,
  Brown, Santoro, Gupta, Garriga-Alonso et~al.}]{srivastava2022beyond}
Aarohi Srivastava, Abhinav Rastogi, Abhishek Rao, Abu Awal~Md Shoeb, Abubakar
  Abid, Adam Fisch, Adam~R Brown, Adam Santoro, Aditya Gupta, Adri{\`a}
  Garriga-Alonso, et~al. 2022.
\newblock Beyond the imitation game: Quantifying and extrapolating the
  capabilities of language models.
\newblock \emph{arXiv preprint arXiv:2206.04615}.

\bibitem[{Su et~al.(2022)Su, Zhou, Yu, Chen, Zhu, Yu, and Zhou}]{su2022welm}
Hui Su, Xiao Zhou, Houjing Yu, Yuwen Chen, Zilin Zhu, Yang Yu, and Jie Zhou.
  2022.
\newblock Welm: A well-read pre-trained language model for chinese.
\newblock \emph{arXiv preprint arXiv:2209.10372}.

\bibitem[{Sun et~al.(2020)Sun, Yu, Yu, and Cardie}]{sun2020investigating}
Kai Sun, Dian Yu, Dong Yu, and Claire Cardie. 2020.
\newblock \href {https://transacl.org/ojs/index.php/tacl/article/view/1882}
  {Investigating prior knowledge for challenging chinese machine reading
  comprehension}.
\newblock \emph{Trans. Assoc. Comput. Linguistics}, 8:141--155.

\bibitem[{Suzgun et~al.(2022)Suzgun, Scales, Sch{\"a}rli, Gehrmann, Tay, Chung,
  Chowdhery, Le, Chi, Zhou et~al.}]{suzgun2022challenging}
Mirac Suzgun, Nathan Scales, Nathanael Sch{\"a}rli, Sebastian Gehrmann, Yi~Tay,
  Hyung~Won Chung, Aakanksha Chowdhery, Quoc~V Le, Ed~H Chi, Denny Zhou, et~al.
  2022.
\newblock Challenging big-bench tasks and whether chain-of-thought can solve
  them.
\newblock \emph{arXiv preprint arXiv:2210.09261}.

\bibitem[{Tay et~al.(2022{\natexlab{a}})Tay, Dehghani, Tran, Garcia, Bahri,
  Schuster, Zheng, Houlsby, and Metzler}]{tay2022unifying}
Yi~Tay, Mostafa Dehghani, Vinh~Q Tran, Xavier Garcia, Dara Bahri, Tal Schuster,
  Huaixiu~Steven Zheng, Neil Houlsby, and Donald Metzler. 2022{\natexlab{a}}.
\newblock Unifying language learning paradigms.
\newblock \emph{arXiv preprint arXiv:2205.05131}.

\bibitem[{Tay et~al.(2022{\natexlab{b}})Tay, Dehghani, Tran, Garcia, Wei, Wang,
  Chung, Bahri, Schuster, Zheng et~al.}]{tay2022ul2}
Yi~Tay, Mostafa Dehghani, Vinh~Q Tran, Xavier Garcia, Jason Wei, Xuezhi Wang,
  Hyung~Won Chung, Dara Bahri, Tal Schuster, Steven Zheng, et~al.
  2022{\natexlab{b}}.
\newblock Ul2: Unifying language learning paradigms.
\newblock In \emph{The Eleventh International Conference on Learning
  Representations}.

\bibitem[{Tay et~al.(2022{\natexlab{c}})Tay, Wei, Chung, Tran, So, Shakeri,
  Garcia, Zheng, Rao, Chowdhery et~al.}]{tay2022transcending}
Yi~Tay, Jason Wei, Hyung~Won Chung, Vinh~Q Tran, David~R So, Siamak Shakeri,
  Xavier Garcia, Huaixiu~Steven Zheng, Jinfeng Rao, Aakanksha Chowdhery, et~al.
  2022{\natexlab{c}}.
\newblock Transcending scaling laws with 0.1\% extra compute.
\newblock \emph{arXiv preprint arXiv:2210.11399}.

\bibitem[{Taylor et~al.(2022)Taylor, Kardas, Cucurull, Scialom, Hartshorn,
  Saravia, Poulton, Kerkez, and Stojnic}]{taylor2022galactica}
Ross Taylor, Marcin Kardas, Guillem Cucurull, Thomas Scialom, Anthony
  Hartshorn, Elvis Saravia, Andrew Poulton, Viktor Kerkez, and Robert Stojnic.
  2022.
\newblock Galactica: A large language model for science.
\newblock \emph{arXiv preprint arXiv:2211.09085}.

\bibitem[{Tiedemann(2020)}]{tiedemann-2020-tatoeba}
J{"o}rg Tiedemann. 2020.
\newblock \href {https://aclanthology.org/2020.wmt-1.139} {The {T}atoeba
  {T}ranslation {C}hallenge {--} {R}ealistic data sets for low resource and
  multilingual {MT}}.
\newblock In \emph{Proceedings of the Fifth Conference on Machine Translation},
  pages 1174--1182. Association for Computational Linguistics.

\bibitem[{Tikhonov and Ryabinin(2021)}]{tikhonov2021heads}
Alexey Tikhonov and Max Ryabinin. 2021.
\newblock \href {http://arxiv.org/abs/2106.12066} {It's all in the heads: Using
  attention heads as a baseline for cross-lingual transfer in commonsense
  reasoning}.

\bibitem[{Touvron et~al.(2023)Touvron, Lavril, Izacard, Martinet, Lachaux,
  Lacroix, Rozi{\`e}re, Goyal, Hambro, Azhar et~al.}]{touvron2023llama}
Hugo Touvron, Thibaut Lavril, Gautier Izacard, Xavier Martinet, Marie-Anne
  Lachaux, Timoth{\'e}e Lacroix, Baptiste Rozi{\`e}re, Naman Goyal, Eric
  Hambro, Faisal Azhar, et~al. 2023.
\newblock Llama: Open and efficient foundation language models.
\newblock \emph{arXiv preprint arXiv:2302.13971}.

\bibitem[{Vaswani et~al.(2017)Vaswani, Shazeer, Parmar, Uszkoreit, Jones,
  Gomez, Kaiser, and Polosukhin}]{vaswani2017attention}
Ashish Vaswani, Noam Shazeer, Niki Parmar, Jakob Uszkoreit, Llion Jones,
  Aidan~N Gomez, {\L}ukasz Kaiser, and Illia Polosukhin. 2017.
\newblock Attention is all you need.
\newblock \emph{Advances in neural information processing systems}, 30.

\bibitem[{Vu et~al.(2022)Vu, Barua, Lester, Cer, Iyyer, and
  Constant}]{vu2022overcoming}
Tu~Vu, Aditya Barua, Brian Lester, Daniel Cer, Mohit Iyyer, and Noah Constant.
  2022.
\newblock Overcoming catastrophic forgetting in zero-shot cross-lingual
  generation.
\newblock \emph{arXiv preprint arXiv:2205.12647}.

\bibitem[{Wang and Komatsuzaki(2021)}]{wang2021gpt}
Ben Wang and Aran Komatsuzaki. 2021.
\newblock Gpt-j-6b: A 6 billion parameter autoregressive language model.

\bibitem[{Wang et~al.(2022{\natexlab{a}})Wang, Roberts, Hesslow, Scao, Chung,
  Beltagy, Launay, and Raffel}]{wang2022language}
Thomas Wang, Adam Roberts, Daniel Hesslow, Teven~Le Scao, Hyung~Won Chung,
  Iz~Beltagy, Julien Launay, and Colin Raffel. 2022{\natexlab{a}}.
\newblock What language model architecture and pretraining objective work best
  for zero-shot generalization?
\newblock \emph{arXiv preprint arXiv:2204.05832}.

\bibitem[{Wang et~al.(2022{\natexlab{b}})Wang, Kordi, Mishra, Liu, Smith,
  Khashabi, and Hajishirzi}]{wang2022self}
Yizhong Wang, Yeganeh Kordi, Swaroop Mishra, Alisa Liu, Noah~A Smith, Daniel
  Khashabi, and Hannaneh Hajishirzi. 2022{\natexlab{b}}.
\newblock Self-instruct: Aligning language model with self generated
  instructions.
\newblock \emph{arXiv preprint arXiv:2212.10560}.

\bibitem[{Wang et~al.(2022{\natexlab{c}})Wang, Mishra, Alipoormolabashi, Kordi,
  Mirzaei, Arunkumar, Ashok, Dhanasekaran, Naik, Stap, Pathak, Karamanolakis,
  Lai, Purohit, Mondal, Anderson, Kuznia, Doshi, Patel, Pal, Moradshahi,
  Parmar, Purohit, Varshney, Kaza, Verma, Puri, Karia, Sampat, Doshi, Mishra,
  Reddy, Patro, Dixit, Shen, Baral, Choi, Smith, Hajishirzi, and
  Khashabi}]{wang2022super}
Yizhong Wang, Swaroop Mishra, Pegah Alipoormolabashi, Yeganeh Kordi, Amirreza
  Mirzaei, Anjana Arunkumar, Arjun Ashok, Arut~Selvan Dhanasekaran, Atharva
  Naik, David Stap, Eshaan Pathak, Giannis Karamanolakis, Haizhi~Gary Lai,
  Ishan Purohit, Ishani Mondal, Jacob Anderson, Kirby Kuznia, Krima Doshi,
  Maitreya Patel, Kuntal~Kumar Pal, Mehrad Moradshahi, Mihir Parmar, Mirali
  Purohit, Neeraj Varshney, Phani~Rohitha Kaza, Pulkit Verma, Ravsehaj~Singh
  Puri, Rushang Karia, Shailaja~Keyur Sampat, Savan Doshi, Siddhartha Mishra,
  Sujan Reddy, Sumanta Patro, Tanay Dixit, Xudong Shen, Chitta Baral, Yejin
  Choi, Noah~A. Smith, Hannaneh Hajishirzi, and Daniel Khashabi.
  2022{\natexlab{c}}.
\newblock Super-naturalinstructions: Generalization via declarative
  instructions on 1600+ nlp tasks.
\newblock \emph{arXiv preprint arXiv:2204.07705}.

\bibitem[{Wang et~al.(2022{\natexlab{d}})Wang, Pan, Yu, Yu, Chen, and
  Ji}]{wang2022zemi}
Zhenhailong Wang, Xiaoman Pan, Dian Yu, Dong Yu, Jianshu Chen, and Heng Ji.
  2022{\natexlab{d}}.
\newblock Zemi: Learning zero-shot semi-parametric language models from
  multiple tasks.
\newblock \emph{arXiv preprint arXiv:2210.00185}.

\bibitem[{Webson and Pavlick(2021)}]{webson-pavlick-2021}
Albert Webson and Ellie Pavlick. 2021.
\newblock \href {http://arxiv.org/abs/2109.01247} {Do prompt-based models
  really understand the meaning of their prompts?}

\bibitem[{Wei et~al.(2021)Wei, Bosma, Zhao, Guu, Yu, Lester, Du, Dai, and
  Le}]{wei2021finetuned}
Jason Wei, Maarten Bosma, Vincent~Y Zhao, Kelvin Guu, Adams~Wei Yu, Brian
  Lester, Nan Du, Andrew~M Dai, and Quoc~V Le. 2021.
\newblock Finetuned language models are zero-shot learners.
\newblock \emph{arXiv preprint arXiv:2109.01652}.

\bibitem[{Williams et~al.(2018)Williams, Nangia, and Bowman}]{N18-1101}
Adina Williams, Nikita Nangia, and Samuel Bowman. 2018.
\newblock \href {http://aclweb.org/anthology/N18-1101} {A broad-coverage
  challenge corpus for sentence understanding through inference}.
\newblock In \emph{Proceedings of the 2018 Conference of the North American
  Chapter of the Association for Computational Linguistics: Human Language
  Technologies, Volume 1 (Long Papers)}, pages 1112--1122. Association for
  Computational Linguistics.

\bibitem[{Wu and Dredze(2019)}]{wu-dredze-2019-beto}
Shijie Wu and Mark Dredze. 2019.
\newblock \href {https://doi.org/10.18653/v1/D19-1077} {Beto, bentz, becas: The
  surprising cross-lingual effectiveness of {BERT}}.
\newblock In \emph{Proceedings of the 2019 Conference on Empirical Methods in
  Natural Language Processing and the 9th International Joint Conference on
  Natural Language Processing (EMNLP-IJCNLP)}, pages 833--844, Hong Kong,
  China. Association for Computational Linguistics.

\bibitem[{Xu et~al.(2020)Xu, Hu, Zhang, Li, Cao, Li, Xu, Sun, Yu, Yu
  et~al.}]{xu2020clue}
Liang Xu, Hai Hu, Xuanwei Zhang, Lu~Li, Chenjie Cao, Yudong Li, Yechen Xu, Kai
  Sun, Dian Yu, Cong Yu, et~al. 2020.
\newblock Clue: A chinese language understanding evaluation benchmark.
\newblock \emph{arXiv preprint arXiv:2004.05986}.

\bibitem[{Xue et~al.(2020)Xue, Constant, Roberts, Kale, Al-Rfou, Siddhant,
  Barua, and Raffel}]{xue2020mt5}
Linting Xue, Noah Constant, Adam Roberts, Mihir Kale, Rami Al-Rfou, Aditya
  Siddhant, Aditya Barua, and Colin Raffel. 2020.
\newblock mt5: A massively multilingual pre-trained text-to-text transformer.
\newblock \emph{arXiv preprint arXiv:2010.11934}.

\bibitem[{Yang et~al.(2019)Yang, Zhang, Tar, and Baldridge}]{pawsx2019emnlp}
Yinfei Yang, Yuan Zhang, Chris Tar, and Jason Baldridge. 2019.
\newblock {PAWS-X: A Cross-lingual Adversarial Dataset for Paraphrase
  Identification}.
\newblock In \emph{Proc. of EMNLP}.

\bibitem[{Yong and Nikoulina(2022)}]{yong2022adapting}
Zheng-Xin Yong and Vassilina Nikoulina. 2022.
\newblock Adapting bigscience multilingual model to unseen languages.
\newblock \emph{arXiv preprint arXiv:2204.04873}.

\bibitem[{Yong et~al.(2022)Yong, Schoelkopf, Muennighoff, Aji, Adelani,
  Almubarak, Bari, Sutawika, Kasai, Baruwa et~al.}]{yong2022bloom+}
Zheng-Xin Yong, Hailey Schoelkopf, Niklas Muennighoff, Alham~Fikri Aji,
  David~Ifeoluwa Adelani, Khalid Almubarak, M~Saiful Bari, Lintang Sutawika,
  Jungo Kasai, Ahmed Baruwa, et~al. 2022.
\newblock Bloom+ 1: Adding language support to bloom for zero-shot prompting.
\newblock \emph{arXiv preprint arXiv:2212.09535}.

\bibitem[{Zaken et~al.(2021)Zaken, Ravfogel, and Goldberg}]{zaken2021bitfit}
Elad~Ben Zaken, Shauli Ravfogel, and Yoav Goldberg. 2021.
\newblock Bitfit: Simple parameter-efficient fine-tuning for transformer-based
  masked language-models.
\newblock \emph{arXiv preprint arXiv:2106.10199}.

\bibitem[{Zeng et~al.(2022)Zeng, Liu, Du, Wang, Lai, Ding, Yang, Xu, Zheng, Xia
  et~al.}]{zeng2022glm}
Aohan Zeng, Xiao Liu, Zhengxiao Du, Zihan Wang, Hanyu Lai, Ming Ding, Zhuoyi
  Yang, Yifan Xu, Wendi Zheng, Xiao Xia, et~al. 2022.
\newblock Glm-130b: An open bilingual pre-trained model.
\newblock \emph{arXiv preprint arXiv:2210.02414}.

\bibitem[{Zhang et~al.(2022)Zhang, Roller, Goyal, Artetxe, Chen, Chen, Dewan,
  Diab, Li, Lin et~al.}]{zhang2022opt}
Susan Zhang, Stephen Roller, Naman Goyal, Mikel Artetxe, Moya Chen, Shuohui
  Chen, Christopher Dewan, Mona Diab, Xian Li, Xi~Victoria Lin, et~al. 2022.
\newblock Opt: Open pre-trained transformer language models.
\newblock \emph{arXiv preprint arXiv:2205.01068}.

\bibitem[{Zhang et~al.(2019)Zhang, Baldridge, and He}]{zhang2019paws}
Yuan Zhang, Jason Baldridge, and Luheng He. 2019.
\newblock Paws: Paraphrase adversaries from word scrambling.
\newblock \emph{arXiv preprint arXiv:1904.01130}.

\bibitem[{Zhong et~al.(2021)Zhong, Lee, Zhang, and Klein}]{zhong2021adapting}
Ruiqi Zhong, Kristy Lee, Zheng Zhang, and Dan Klein. 2021.
\newblock \href {http://arxiv.org/abs/2104.04670} {Meta-tuning language models
  to answer prompts better}.
\newblock \emph{CoRR}, abs/2104.04670.

\bibitem[{Zhu et~al.(2022)Zhu, Jain, Suresh, Ravindran, Tipirneni, and
  Reddy}]{zhu2022xlcost}
Ming Zhu, Aneesh Jain, Karthik Suresh, Roshan Ravindran, Sindhu Tipirneni, and
  Chandan~K. Reddy. 2022.
\newblock \href {http://arxiv.org/abs/2206.08474} {Xlcost: A benchmark dataset
  for cross-lingual code intelligence}.

\end{thebibliography}
